\documentclass[stsy,nonblindrev]{informs3}

\OneAndAHalfSpacedXI
\RequirePackage[numbers]{natbib}
\RequirePackage[colorlinks,citecolor=blue,linkcolor=blue,urlcolor=blue,hypertexnames=false]{hyperref}
\RequirePackage{graphicx}
\usepackage{float}
\usepackage{caption}
\usepackage{subcaption}
\usepackage{pdfpages}
\usepackage[dvipsnames]{xcolor}

\newcommand\abs[1]{\left|#1\right|}
\def\indic#1{\mathbb{I}\left({#1}\right)}
\def\E{\mathbb{E}}
\def\P{\mathbb{P}}
\providecommand{\argmax}{\mathop\mathrm{arg max}}

\usepackage{natbib}
 \bibpunct[, ]{(}{)}{,}{a}{}{,}%
 \def\bibfont{\small}%
\TheoremsNumberedThrough     % Preferred (Theorem 1, Lemma 1, Theorem 2)
\ECRepeatTheorems

\EquationsNumberedThrough    % Default: (1), (2), ...
\begin{document}
%%%%%%%%%%%%%%%%

%\setcounter{page}{1}
%\DOI{10.1287/xxxx.0000.0000}%


\RUNTITLE{Diffusion Approximations for Thompson Sampling in the Small Gap Regime}

% \TITLE{Bundling Information Goods of Decreasing Value}
\TITLE{Diffusion Approximations for Thompson Sampling in the Small Gap Regime}

\ARTICLEAUTHORS{%
\AUTHOR{Lin Fan}
\AFF{Kellogg School of Management, Northwestern University, Evanston, IL 60208, \EMAIL{lin.fan@kellogg.northwestern.edu}} %, \URL{}}
\AUTHOR{Peter W. Glynn}
\AFF{Department of Management Science and Engineering, Stanford University, Stanford, CA 94305, \EMAIL{glynn@stanford.edu}}
} % end of the block

\ABSTRACT{%
% Enter your abstract
We study the process-level dynamics of Thompson sampling and related sampling-based bandit algorithms in the ``small gap'' regime, where the gaps between the arm means are of order $\sqrt{\gamma}$ or smaller and the time horizon is of order $1/\gamma$, with $\gamma \downarrow 0$.
In this regime, as $\gamma \downarrow 0$, we show that the process-level dynamics of such algorithms converge weakly to the solutions to certain stochastic differential equations and stochastic ordinary differential equations.
Our weak convergence theory is developed using the Continuous Mapping Theorem, which provides a direct and modular theoretical approach that can be adapted to analyze a variety of sampling-based bandit algorithms and handle weakly dependent reward processes.
A central finding is an algorithmic invariance principle: in the small gap regime, the limit dynamics of a broad class of sampling-based algorithms---including Thompson sampling with general single-parameter exponential family likelihoods, as well as non-parametric bandit algorithms based on bootstrap re-sampling---all coincide with those of Thompson sampling with Gaussian likelihoods.
Moreover, in the small gap regime, the regret performance of these algorithms is generally insensitive to model mis-specification, changing continuously with increasing degrees of mis-specification.
}%

%\FUNDING{This research was supported by [grant number, funding agency].}

%Supplemental Material:
%Data Ethics & Reproducibility Note:

%\KEYWORDS{Stochastic programming, Decision support,Uncertainty, Disaster response, Optimization}

%\HISTORY{Received: Month DD, YYYY; Accepted: Month DD, YYYY; Published Online: Month DD, YYYY}

\maketitle
%%%%%%%%%%%%%%%%%%%%%%%%%%%%%%%%%%%%%%%%%%%%%%%%%%%%%%%%%%%%%%%%%%%%%%

% Text of your paper here

\section{Introduction} \label{diffusion_introduction}

The multi-armed bandit problem is a widely studied model that is both useful in practical applications and a valuable theoretical paradigm exhibiting the trade-off between exploration and exploitation in sequential decision-making under uncertainty.
Theoretical research in this area has focused overwhelmingly on studying the performance of algorithms through establishing upper and lower bounds on the expected (pseudo-)regret; see \cite{lattimore_etal2020} for a recent detailed account of bandit theory.
The regret $\text{Reg}(n) := \sum_k N_k(n) \Delta_k$ is the sum over each arm $k$ of the number of times $N_k(n)$ it is played over time horizon $n$, weighted by its mean reward sub-optimality gap $\Delta_k := \max_j \mu_j - \mu_k$, where $\mu_j$ is the mean reward of arm $j$.
While expected regret $\E[\text{Reg}(n)]$ is the most fundamental performance measure, the performance implications of $\text{Reg}(n)$ depend on aspects of its distribution beyond the mean, which may be crucial to understand in some applications.
For example, in settings where bandit algorithms are deployed with only a limited number of runs so that the law of large numbers does not ``kick in'', or in settings where risk sensitivity is a key concern, understanding the variability of $\text{Reg}(n)$ can be as important as understanding $\E[\text{Reg}(n)]$ for designing effective algorithms.

In this paper, our main focus is on Thompson sampling (TS) \citep{thompson_1933}, which is a Bayesian approach for balancing exploration and exploitation that has become one of the most popular bandit algorithms \citep{chapelle_etal2011, agrawal_etal2012, kaufmann_etal2012, russo_etal2014, russo_etal2016, russo_etal2019, russo_2020}.
The TS principle specifies that at any given time, an arm is played with probability equal to the posterior probability that its mean reward is the highest among all arms; a precise description of TS is provided in Section \ref{diffusion_mab_0}.
Our specific interest is in studying algorithm behavior in the challenging ``small gap'' regime in which the sub-optimality gaps $\Delta_k$ are of order $\sqrt{\gamma}$ or smaller and the time horizon (total number of arm plays) $n$ is large and of order $1/\gamma$, with $\gamma \downarrow 0$.
Thus, this analysis provides insight into algorithm behavior when $n$ is not yet large enough to have confidently identified the optimal arm.
Sending $\gamma \downarrow 0$, we show that the dynamics of TS (and related sampling-based bandit algorithms), viewed as stochastic processes, converge weakly (in distribution) to a diffusion process characterized by stochastic differential equations (SDEs) and (equivalently) stochastic ordinary differential equations (ODEs) (also known as ``random time change'' equations; see, e.g., Chapter 6 of \cite{ethier_etal1986}).

This small gap regime is closely related to so-called ``minimax'' (also called ``worst-case'') settings in the bandit literature, which is one of the key settings that guides the optimal design of bandit algorithms; see Chapters 15-16 of \cite{lattimore_etal2020}.
For illustration, consider the Gaussian Thompson sampler, i.e., the TS principle implemented using the posterior updating mechanics of Gaussian likelihoods and priors, which is known to have nearly minimax optimal expected regret.
For the Gaussian Thompson sampler, minimax bandit settings are ``statistically hardest'' and have sub-optimality gaps $\Delta_k$ of order $1/\sqrt{n}$ for a time horizon of $n$ (suppressing the dependence on the total number of arms $K$, which we treat as a fixed constant in this paper) \citep{agrawal_etal2013b, agrawal_etal2017}.
In such settings, there is not enough statistical information in the rewards collected for bandit algorithms to fully distinguish between sub-optimal and optimal arms.
So, essentially all arms with order $1/\sqrt{n}$ sub-optimality gaps are played $O_{\P}(n)$ times over a horizon of $n$, resulting in $O_{\P}(\sqrt{n})$ regret overall.
Moreover, as alluded to above, the study of such settings provides insight about the early stages of general bandit experiments, when algorithms are just starting to be able to distinguish between arms.

Our main contributions in this paper are summarized in points 1)-3) below.
Recently, \cite{kuang_etal2021}, working within a general framework for sequential experimentation with iid rewards, independently established similar results on weak convergence to SDEs and limit representation by stochastic ODEs for versions of the Gaussian Thompson sampler, which correspond most directly to our Theorem \ref{diffusion_thm1} and Proposition \ref{time_change_representation_conversion}.
In Section \ref{diffusion_related}, we discuss this overlap and the main differences between the two papers with respect to these two results.
Beyond this overlap, the two papers are developed in complementary ways, each with a different focus.

\begin{enumerate}
    \item[1)] We develop a direct and modular weak convergence analysis approach for establishing SDE and stochastic ODE approximations for the dynamics of sampling-based bandit algorithms, including TS and related algorithms, in the small gap regime.
    We start with explicit discrete-time evolution equations describing the dynamics of the bandit systems, and after re-scaling, directly pass to the limit using the Continuous Mapping Theorem (CMT) and elementary arguments to obtain the SDE and stochastic ODE weak limits.
    Our CMT-based approach shows concretely how the weak limits arise from the pre-limit dynamics, which in turn allows the analysis to be readily adapted to new algorithms and reward structures.
    Another major benefit of our CMT-based approach is its modularity.
    In ``rested bandit'' settings, we can establish weak convergence of each arm's centered and re-scaled cumulative reward process to Brownian motion separately from the analysis of the algorithm’s sampling behavior.
    These capabilities are the key to our subsequent developments in Contributions 2) and 3) below---allowing us to handle more general bandit settings going well beyond the Gaussian Thompson sampler with iid rewards, including more complex sampling-based bandit algorithms and (weakly) dependent reward processes.
    \item[2)] We apply our CMT-based approach to study in detail the process-level dynamics of versions of the Gaussian Thompson sampler in the small gap regime, establishing weak convergence to SDEs (Theorems \ref{diffusion_thm1} and \ref{diffusion_thm3} in Sections \ref{diffusion_mab_1} and \ref{diffusion_additional}) and also weak convergence to stochastic ODEs (Theorems \ref{diffusion_thm2} and \ref{diffusion_thm3} in Sections \ref{diffusion_mab_2} and \ref{diffusion_additional}).
    These diffusion approximations essentially only require that the centered and re-scaled cumulative reward processes converge weakly to Brownian motion; the rewards do not need to be Gaussian or even iid---Theorem \ref{diffusion_thm2} is developed for stationary, weakly dependent reward processes in rested bandit settings.
    At the limit level, we establish the full (distributional) equivalence between corresponding SDE and stochastic ODE representations, specifically showing how a solution to an SDE can be expressed as a solution to the corresponding stochastic ODE (Proposition \ref{time_change_representation_conversion} in Section \ref{diffusion_mab_2}) and vice versa (Theorem \ref{diffusion_thm_sde_sode_equivalence} in Section \ref{diffusion_mab_2}).
    Importantly, Theorem \ref{diffusion_thm_sde_sode_equivalence} provides a uniqueness theory for solutions to stochastic ODEs, leveraging uniqueness theory for SDEs.
    Moreover, we provide a rigorous treatment of stochastic ODE theory in Section \ref{diffusion_mab_2}.
    These fundamental aspects of stochastic ODEs are needed for our Theorems \ref{diffusion_thm2} and \ref{diffusion_thm3}, and are also used in subsequent developments.
    \item[3)] Leveraging our CMT-based approach and building on our analysis of the Gaussian Thompson sampler, we develop an \textit{algorithmic invariance principle} that is satisfied by many versions of TS and related sampling-based bandit algorithms in the small gap regime.
    Specifically, the limit dynamics of such algorithms, as captured by their corresponding SDEs and stochastic ODEs, all coincide with those of the Gaussian Thompson sampler.
    Thus, when sub-optimality gaps are small, the Gaussian Thompson sampler provides unifying insight about the sampling behavior and process-level dynamics of many versions of TS and related sampling-based bandit algorithms.
    (Note that our algorithmic invariance principle is predominantly an approximation of the sampling behavior---how exploration and exploitation are balanced---of such algorithms. There is, in a sense, a secondary aspect of ``invariance'', where the centered and re-scaled cumulative reward processes converge weakly to Brownian motions as $\gamma \downarrow 0$ in the small gap regime.)
    Our algorithmic invariance principle includes \textit{exponential family (EF) Thompson samplers}, i.e., the TS principle implemented using the posterior updating mechanics of a general single-parameter exponential family likelihood and a general (bounded support) prior distribution (Theorem \ref{diffusion_prop3} in Section \ref{diffusion_general}).
    It also includes the \textit{non-parametric bootstrap sampler}, which is similar to the TS principle, but involves non-parametric bootstrap re-sampling instead of posterior sampling (Theorem \ref{diffusion_prop5} in Section \ref{diffusion_bootstrap}).
    For these results, we develop custom Gaussian approximations for posterior distributions and bootstrap sampling distributions that hold uniformly over data-generating distributions (Propositions \ref{diffusion_prop1} and \ref{diffusion_prop2} in Appendices \ref{diffusion_appA1} and \ref{diffusion_appA2}).
    Additionally, in the small gap regime, the regret performance of these algorithms is insensitive to mis-specification of reward distributions, changing continuously with increasing degrees of mis-specification (Proposition \ref{diffusion_prop4} in Section \ref{diffusion_mis-specification}).
    This contrasts with the ``instance-based'' bandit setting of \cite{lai_etal1985}, corresponding to time scales that are large relative to $1/\gamma$, over which optimized bandit algorithms achieve expected regret growing logarithmically with time.
    In that setting, regret is sensitive to small probability mis-identification of the optimal arm, and the algorithms become highly sensitive to even slight degrees of model mis-specification \citep{fan_etal2021}.
\end{enumerate}

The rest of the paper is structured as follows.
Related work is further discussed in Section \ref{diffusion_related}.
We then introduce the model and setup used throughout the paper in Section \ref{diffusion_mab_0}.
In Section \ref{diffusion_mab_1}, we provide an intuitive derivation leading to the SDE convergence result in Theorem \ref{diffusion_thm1} for the Gaussian Thompson sampler and iid reward processes, with the proofs given in Section \ref{diffusion_sde}.
Similarly, in Section \ref{diffusion_mab_2}, we provide an intuitive derivation leading to the stochastic ODE convergence result in Theorem \ref{diffusion_thm2} for the Gaussian Thompson sampler and general stationary reward processes, with the proofs given in Section \ref{diffusion_rode}.
Along the way, we develop Theorem \ref{diffusion_thm_sde_sode_equivalence} and Proposition \ref{time_change_representation_conversion}, which establish distributional equivalences between the solution to the SDE and solutions to the corresponding stochastic ODE, with the proofs given in Section \ref{diffusion_rode}.
In Section \ref{diffusion_additional}, we develop extensions in Corollary \ref{diffusion_prop00} and Theorem \ref{diffusion_thm3} to the earlier diffusion approximations.
In Section \ref{diffusion_general}, we show via Theorem \ref{diffusion_prop3} that the EF Thompson sampler has the same weak limit in the small gap regime as the Gaussian Thompson sampler.
In Section \ref{diffusion_bootstrap}, the same is shown via Theorem \ref{diffusion_prop5} for the non-parametric bootstrap sampler.  
The proofs for Theorems \ref{diffusion_prop3} and \ref{diffusion_prop5} are given in Section \ref{sec:sections_2_4_proofs}; as supporting results, the Gaussian approximation for posterior distributions in Proposition \ref{diffusion_prop1} and bootstrap sampling distributions in Proposition \ref{diffusion_prop2} are developed in Appendices \ref{diffusion_appA1} and \ref{diffusion_appA2}, respectively.
In Section \ref{diffusion_mis-specification}, we study the insensitivity of these sampling-based bandit algorithms in the small gap regime to mis-specification of the reward distribution in Proposition \ref{diffusion_prop4}.
We then conclude the paper with a quick study of batched updating in Section \ref{diffusion_batched}.
Useful technical lemmas can be found in Appendix \ref{diffusion_appC}.

\subsection{Related Work} \label{diffusion_related}

While completing the initial version of our paper, we became aware of the independent and concurrent work of \cite{kuang_etal2021} (abbreviated KW in the discussion below), which was posted on arXiv \citep{wager_xu_arXiv} prior to our manuscript.
Our present paper is an updated and expanded version of our previous arXiv postings, and also forms a chapter of the first author's doctoral dissertation \citep{fan_phd2023}.

As mentioned earlier, the overlap between our work and KW is that both obtain similar results on weak convergence to SDEs and limit representation by stochastic ODEs for the dynamics of versions of the Gaussian Thompson sampler in small gap regimes with iid rewards.
In terms of formal theoretical results, the overlap is essentially between our Theorem \ref{diffusion_thm1} and KW Theorem 1 (applied to the Gaussian Thompson sampler), and also our Proposition \ref{time_change_representation_conversion} and KW Theorem 3.
Below, we discuss the key differences between our work and KW that are relevant to this overlap.
We discuss additional related work later in this section.

Both our work and KW establish that a solution to an SDE can be expressed as a solution to the corresponding stochastic ODE (our Proposition \ref{time_change_representation_conversion} and KW Theorem 3, obtained via different proofs).
However, as discussed in Contribution 2), we additionally establish the converse direction in Theorem \ref{diffusion_thm_sde_sode_equivalence}, provide a rigorous treatment of stochastic ODEs in Section \ref{diffusion_mab_2}, and develop the theory for weak convergence to stochastic ODEs.
Besides a more complete mathematical description of the limit and theory of weak convergence, our developments are necessary for establishing diffusion approximations in more general settings involving, e.g., weakly dependent reward processes in rested bandits.

Notably, the theoretical approach for establishing weak convergence in our paper differs from that in KW.
The KW paper formulates the \textit{Sequentially Randomized Markov Experiments} framework, which models the dynamics of sampling-based bandit algorithms (e.g., tempered greedy, Luce's rule \citep{luce_1959}, Gaussian Thompson sampler, and exploration sampling \citep{kasy_etal2021}; see their examples 1-4) for iid reward processes as Markov chains, with states consisting of the number of plays of each arm and the cumulative rewards received for each arm.
Then, in a small gap regime, KW establish weak convergence of re-scaled versions of the Markov chains to SDEs by establishing convergence of the infinitesimal generators of the re-scaled Markov chains, leveraging general theory in \cite{stroock_etal1979}.

By contrast, our CMT-based approach is developed from the ground up, which together with its explicit construction of weak limits as discussed in Contribution 1), allows it to be readily modified to suit new bandit settings, including new algorithms and reward structures.
We restrict the scope of our paper to versions of TS and related sampling-based bandit algorithms, though our CMT-based approach also applies to the Sequentially Randomized Markov Experiments algorithm class of KW.

Another significant advantage of our CMT-based approach is its modularity, as discussed in Contribution 1).
Moreover, our approach does not require the pre-limit system to have any Markovian structure. 
Together, these two aspects allow us to develop diffusion approximations in more general bandit settings involving, e.g., weakly dependent reward processes (our Theorem \ref{diffusion_thm2}) or more complex algorithms like the non-parametric bootstrap sampler (our Theorem \ref{diffusion_prop5}).
By contrast, with a generator-based approach, the pre-limit system must be Markovian (or approximately Markovian), which means that any additional structure---such as serial dependence in rewards or the re-sampling mechanism of the bootstrap---must be accounted for in the state.
For instance, for the non-parametric bootstrap sampler under iid rewards, the empirical CDF of each arm's rewards would need to be part of the Markov state, substantially complicating the generator convergence analysis.

On the other hand, in addition to their framework, KW develop a number of
results and insights that we do not pursue. 
In particular, KW use their diffusion limits to study regret consequences of sampling-based algorithms in their small gap regime.
For example, for a two-armed version of the Gaussian Thompson sampler, they find that regret behavior depends sharply on the prior-``smoothing'' parameter: when the diffusion-scaled gap is large, less informative, undersmoothed priors exhibit better regret behavior than smoothed priors with positive diffusion-scale prior precision.
They also study a related instability phenomenon near time zero on the diffusion time scale, showing that in one-armed adaptive experiments, sensitive sampling rules, including an undersmoothed Gaussian Thompson sampler, can exhibit large swings in sampling probabilities.

\vspace{2mm}
\noindent \textbf{Additional Related Work}

Related to both our work and KW, \cite{kalvit_etal2021} study a canonical version of the upper confidence bound (UCB) algorithm for two-armed bandit settings under various scalings of the sub-optimality gap with the time horizon $n$.
They show that UCB arm allocation fractions are asymptotically deterministic, derive sharp minimax regret asymptotics when the gap scales as $\sqrt{\log(n)/n}$, and obtain a process-level diffusion limit when the gap scales as $1/\sqrt{n}$.
Their results also highlight a qualitative contrast with TS: UCB arm allocations converge in probability to $1/2$ as $n \to \infty$ when the gap is $o(\sqrt{\log(n)/n})$, whereas TS can exhibit non-degenerate, imbalanced limiting allocations in certain zero-gap Bernoulli bandit settings.
Since UCB uses a deterministic, discontinuous allocation rule, our CMT-based approach and the generator-based approach of KW do not directly apply (in their present forms) to the UCB weak convergence results in \cite{kalvit_etal2021}.

Recently, \cite{hirano_etal2023} have developed asymptotic representation theorems for batched adaptive experiments, extending Le Cam's classical ``limits of experiments'' theory for non-adaptive experiments.
Using $1/\sqrt{n}$ local parametrizations, they obtain representations in terms of a limit bandit experiment, which can be used for local asymptotic power analysis for data obtained from batched bandit experiments.

Subsequent to our 2021 arXiv postings (and that of KW and \cite{kalvit_etal2021}), there have been additional works leveraging continuous-time scaling and diffusion approximations to study bandit problems, including \cite{zhu_etal2023}, \cite{adusumilli_2025}, and \cite{adusumilli_2026}. 
These works exploit the tractability of the limit approximations to study policy design for the pre-limit bandit systems and, in some cases, characterize near-optimal policies.

Stepping back, diffusion approximations have a long history in the Bayesian analysis of multi-armed bandit problems.
In the discounted infinite-horizon setting, \cite{chang_etal1987} use asymptotic analysis of Brownian optimal stopping boundaries to obtain approximations to Gittins indices as the discount factor tends to one.
Building on this limiting diffusion representation, \cite{brezzi_etal2002} develop simple closed-form approximations to these indices, and \cite{yao_2006} develops a corrected diffusion approximation for normal rewards processes using Chernoff's continuity correction.
These works differ from ours in their object of analysis---the Gittins/dynamic allocation indices, characterized via an optimal stopping problem, rather than the process-level dynamics of more heuristic sampling-based bandit algorithms---and in their scaling regime, which involves the discount factor tending to one rather than sub-optimality gaps going to zero.

Finally, a broader recent literature uses diffusion models and approximations to study adaptive experimentation and stopping problems.
\cite{chick_etal2009,chick_etal2012} develop economically-motivated Bayesian simulation selection procedures: their diffusion/free boundary analyses yield tractable approximations to optimal stopping in single-alternative settings and are used to construct sequential allocation and stopping procedures in multi-alternative settings.
\cite{harrison_etal2015} and \cite{araman_etal2022} study Bayesian sequential experimentation/control problems with binary priors and terminal decisions of when to stop learning and take an action; the former formulates a continuous time Bayesian learning model, whereas the latter derives a diffusion approximation in a high frequency, low informativeness regime.
These works differ from ours in their focus on optimal stopping/control and dynamic programming, rather than process-level weak limits for sampling-based bandit algorithms in the small gap regime.
Moreover, \cite{harrison_etal2015,araman_etal2022} focus on binary priors, whereas we consider general priors.

\section{Model and Preliminaries} \label{diffusion_mab_0}

\noindent \textbf{Bandit Problems and Thompson Sampling}

A general sampling-based bandit algorithm operates as follows.
We have a filtration $\mathcal{H} = (\mathcal{H}_j, \, j \ge 0)$ that the bandit process is adapted to, with
\begin{align}
    \mathcal{H}_j = \sigma(I(1),Y(1),\dots,I(j),Y(j)) \label{h_j}
\end{align} 
corresponding to the data collected through some time $j$, where at each time $i$ and for each arm $k \in [K] := \{1,\dots,K\}$, $I_k(i) = 1$ if arm $k$ is selected and otherwise $I_k(i) = 0$ (so that $\sum_k I_k(i) = 1$), and $Y(i)$ is the reward received for the selected arm.
For the settings in this paper, the data can be summarized by sufficient statistics $(N(j),G(j)) = ((N_k(j),G_k(j)), \, k \in [K])$ measurable with respect to $\mathcal{H}_j$, where for each arm $k \in [K]$,
\begin{align}
    N_k(j) = \sum_{i=1}^j I_k(i) \label{n_k}
\end{align}
is the number of plays and
\begin{align}
    G_k(j) = \sum_{i=1}^j I_k(i)Y(i) \label{g_k}
\end{align}
is the cumulative reward.

The algorithm selects an arm in the time period $j+1$ by generating $I(j+1)$ as an independent $K$-dimensional multinomial random variable with a single trial and success probability vector $\pi(N(j),G(j)) \in \Delta^K$, where $\Delta^K$ denotes the $K$-dimensional probability simplex and $\pi : \mathbb{N}^K \times \mathbb{R}^K \to \Delta^K$.
Given $I(j+1)$, a reward $Y(j+1)$ is received for the selected arm, and the sufficient statistics $(N(j+1),G(j+1))$ are updated accordingly.

TS is an important example of a sampling-based bandit algorithm and our primary focus throughout the paper.
When studying TS, we will restrict attention to TS designed for parametric reward models parameterized by mean.
The actual data presented to the algorithm will be assumed iid or weakly dependent.
(As mentioned in the Introduction, we will begin with the Gaussian Thompson sampler in Sections \ref{diffusion_mab_1} and \ref{diffusion_mab_2} before generalizing to EF Thompson samplers in Section \ref{diffusion_general}.)
As a Bayesian algorithm, TS maintains a posterior distribution for the mean reward of each arm, and in each time period, it samples a mean from each posterior and plays the arm corresponding to the highest sampled mean, after which a corresponding reward is received and the posterior is updated with the new information.
More precisely, for each arm $k$, we start with an independent prior distribution $\nu_k^0$ for the unknown mean $\mu_k$.
From posterior updating, at each time $j = 1,2,\dots$ and for each arm $k$, we have a posterior distribution $\nu_k(N_k(j),G_k(j))$, which depends on the sufficient statistics $(N_k(j),G_k(j))$ for that arm.
At time $j$, we draw an independent sample $\widetilde{\mu}_k(j) \sim \nu_k(N_k(j),G_k(j))$ for each arm $k$, and we play the arm $\argmax_k \widetilde{\mu}_k(j)$.
So, for TS, $\pi_k(N(j),G(j)) := \P(k = \argmax_l \widetilde{\mu}_l(j))$, i.e., each arm is played according to the posterior probability that it has the highest mean reward.

\vspace{2mm}
\noindent \textbf{Reward Feedback Mechanisms}

We consider two ways of generating reward feedback.
For each arm $k \in [K]$, let $X_k(i)$, $i = 1,2,\dots$ be the sequence of rewards for the arm.
The first way is called the \textit{random table model}, where at time $j$, $Y(j) = X_k(j)$ for the selected arm $k \in [K]$ ($I_k(j) = 1$).
The second way is called the \textit{reward stack model}, where at time $j$, $Y(j) = X_k(N_k(j-1)+1)$ for the selected arm $k \in [K]$ ($I_k(j) = 1$), where $N_k(j-1)$ is the number of plays of arm $k$ through time $j-1$, as defined in (\ref{n_k}) above.
(The random table and reward stack terminology is taken from \cite{lattimore_etal2020}; see Chapter 4.6, page 53.)

When the reward sequence for each arm is iid with independence across arms, the random table model and the reward stack model generate reward feedback in distributionally equivalent ways.
We will also consider the setting where for each arm $k \in [K]$, the rewards $X_k(i)$, $i = 1,2,\dots$ are a \textit{stationary sequence} (which allows for serial dependence), i.e., for any fixed integers $1 \le i_1 \le i_2 \le \dots \le i_l < \infty$, the finite-dimensional distributions $(X_k(i_1 + j),X_k(i_2 + j),\dots,X_k(i_l + j))$ are the same for all $j \ge 0$.
In the stationary rewards setting, we will use the reward stack model.
This leads to a \textit{rested bandit}, where the reward process for each arm is in steady state and evolves according to a stochastic process only when the arm is played, otherwise staying ``frozen''.

In Section \ref{diffusion_mab_1}, we will see how the random table model leads to an SDE characterization of TS dynamics.
In Section \ref{diffusion_mab_2}, we will see how the reward stack model leads to a stochastic ODE characterization, both for iid and stationary reward processes.

\vspace{2mm}
\noindent \textbf{Function Spaces and Weak Convergence}

Throughout this paper, $D_m[a,\infty)$ denotes the space of functions with domain $[a,\infty)$ and range $\mathbb{R}^m$ (writing $D[a,\infty)$ when $m=1$), that are right-continuous and have limits from the left.
We use the Skorohod metric on this space, as defined in Chapter 3, Section 5 of \cite{ethier_etal1986}.
Weak convergence is always denoted using $\Rightarrow$, both for stochastic processes taking values in $D_m[a,\infty)$ and for random variables taking values in $\mathbb{R}^m$.
Complete mathematical details for the spaces $D_m[a,\infty)$ equipped with the Skorohod metric, as well as the theory of weak convergence in such spaces, can be found in Chapter 3 of \cite{ethier_etal1986}.

\vspace{2mm}
\noindent \textbf{Small Gap Regime}

As mentioned in the Introduction, throughout the paper, we consider a sequence of bandit models indexed by a positive, real-valued parameter $\gamma$, with $\gamma \downarrow 0$.
We will consider bandit instances with arm mean separation on the scale of $\sqrt{\gamma}$, over time horizons on the scale of $1/\gamma$.
When working within the corresponding $\gamma$-scale system, we will write a $\gamma$ superscript on all objects defined previously to indicate we are working with the same object defined appropriately in the $\gamma$-scale system.
For any reward feedback mechanism, we will use the discrete-time filtration $\mathcal{H}^\gamma = (\mathcal{H}_j^\gamma, \, j \ge 0)$, with
\begin{align}
    \mathcal{H}_j^\gamma = \sigma(I^\gamma(1),Y^\gamma(1),\dots,I^\gamma(j),Y^\gamma(j)), \label{h_j_gamma}
\end{align}
to keep track of the algorithm's information.
Below, we state and discuss two setups for the small gap regime (Assumptions \ref{assumption0} and \ref{assumption1}) that we will use to develop our limit theory throughout the paper.

In the iid rewards setting, we will work under Assumption \ref{assumption0}, given as follows.
\begin{assumption}[\normalfont Small Gap Regime with IID Rewards] \label{assumption0}
For each $\gamma$ and each arm $k \in [K]$, the rewards $X_k^{\gamma}(i)$, $i = 1,2,\dots$ are independent across both $i$ and $k$, and are distributed according to $Q_k^{\gamma}$, with mean $\mu_k^{\gamma}$ and variance $(\sigma_k^{\gamma})^2$.
There exist some $\alpha > 0$, some $\mu_* \in \mathbb{R}$, and for each arm $k$, some fixed $d_k \in \mathbb{R}$, $\sigma_k > 0$ such that
\begin{align}
    & \mu_k^\gamma = \mu_* + \sqrt{\gamma} d_k^{\gamma}, \quad \lim_{\gamma \downarrow 0} d_k^{\gamma} = d_k \label{assumption0_mu} \\
    & \lim_{\gamma \downarrow 0} \sigma_k^{\gamma} = \sigma_k \label{assumption0_sigma} \\
    & \sup_{\gamma > 0} \E\left[\abs{X_k^{\gamma}(i)}^{2+\alpha}\right] < \infty. \label{assumption0_lyapunov}
\end{align}
\end{assumption}

\begin{remark} \label{rmk3}
    In the iid rewards setting, for our analysis, it suffices for the rewards to have uniformly bounded $2+\alpha$ moments, with arbitrarily small $\alpha > 0$ (as in (\ref{assumption0_lyapunov})).
    On the other hand, the weak convergence results of \cite{kuang_etal2021} require uniformly bounded fourth moments.
\end{remark}

In the small gap regime setup of Assumption \ref{assumption0}, for each arm $k$, we will use the notation $\Delta_k^{\gamma} := \max_l d_l^{\gamma} - d_k^{\gamma}$.
As $\gamma \downarrow 0$, $\Delta_k^{\gamma} \to \Delta_k := \max_l d_l - d_k$.
The essential idea behind the small gap regime is that the arm means $\mu_k^{\gamma}$ are all clustered near some fixed $\mu_* \in \mathbb{R}$, with small differences/gaps between the means on the scale of $\sqrt{\gamma}$.
In order to begin distinguishing between arms, one must play each arm on the scale of $1/\gamma$ times, so that the standard errors for estimating the means are on the scale of $\sqrt{\gamma}$, comparable in size to the gaps between the arm means.
Playing the arms significantly fewer times results in their means being essentially indistinguishable.
The conditions in Assumption \ref{assumption0} enable centered and re-scaled cumulative reward processes to be well-approximated by Brownian motions.

In the stationary rewards setting, we will work under Assumption \ref{assumption1}, given as follows.

\begin{assumption}[\normalfont Small Gap Regime with Stationary Rewards] \label{assumption1}
For each $\gamma$ and each arm $k \in [K]$, the rewards $X_k^\gamma(i)$, $i = 1,2,\dots$ are a stationary sequence with mean $\mu_k^\gamma$ (which satisfies the scaling in (\ref{assumption0_mu}) from Assumption \ref{assumption0}), with independence across different arms $k$.
For each arm $k$, there exists $\sigma_k > 0$, such that the family of processes (indexed by $\gamma$)
\begin{align}
Z^{\gamma}_k(t) & = \sqrt{\gamma} \frac{1}{\sigma_k} \sum_{i=1}^{\lfloor t/\gamma \rfloor} (X_k^{\gamma}(i) - \mu_k^{\gamma}) \label{stationary_partial_sums}
\end{align}
is tight in $D[0,\infty)$, and for any continuous function $f : \mathbb{R} \to \mathbb{R}$ vanishing at infinity, and any $u \in [0,\infty)^K$, $v > 0$,
\begin{align}
    \lim_{\gamma \downarrow 0} \E\left[ \, \abs{ \, \E\left[ f(Z_k^\gamma(u_k + v)) \mid \mathcal{F}_u^\gamma \right] - \E\left[ f(Z_k^\gamma(u_k) + \sqrt{v} \mathcal{N}) \mid Z_k^\gamma(u_k) \right] \, } \, \right] = 0, \label{nonanticipative_fdd}
\end{align}
where $\mathcal{N}$ is a standard Gaussian random variable independent of $Z_k^\gamma$, and the filtration $\mathcal{F}^\gamma = (\mathcal{F}^\gamma_u, \, u \in [0,\infty)^K)$ is defined as
\begin{align}
    \mathcal{F}_u^\gamma & = \sigma(X_k^\gamma(i_k), \, i_k \le \lfloor u_k/\gamma \rfloor, \, k \in [K] ). \label{h_u_gamma}
\end{align}
\end{assumption}

\begin{remark} \label{rmk4}
    The condition in (\ref{nonanticipative_fdd}) implies that the finite-dimensional distributions of the $Z_k^\gamma$ converge to the corresponding multivariate normal distributions.
    The condition in (\ref{nonanticipative_fdd}) also ensures weak convergence to ``non-anticipative'' (see Definition \ref{sode_non-anticipative}) solutions to the limit stochastic ODEs.
    Together, the tightness assumption and (\ref{nonanticipative_fdd}) ensure that $Z_k^\gamma \Rightarrow B_k$ in $D[0,\infty)$ as $\gamma \downarrow 0$, where $B_k$ is standard Brownian motion.
    Here, because the limit processes are Brownian motions with continuous sample paths, it suffices to verify tightness in $D[0,\infty)$ of the family $Z_k^\gamma$ in (\ref{stationary_partial_sums}) separately for each $k \in [K]$, instead of verifying tightness in $D_K[0,\infty)$ (using, e.g., Lemma \ref{diffusion_lemma5}). 
    In (\ref{stationary_partial_sums}), $\sigma_k^2$ is the long-run variance constant corresponding to the time-average of the stationary reward process; normalizing by $\sigma_k$ in (\ref{stationary_partial_sums}) ensures weak convergence to standard Brownian motion.
    If the arm rewards are not just stationary but also independent, then the conditions of Assumption \ref{assumption1} follow from those of Assumption \ref{assumption0}.
\end{remark}

\vspace{2mm}
\noindent \textbf{Key Processes for Describing Dynamics}

Here, we record the key processes that will be used throughout the paper for describing the dynamics of TS and related algorithms in the small gap regime. 
To keep track of the amount of sampling effort allocated to each of the $K$ arms, we use $U^{\gamma} = (U^{\gamma}_k, \, k \in [K]) \in D_K[0,\infty)$, defined as:
\begin{align}
U^{\gamma}_k(t) = \gamma \sum_{i=1}^{\lfloor t/\gamma \rfloor} I_k^{\gamma}(i), \label{diffusion_r_process1} 
\end{align}
which is a re-scaling of (\ref{n_k}).
To keep track of the rewards received for each arm under the random table model, we will use $S^{\gamma} = (S^{\gamma}_k, \, k \in [K]) \in D_K[0,\infty)$, defined as:
\begin{align}
S^{\gamma}_k(t) = \sqrt{\gamma} \frac{1}{\sigma_k} \sum_{i=1}^{\lfloor t/\gamma \rfloor} I_k^{\gamma}(i) (X_k^{\gamma}(i) - \mu_k^{\gamma}), \label{diffusion_s_process1}
\end{align}
which is a centering and re-scaling of (\ref{g_k}).
To keep track of the rewards received for each arm under the reward stack model, we will use $Z^{\gamma} \circ U^{\gamma} = (Z_k^{\gamma}(U_k^{\gamma}), \, k \in [K]) \in D_K[0,\infty)$, where
\begin{align}
Z^{\gamma}_k(U^{\gamma}_k(t)) & = \sqrt{\gamma} \frac{1}{\sigma_k} \sum_{i=1}^{U^{\gamma}_k(t)/\gamma} (X_k^{\gamma}(i) - \mu_k^{\gamma}), \label{diffusion_z_process1}
\end{align}
with $Z^{\gamma} = (Z^{\gamma}_k, \, k \in [K])$ and each $Z_k^{\gamma}$ as defined in (\ref{stationary_partial_sums}), and with $U_k^\gamma$ as defined in (\ref{diffusion_r_process1}).
The processes in (\ref{diffusion_z_process1}), like those in (\ref{diffusion_s_process1}), are also a centering and re-scaling of (\ref{g_k}).
(For vector-valued functions $f$ and $g$, we use $f \circ g$ to denote component-wise composition of $f$ and $g$.)
In (\ref{diffusion_s_process1}) and (\ref{diffusion_z_process1}), the $\mu_k^\gamma$ and $\sigma_k$ are the means and scaling factors from either Assumption \ref{assumption0} or Assumption \ref{assumption1}.

\section{Derivations of Diffusion Approximations} \label{diffusion_mab}

In the following sections, we derive an SDE approximation (Section \ref{diffusion_mab_1}) and a stochastic ODE approximation (Section \ref{diffusion_mab_2}) for the Gaussian Thompson sampler, i.e., TS implemented using posterior updating based on Gaussian likelihoods and priors.
For the Gaussian likelihood, we use a fixed variance $c_*^2 > 0$, which may or may not correspond to the $\sigma_k^2$, the limit variances (in (\ref{assumption0_sigma})) or long-run variances (in (\ref{stationary_partial_sums})), of the arm reward processes.
Later in the paper, we will complement the theory of this section by studying EF Thompson samplers in Section \ref{diffusion_general}, the non-parametric bootstrap sampler in Section \ref{diffusion_bootstrap}, and then model mis-specification issues in Section \ref{diffusion_mis-specification}.

Before continuing on to the derivation of diffusion approximations, we first discuss a technical issue that can arise.
The sampling behavior of TS can be highly erratic at the very beginning of a bandit experiment in the small gap regime (as in Assumptions \ref{assumption0} and \ref{assumption1}) when little data has been collected and the algorithm is only performing exploration.
This can create mathematical difficulties such as the breakdown of Lipschitz continuity in SDE approximations in an arbitrarily small initial time interval (in continuous time), which in turn makes it challenging to establish that the SDEs (and stochastic ODEs) have unique solutions.
Below, we discuss two ways of ``smoothing'' the initial behavior of TS to restore suitable Lipschitz continuity of the SDEs.

\vspace{2mm}
\noindent \textbf{1) Smoothing via Concentrated Priors}

One way to smooth out the initial behavior of TS is to use concentrated priors on the arm means.
We use this approach in Sections \ref{diffusion_mab_1} and \ref{diffusion_mab_2}.
In Assumptions \ref{assumption0} and \ref{assumption1}, the arm means $\mu_k^\gamma$ are concentrated around $\mu_*$ with sub-optimality gaps $\sqrt{\gamma} \Delta_k^{\gamma}$, where the $\Delta_k^{\gamma}$ are unknown.
(Recall that $\Delta_k^{\gamma} \to \Delta_k$ for some $\Delta_k \ge 0$ as $\gamma \downarrow 0$.)
We assume that the centering parameter $\mu_*$ and scale parameter $\gamma$ are known, and we use an independent $N(\mu_*,\gamma/b)$ prior for each arm in the Gaussian Thompson sampler, with a chosen fixed $b > 0$.
Translated into practice, this means that the experimenter knows, perhaps from related experiments run in the past, that the arm means are in a ``small ($\sqrt{\gamma}$-scale) neighborhood'' of some $\mu_*$.
So, the experimenter chooses priors with probability mass concentrated in such a neighborhood of $\mu_*$.
Then, over time horizons scaling as $1/\gamma$, the bandit algorithm begins to identify the sub-optimality gaps and maximize cumulative reward.

Importantly, the use of $\gamma$-scale variance priors together with $(1/\gamma)$-scale time horizons ensures the SDE approximations have suitable Lipschitz continuity properties, and thus a unique strong solution.
The use of $\gamma$-scale variance priors together with data collected over $(1/\gamma)$-scale time horizons naturally enables Bayesian inference about the $\sqrt{\gamma}$-scale sub-optimality gaps.
If the prior is less concentrated with variance scaling as $\omega(\gamma)$ as $\gamma \downarrow 0$, then it will be asymptotically dominated by the data collected over $(1/\gamma)$-scale time horizons.
And if the prior is more concentrated with variance scaling as $o(\gamma)$ as $\gamma \downarrow 0$, then it will asymptotically dominate the data collected.

\begin{remark} \label{not_translation_invariant}
    The Gaussian Thompson sampler with concentrated priors on the arm means is not translation-invariant. 
    To avoid writing the centering parameter $\mu_*$ repeatedly in the derivations for Theorems \ref{diffusion_thm1} and \ref{diffusion_thm2} only, there we simply set $\mu_* = 0$.
    For all other results, there is translation invariance, and $\mu_*$ always gets canceled out.
\end{remark}

\vspace{2mm}
\noindent \textbf{2) Smoothing via $\epsilon$-warm-start}

A second way to smooth out the initial erratic behavior of TS is to sample all arms with fixed, positive probabilities for an arbitrarily small initial time interval $\epsilon$ in continuous time, and then run TS afterwards.
We refer to this initialization procedure as \textit{$\epsilon$-warm-start}, which is defined below at a general level.
We will use it in Section \ref{diffusion_additional} and in Section \ref{diffusion_applications}.
\begin{definition}[$\epsilon$-warm-start] \label{def:epsilon-warm-start}
    Fix some positive (user-specified) probabilities $q_k$, $k \in [K]$, with $\sum_k q_k = 1$.
    For the initial $\lfloor \epsilon/\gamma \rfloor$ time periods, sample each arm $k$ with probability $q_k$.
    Then, run the desired bandit algorithm from time $\lfloor \epsilon/\gamma \rfloor + 1$ onward.
\end{definition}
Using $\epsilon$-warm-start, we can ensure suitable Lipschitz continuity of the SDE approximation, and thus a unique strong solution.
Moreover, the prior used in TS can be general and need not be concentrated in any way.
When the desired algorithm is TS, we can also think of $\epsilon$-warm-start as an empirical Bayes approach, where a tiny fraction of data is collected initially to learn a prior centered around $\mu_*$ with variance scaling as $\gamma$, after which TS using the learned prior is deployed.

\subsection{SDE Approximation} \label{diffusion_mab_1}

In this section, we work under Assumption \ref{assumption0} with iid rewards for each arm, and we use the random table model of reward feedback, as introduced in Section \ref{diffusion_mab_0}.
This leads to the SDE approximation for the Gaussian Thompson sampler in Theorem \ref{diffusion_thm1} below.

To begin, we show that the dynamics in this setting can be described by the evolution of the processes $(U^\gamma,S^\gamma)$ as defined in (\ref{diffusion_r_process1}) and (\ref{diffusion_s_process1}).
At time $j+1$, conditional on $\mathcal{H}^{\gamma}_{j}$ (defined in (\ref{h_j_gamma})), the Gaussian Thompson sampler draws a sample from the posterior distribution of each arm $k$:
\begin{align}
\widetilde{\mu}_k^{\gamma}(j+1) \sim N\left(\frac{\gamma \sum_{i=1}^{j} I_k^{\gamma}(i) X_k^{\gamma}(i)}{U^{\gamma}_k(j \gamma) + b c_*^2}, \frac{c_*^2 \gamma}{U^{\gamma}_k(j \gamma) + b c_*^2}\right). \label{diffusion_draw1}
\end{align}
So, the probability of playing arm $k$ can be expressed as:
\begin{align}
& \P\left( k = \argmax_{l \in [K]} \widetilde{\mu}_l^{\gamma}(j+1) \; \bigl| \; \mathcal{H}^{\gamma}_{j} \right) \label{diffusion_prob1} \\
& = \P\left( k = \argmax_{l \in [K]} \left\{\frac{S^{\gamma}_l(j \gamma) \sigma_l + U^{\gamma}_l(j \gamma) d_l^{\gamma}}{U^{\gamma}_l(j \gamma) + b c_*^2} + \frac{c_*}{\sqrt{U^{\gamma}_l(j \gamma) + b c_*^2}} \mathcal{N}_l\right\} \; \Biggl| \; U^{\gamma}(j \gamma),S^{\gamma}(j \gamma) \right) \label{diffusion_prob2} \\
& = p_k^{\gamma}(U^{\gamma}(j \gamma),S^{\gamma}(j \gamma)), \label{diffusion_prob3}
\end{align}
where the $\mathcal{N}_l$ are independent standard Gaussian random variables, and for $u = (u_k, \, k \in [K]) \in [0,\infty)^K$ and $s = (s_k, \, k \in [K]) \in \mathbb{R}^K$,
\begin{align}
    p^{\gamma}_k(u,s) = \P\left( k = \argmax_{l \in [K]} \left\{\frac{s_{l} \sigma_{l} + u_{l} d^{\gamma}_{l}}{u_{l} + b c_*^2} + \frac{c_*}{\sqrt{u_{l} + b c_*^2}}\mathcal{N}_l \right\} \right). \label{prelimit_sampling_probability}
\end{align}
We can then re-express $U^{\gamma}_k(t)$ and $S^{\gamma}_k(t)$ from (\ref{diffusion_r_process1})-(\ref{diffusion_s_process1}) as
\begin{align}
U^{\gamma}_k(t) & = \gamma \sum_{i=0}^{\lfloor t/\gamma \rfloor - 1} p_k^{\gamma}(U^{\gamma}(i \gamma),S^{\gamma}(i \gamma)) + M^{\gamma}_k(t) \label{diffusion_discrete_sde1} \\
S^{\gamma}_k(t) & = \sum_{i=0}^{\lfloor t/\gamma \rfloor - 1} \sqrt{\displaystyle p_k^{\gamma}(U^{\gamma}(i \gamma),S^{\gamma}(i \gamma))} \left( B^{\gamma}_k((i+1) \gamma) - B^{\gamma}_k(i \gamma) \right), \label{diffusion_discrete_sde2}
\end{align}
where $M^{\gamma} = (M_k^{\gamma}, \, k \in [K]) \in D_K[0,\infty)$ and $B^{\gamma} = (B_k^{\gamma}, \, k \in [K]) \in D_K[0,\infty)$ are defined as:
\begin{align}
M^{\gamma}_k(t) & = \gamma \sum_{i=0}^{\lfloor t/\gamma \rfloor - 1} \left( I_k^{\gamma}(i+1) - p_k^{\gamma}(U^{\gamma}(i \gamma),S^{\gamma}(i \gamma)) \right) \label{diffusion_discrete_sde4} \\
B^{\gamma}_k(t) & = \sqrt{\gamma} \frac{1}{\sigma_k} \sum_{i=0}^{\lfloor t/\gamma \rfloor - 1} \frac{I_k^{\gamma}(i+1) (X_k^{\gamma}(i+1) - \mu_k^{\gamma})}{\sqrt{\displaystyle p_k^{\gamma}(U^{\gamma}(i \gamma),S^{\gamma}(i \gamma))}}, \label{diffusion_discrete_sde5}
\end{align}
and $(I_k^{\gamma}(i+1), \, k \in [K])$ is a multinomial random variable with a single trial and success probabilities $p_k^{\gamma}(\displaystyle U^{\gamma}(i \gamma),S^{\gamma}(i \gamma))$.

As $\gamma \downarrow 0$, we show that $M^{\gamma}$ converges weakly to the $D_K[0,\infty)$ zero process, and $B^{\gamma}$ converges weakly to standard $K$-dimensional Brownian motion.
Additionally, since $d_k^{\gamma} \to d_k$ from (\ref{assumption0_mu}),
we have 
\begin{align}
    p_k^{\gamma}(u,s) \to p_k(u,s) \label{diffusion_uniform_compact}
\end{align}
uniformly for $(u,s)$ in compact subsets of $[0,\infty)^K \times \mathbb{R}^K$, where
\begin{align}
    p_k(u,s) = \P\left( k = \argmax_{l \in [K]} \left\{\frac{s_{l} \sigma_{l} + u_{l} d_{l}}{u_{l} + b c_*^2} + \frac{c_*}{\sqrt{u_{l} + b c_*^2}}\mathcal{N}_l \right\} \right). \label{limit_sampling_probability}
\end{align}
Thus, we expect (\ref{diffusion_discrete_sde1})-(\ref{diffusion_discrete_sde2}) to be a discrete approximation to the SDE in integral form:
\begin{align}
U_k(t) & = \int_0^t p_k(U(v),S(v)) dv \label{diffusion_sde_rep1} \\
S_k(t) & = \int_0^t \sqrt{p_k(U(v),S(v))} dB_k(v), \quad k \in [K] \label{diffusion_sde_rep2}
\end{align}
with standard $K$-dimensional Brownian motion $B$.

To conclude the above derivation, the formal SDE characterization is stated in Theorem \ref{diffusion_thm1} below.
The proof of Theorem \ref{diffusion_thm1} can be found in Section \ref{diffusion_sde}, along with the development of the supporting results for the proof.
The rigorous argument closely follows the derivation above.
The main technical tool is the CMT, together with the property that stochastic integration is a continuous mapping of the integrand and integrator processes, which allows us to pass from the pre-limit in (\ref{diffusion_discrete_sde1})-(\ref{diffusion_discrete_sde2}) to the limit in (\ref{diffusion_sde_rep1})-(\ref{diffusion_sde_rep2}).

As mentioned previously, for each function $p_k$ in (\ref{limit_sampling_probability}), both $p_k$ and $\sqrt{p_k}$ are \textit{locally Lipschitz continuous}, which means that they are Lipschitz continuous on any compact subset of the domain $[0,\infty)^K \times \mathbb{R}^K$.
In particular, in (\ref{limit_sampling_probability}), we have $b > 0$ resulting from the use of concentrated priors in the pre-limit, which ensures local Lipschitz continuity.
These functions are also bounded, which together with local Lipschitz continuity, ensures that the SDEs in (\ref{diffusion_sde_rep1})-(\ref{diffusion_sde_rep2}) have a (global) unique strong solution, as summarized in Remark \ref{sde_strong_solution_existence_uniqueness} below.
\begin{remark} \label{sde_strong_solution_existence_uniqueness}
    For the definition of a strong solution to an SDE and uniqueness of strong solutions, see Definitions 2.1 and 2.3 from Chapter 5.2 of \cite{karatzas_etal1998}. 
    The existence of a strong solution under boundedness (and thus (sub-)linear growth) and local Lipschitz continuity of the $p_k$ and $\sqrt{p_k}$ functions follows from Theorem 3.11 from Chapter 5 of \cite{ethier_etal1986}.
    Uniqueness follows from Theorem 2.5 from Chapter 5.2 of \cite{karatzas_etal1998}.
\end{remark}

Before stating Theorem \ref{diffusion_thm1}, in Remark \ref{rmk0} below, we note an expression for regret that will be used throughout the rest of the paper.
\begin{remark} \label{rmk0}
    Under the setup of Assumption \ref{assumption0}, for a particular $\gamma$ value, the overall regret $\text{Reg}^{\gamma}(n)$ at time $n$ is related to the $U^{\gamma}_k$ processes by:
    \begin{align}
        \text{Reg}^{\gamma}(n) = \frac{1}{\sqrt{\gamma}} \sum_{k \in [K]} U^{\gamma}_k(n \gamma) \Delta_k^{\gamma}. \label{diffusion_regret_connection}
    \end{align}
\end{remark}

\begin{theorem} \label{diffusion_thm1}
Consider a $K$-armed bandit in the small gap regime of Assumption \ref{assumption0} (with iid rewards for each arm) and the random table model of reward feedback.
For the Gaussian Thompson sampler with concentrated priors on the arm means, we have
\begin{align}
    (U^{\gamma},S^{\gamma}) \Rightarrow (U,S) \label{diffusion_sde0}
\end{align}
as $\gamma \downarrow 0$ in $D_{2K}[0,\infty)$, where $(U,S)$ is the unique strong solution to the SDE:
\begin{align}
dU_k(t) & = p_k(U(t),S(t)) dt \label{diffusion_sde1} \\
dS_k(t) & = \sqrt{p_k(U(t),S(t))} dB_k(t) \label{diffusion_sde2} \\
U_k(0) & = S_k(0) = 0, \quad k \in [K], \label{diffusion_sde3}
\end{align}
with standard $K$-dimensional Brownian motion $B$ and functions $p_k$ as expressed in (\ref{limit_sampling_probability}).

Moreover, for regret,
\begin{align}
    \sqrt{\gamma} \, \textup{Reg}^\gamma(\lfloor \cdot / \gamma \rfloor) \Rightarrow \sum_{k \in [K]} U_k(\cdot) \Delta_k \label{regret_weak_convergence}
\end{align}
as $\gamma \downarrow 0$ in $D[0,\infty)$.
\end{theorem}

\subsection{Stochastic ODE Approximation} \label{diffusion_mab_2}

In this section, we work under more general conditions than in Section \ref{diffusion_mab_1}, where we used Assumption \ref{assumption0} with iid rewards for each arm. 
Here, we work under Assumption \ref{assumption1} with general stationary sequences of rewards for each arm, and we use the reward stack model of reward feedback, as introduced in Section \ref{diffusion_mab_0}.
As discussed in Section \ref{diffusion_mab_0}, we can think of this setup as a rested bandit, where the rewards for each arm evolve according to a stochastic process when the arm is played and stays frozen otherwise.
This leads to the stochastic ODE approximation for the Gaussian Thompson sampler in Theorem \ref{diffusion_thm2}.
To conclude this section, via Theorem \ref{diffusion_thm_sde_sode_equivalence} and Proposition \ref{time_change_representation_conversion}, we establish the distributional equivalence between general SDE and stochastic ODE limit representations.

Similar to the derivation of the SDE approximation, we first show that the dynamics can be described by the evolution of the processes $(U^\gamma,Z^\gamma \circ U^\gamma)$ as defined in (\ref{diffusion_r_process1}) and (\ref{diffusion_z_process1}).
At time $j+1$, conditional on $\mathcal{H}^{\gamma}_{j}$ (defined in (\ref{h_j_gamma})), the Gaussian Thompson sampler draws a sample from the posterior distribution of each arm $k$:
\begin{align}
\widetilde{\mu}_k^{\gamma}(j+1) \sim N\left(\frac{\gamma \sum_{i=1}^{U^{\gamma}_k(j \gamma)/\gamma} X_k^{\gamma}(i)}{U^{\gamma}_k(j \gamma) + b c_*^2}, \frac{c_*^2 \gamma}{U^{\gamma}_k(j \gamma) + b c_*^2}\right). \label{diffusion_draw2}
\end{align}
So, the probability of playing arm $k$ can be expressed as:
\begin{align}
& \P\left( k = \argmax_{l \in [K]} \widetilde{\mu}_l^{\gamma}(j+1) \; \bigl| \; \mathcal{H}^{\gamma}_{j} \right) \label{diffusion_prob11} \\
& = \P\left( k = \argmax_{l \in [K]} \left\{\frac{Z^{\gamma}_l(U^{\gamma}_l(j \gamma)) \sigma_l + U^{\gamma}_l(j \gamma) d_l^{\gamma}}{U^{\gamma}_l(j \gamma) + b c_*^2} + \frac{c_*}{\sqrt{U^{\gamma}_l(j \gamma) + b c_*^2}} \mathcal{N}_l\right\} \; \Biggl| \; U^{\gamma}(j \gamma),Z^{\gamma} \circ U^{\gamma}(j \gamma) \right) \label{diffusion_prob12} \\
& = p_k^{\gamma}(U^{\gamma}(j \gamma),Z^{\gamma} \circ U^{\gamma}(j \gamma)), \label{diffusion_prob13}
\end{align}
where the $\mathcal{N}_l$ are independent standard Gaussian random variables, and functions $p_k^{\gamma}$ are given by (\ref{prelimit_sampling_probability}).

We can then re-express $U_k^{\gamma}(t)$ as
\begin{align}
U^{\gamma}_k(t) & = \gamma \sum_{i=0}^{\lfloor t/\gamma \rfloor - 1} p_k^{\gamma}(U^{\gamma}(i \gamma),Z^{\gamma} \circ U^{\gamma}(i \gamma)) + M^{\gamma}_k(t), \quad k \in [K], \label{diffusion_discrete_ode1}
\end{align}
where $M^{\gamma} = (M_k^{\gamma}, \, k \in [K]) \in D_K[0,\infty)$ is defined as:
\begin{align}
M^{\gamma}_k(t) & = \gamma \sum_{i=0}^{\lfloor t/\gamma \rfloor - 1} \left( I^{\gamma}_k(i+1) - p_k^{\gamma}(U^{\gamma}(i \gamma),Z^{\gamma} \circ U^{\gamma}(i \gamma)) \right), \label{diffusion_discrete_ode3}
\end{align}
and $(I^{\gamma}_k(i+1), \, k \in [K])$ is a multinomial random variable with a single trial and success probabilities $p_k^{\gamma}(U^{\gamma}(i \gamma),Z^{\gamma} \circ U^{\gamma}(i \gamma))$.

As $\gamma \downarrow 0$, we show that $M^{\gamma}$ converges weakly to the $D_K[0,\infty)$ zero process.
Moreover, as discussed in Remark \ref{rmk4}, $Z^{\gamma}$ converges weakly to standard $K$-dimensional Brownian motion.
As in the previous section, the convergence in (\ref{diffusion_uniform_compact}) holds.
Thus, we expect (\ref{diffusion_discrete_ode1}) to be a discrete approximation to the stochastic ODE in integral form:
\begin{align}
U_k(t) = \int_0^t p_k(U(v),B \circ U(v)) dv, \quad k \in [K], \label{diffusion_ode_rep1}
\end{align}
with standard $K$-dimensional Brownian motion $B$, and functions $p_k$ as expressed in (\ref{limit_sampling_probability}).

To conclude the above derivation, the formal stochastic ODE characterization is stated in Theorem \ref{diffusion_thm2} below.
The proof of Theorem \ref{diffusion_thm2} can be found in Section \ref{diffusion_rode}.
The rigorous argument closely follows the derivation above, using the CMT, together with the property that Riemann integration is a continuous mapping of the integrand and integrator processes, which allows us to pass from the pre-limit in (\ref{diffusion_discrete_ode1}) to the limit in (\ref{diffusion_ode_rep1}).

The uniqueness of non-anticipative weak (in distribution) solutions to the stochastic ODE in Theorem \ref{diffusion_thm2} follows from Theorem \ref{diffusion_thm_sde_sode_equivalence} below (which establishes a general connection between SDE and stochastic ODE solutions), together with standard SDE strong uniqueness theory (as discussed previously in Remark \ref{sde_strong_solution_existence_uniqueness}).
Before stating Theorem \ref{diffusion_thm2}, we specify below in Definition \ref{sode_non-anticipative} what it means to be a non-anticipative solution to the stochastic ODE.
(Definition \ref{sode_non-anticipative} uses the notions of filtrations indexed by directed sets and their associated stopping times, which are discussed below in Remark \ref{kurtz_results} and Definition \ref{directed_set_filtration_stopping_time}.)
Then, in Definition \ref{sode_weak_uniqueness}, we specify what it means to be a unique (in distribution) weak solution to the stochastic ODE.
Definitions \ref{sode_non-anticipative} and \ref{sode_weak_uniqueness} can be found in Chapter 6, Section 2 of \cite{ethier_etal1986}.
Also, one may wonder if the non-anticipative property automatically holds for all solutions to stochastic ODEs of the form in (\ref{diffusion_ode_rep1})---in general, the answer is no; see Remark \ref{necessity_non-anticipative} below.

\begin{definition} \label{sode_non-anticipative}
    Let $B$ be a standard $K$-dimensional Brownian motion on a probability space $(\Omega, \mathbb{F}, \mathbb{P})$, and consider the augmented filtration $\mathcal{F} = (\mathcal{F}_u, \, u \in [0,\infty)^K)$, with
    \begin{align}
        \mathcal{F}_u = \sigma\bigl( B_k(t_k), \, t_k \le u_k, \, k \in [K]\bigr) \vee \sigma(\mathcal{L}), \label{explicit_augmented_Brownian_filtration}
    \end{align}
    where $\mathcal{L} \subset \mathbb{F}$ is the collection of all $\mathbb{P}$-probability zero sets.
    For the Brownian motion $B$ and Borel measurable functions $p_k : [0,\infty)^K \times \mathbb{R}^K \to (0,1)$, $k \in [K]$, we say that a solution $U = (U_k, \, k \in [K])$ to the stochastic ODE (satisfying a.s. (almost surely) for all $t \ge 0$):
    \begin{align}
        U_k(t) = \int_0^t p_k(U(v),B \circ U(v)) dv, \quad k \in [K], \nonumber
    \end{align}
    is \textit{non-anticipative} if there exists a filtration $\mathcal{G} = (\mathcal{G}_u, \, u \in [0,\infty)^K)$ such that the following hold, with $\xi^u(\cdot) := (B_k(u_k + \cdot), \, k \in [K])$ for $u \in [0,\infty)^K$. \vspace{1mm}
    \begin{enumerate}
        \item[(i)] $\mathcal{F}_u \subset \mathcal{G}_u \subset \mathbb{F}$ \, $\forall \, u \in [0,\infty)^K$ \vspace{1mm}
        \item[(ii)] $\P(\xi^u \in \cdot \mid \mathcal{G}_u) = \P(\xi^u \in \cdot \mid \mathcal{F}_u)$ \, $\forall \, u \in [0,\infty)^K$ \vspace{1mm}
        \item[(iii)] for each $t \ge 0$, $U(t)$ is a $\mathcal{G}_u$-stopping time
    \end{enumerate} 
\end{definition}

\begin{remark} \label{kurtz_results}
    In (\ref{explicit_augmented_Brownian_filtration}), each Brownian motion $B_k$ has its own separate clock/time index $u_k$, and the filtration $\mathcal{F}$ is indexed by the directed set $[0,\infty)^K$.
    The definitions of filtrations indexed by directed sets like $[0,\infty)^K$ and their associated stopping times are provided in the following Definition \ref{directed_set_filtration_stopping_time}.
    For further details, see \cite{kurtz_etal1980a} or Chapter 2, Section 8 of \cite{ethier_etal1986}.
\end{remark}

\begin{definition} \label{directed_set_filtration_stopping_time}
    Let $(\Omega, \mathbb{F}, \mathbb{P})$ be a probability space.
    A collection of sub-$\sigma$-algebras $\mathcal{F} = (\mathcal{F}_u, \, u \in [0,\infty)^m)$ of $\mathbb{F}$ is a \textit{filtration} indexed by (directed set) $[0,\infty)^m$ if $u_k \le v_k$ for all $k \in [m]$ implies that $\mathcal{F}_u \subset \mathcal{F}_v$.
    A random variable $\tau$ taking values in $[0,\infty)^m$ is an \textit{$\mathcal{F}$-stopping time} if $\bigcap_{k \in [m]} \{ \tau_k \le u_k \} \in \mathcal{F}_u$ for all $u \in [0,\infty)^m$.
\end{definition}

\begin{definition} \label{sode_weak_uniqueness}
    Given a standard $K$-dimensional Brownian motion $B$ on a probability space $(\Omega, \mathbb{F}, \mathbb{P})$, we say that the stochastic ODE (in integral form) in (\ref{diffusion_ode_rep1}) has a \textit{weak solution} if there exists a probability space $(\widetilde{\Omega},\widetilde{\mathbb{F}},\widetilde{\mathbb{P}})$ on which is defined a standard $K$-dimensional Brownian motion $\widetilde{B}$ and a stochastic process $\widetilde{U}$ satisfying $\widetilde{\mathbb{P}}$-a.s. for all $t \ge 0$:
    \begin{align}
        \widetilde{U}_k(t) = \int_0^t p_k(\widetilde{U}(v),\widetilde{B} \circ \widetilde{U}(v)) dv, \quad k \in [K]. \nonumber
    \end{align}
    We say that a (non-anticipative) weak solution is \textit{unique in distribution} if any two (non-anticipative) weak solutions always have the same finite-dimensional distributions.
\end{definition}

\begin{remark} \label{sode_existence}
    As mentioned earlier, we will use the connection between stochastic ODEs and SDEs (in Theorem \ref{diffusion_thm_sde_sode_equivalence} below) to establish uniqueness (in distribution) of non-anticipative weak solutions to stochastic ODEs.
    However, tightness alone is sufficient to guarantee \textit{existence} of non-anticipative weak solutions to stochastic ODEs of the form in (\ref{diffusion_ode_rep1}), as long as the functions $p_k$ are continuous.
    See Corollary 3.6 from Chapter 6, Section 3 of \cite{ethier_etal1986}.
\end{remark}

\begin{remark} \label{necessity_non-anticipative}
    Given a standard $K$-dimensional Brownian motion $B$ on a probability space $(\Omega, \mathbb{F}, \mathbb{P})$, if a solution $U$ to the stochastic ODE in (\ref{diffusion_ode_rep1}) is not unique for $\mathbb{P}$-almost all sample paths of $B$, then it may not be non-anticipative.
    Thus, if we have uniqueness in distribution but not $\mathbb{P}$-a.s., then the non-anticipative property requires separate treatment.
    See Remark 2.3 and Problem 1 from Chapter 6 of \cite{ethier_etal1986} for an example.
\end{remark}

\begin{theorem} \label{diffusion_thm2}
Consider a $K$-armed bandit in the small gap regime of Assumption \ref{assumption1} (with stationary rewards for each arm) and the reward stack model of reward feedback.
For the Gaussian Thompson sampler with concentrated priors on the arm means, we have
\begin{align}
    (U^{\gamma},Z^{\gamma} \circ U^{\gamma}) \Rightarrow (U,B \circ U) \label{diffusion_ode0}
\end{align}
as $\gamma \downarrow 0$ in $D_{2K}[0,\infty)$, where $U$ is the unique (in distribution) non-anticipative weak solution to the stochastic ODE:
\begin{align}
dU_k(t) & = p_k(U(t),B \circ U(t)) dt \label{diffusion_ode1} \\
U_k(0) & = 0, \quad k \in [K], \label{diffusion_ode2}
\end{align}
with standard $K$-dimensional Brownian motion $B$ and functions $p_k$ as expressed in (\ref{limit_sampling_probability}).

Moreover, for regret, (\ref{regret_weak_convergence}) holds in this stochastic ODE setting.
\end{theorem}

\begin{remark} \label{rmk1}
In the special case that the rewards for each arm are iid (not just stationary), then the current setup (Assumption \ref{assumption1} with the reward stack model) leading to the stochastic ODE representation in Theorem \ref{diffusion_thm2} is probabilistically equivalent to the setup used in Section \ref{diffusion_mab_1} (Assumption \ref{assumption0} with the random table model) leading to the SDE representation in Theorem \ref{diffusion_thm1}.
In particular, under iid rewards, the processes $S^{\gamma}_k(t)$ (defined in (\ref{diffusion_s_process1})) and the processes $Z^{\gamma}_k(U^{\gamma}_k(t))$ (defined in (\ref{diffusion_z_process1})) have the same distribution. 
The processes $U^{\gamma}_k(t)$ are also defined in exactly the same way in both cases.
Thus, under iid rewards, the weak limits, i.e., the unique strong solution to the SDE in Theorem \ref{diffusion_thm1} and the unique (in distribution) weak solution to the stochastic ODE in Theorem \ref{diffusion_thm2}, must also have the same distribution.
\end{remark}

In Theorem \ref{diffusion_thm_sde_sode_equivalence} below, we work with the limit processes and establish in general that the uniqueness of the strong solution to an SDE implies the uniqueness (in distribution) of non-anticipative weak solutions to the corresponding stochastic ODE.
As mentioned earlier, the solution uniqueness to the stochastic ODE in Theorem \ref{diffusion_thm2} is obtained from Theorem \ref{diffusion_thm_sde_sode_equivalence}.
The proof of Theorem \ref{diffusion_thm_sde_sode_equivalence} is provided in Section \ref{diffusion_rode}.

\begin{theorem} \label{diffusion_thm_sde_sode_equivalence}
Suppose $U$ is a non-anticipative solution to the stochastic ODE:
\begin{align}
dU_k(t) & = p_k(U(t), B \circ U(t)) dt \label{sde_sode_equivalence_4} \\
U_k(0) & = 0, \quad k \in [K], \label{sde_sode_equivalence_5}
\end{align}
with standard $K$-dimensional Brownian motion $B$ and Borel measurable functions $p_k : [0,\infty)^K \times \mathbb{R}^K \to (0,1)$.
Then, there exists a standard $K$-dimensional Brownian motion $\widetilde{B}$ such that $(U,S)$, with $S = B \circ U$, is a solution to the SDE:
\begin{align}
dU_k(t) & = p_k(U(t),S(t)) dt \label{sde_sode_equivalence_1} \\
dS_k(t) & = \sqrt{p_k(U(t),S(t))} d\widetilde{B}_k(t) \label{sde_sode_equivalence_2} \\
U_k(0) & = S_k(0) = 0, \quad k \in [K]. \label{sde_sode_equivalence_3}
\end{align}
Therefore, if the SDE in (\ref{sde_sode_equivalence_1})-(\ref{sde_sode_equivalence_3}) has a unique strong solution $(U^*,S^*)$, then for any non-anticipative solution $U$ to the stochastic ODE in (\ref{sde_sode_equivalence_4})-(\ref{sde_sode_equivalence_5}), we have $(U, B \circ U) \overset{d}{=} (U^*, S^*)$.
\end{theorem}

Conversely, in Proposition \ref{time_change_representation_conversion}, we show that we can always convert from the SDE representation to the stochastic ODE representation.
This follows directly from a multivariate version of the well-known result that a continuous local martingale, such as a stochastic integral, can be represented as a Brownian motion with a random time change. 
The proof of Proposition \ref{time_change_representation_conversion} is provided in Section \ref{diffusion_rode}.

\begin{proposition} \label{time_change_representation_conversion}
    Let $(U,S)$ be a strong solution to the SDE in (\ref{sde_sode_equivalence_1})-(\ref{sde_sode_equivalence_3}), with independent standard $K$-dimensional Brownian motion $\widetilde{B}$ and Borel measurable functions $p_k : [0,\infty)^K \times \mathbb{R}^K \to (0,1)$.
    Then, there exists a standard $K$-dimensional Brownian motion $B$ such that we have the representation $(U,S) = (U, B \circ U)$, which satisfies the stochastic ODE in (\ref{sde_sode_equivalence_4})-(\ref{sde_sode_equivalence_5}) with $U$ as a non-anticipative solution.
\end{proposition}

\subsection{Approximations Without Concentrated Priors} \label{diffusion_additional}

From the development of Theorems \ref{diffusion_thm1} and \ref{diffusion_thm2}, it is important for each function $p_k$ in (\ref{limit_sampling_probability}) that both $p_k$ and $\sqrt{p_k}$ are locally Lipschitz continuous, which together with boundedness, ensures that the limit SDE has a unique strong solution (as discussed in Remark \ref{sde_strong_solution_existence_uniqueness}) and the limit stochastic ODE has a unique (in distribution) non-anticipative weak solution.
In Corollary \ref{diffusion_prop00}, we state a result for general sampling-based bandit algorithms that does not involve locally Lipschitz continuous $p_k$ and $\sqrt{p_k}$ limit functions.
In such settings, the uniqueness theory we previously invoked no longer applies.
Nevertheless, the rescaled pre-limit processes, e.g., $(U^\gamma,Z^\gamma \circ U^\gamma)$ in the stochastic ODE setting, will still be tight.
So, every subsequence as $\gamma \downarrow 0$ of pre-limit processes will have a further subsequence that converges weakly to a limit process that satisfies the stochastic ODE.
However, these weak limit processes may be distinct in distribution, so we simply characterize their evolution equations.
The justification for Corollary \ref{diffusion_prop00} follows directly from the proof of Theorem \ref{diffusion_thm2}.

\begin{corollary} \label{diffusion_prop00}
Consider a $K$-armed bandit in the small gap regime of Assumption \ref{assumption1} (with stationary rewards for each arm) and the reward stack model of reward feedback.
For a sampling-based algorithm, suppose that as $\gamma \downarrow 0$, the sampling probabilities $p_k^{\gamma}(u,s) \to p_k(u,s)$ uniformly for $(u,s)$ in compact subsets of $[0,\infty)^K \times \mathbb{R}^K$, where $p_k$ is a continuous function.
Then, the weak limit points of $(U^{\gamma}, Z^{\gamma} \circ U^{\gamma})$ in $D_{2K}[0,\infty)$ as $\gamma \downarrow 0$ are of the form $(U, B \circ U)$ and satisfy the stochastic ODE:
\begin{align}
dU_k(t) & = p_k(U(t),B \circ U(t)) dt \nonumber \\
U_k(0) & = 0, \quad k \in [K], \nonumber
\end{align}
with standard $K$-dimensional Brownian motion $B$.
\end{corollary}

Next, we develop a diffusion approximation for the Gaussian Thompson sampler without use of concentrated priors with variance scaling as $\gamma$.
Unlike in Sections \ref{diffusion_mab_1}-\ref{diffusion_mab_2}, here the decision-maker can use any fixed Gaussian prior with no $\gamma$-dependence.
(More generally, any prior can be used as long as it puts positive probability mass in a neighborhood of the centering parameter $\mu_*$ from Assumptions \ref{assumption0} and \ref{assumption1}.)
Then, the functions $p_k$ in (\ref{limit_sampling_probability}) become:
\begin{align}
    p_k(u,s) = \P\left( k = \argmax_{l \in [K]} \left\{ \frac{s_{l} \sigma_{l}}{u_{l}} + d_{l} + \frac{c_*}{\sqrt{u_{l}}} \mathcal{N}_l \right\} \right), \label{limit_sampling_probability_2}
\end{align}
where the $\mathcal{N}_l$ are independent standard Gaussian random variables.

However, as discussed at the beginning of Section \ref{diffusion_mab}, the function $(u,s) \mapsto p_k(u,s)$ from (\ref{limit_sampling_probability_2}), as well as $(u,s) \mapsto \sqrt{p_k(u,s)}$, are no longer Lipschitz continuous for points near $u_l = 0$, $l \in [K]$.
Nevertheless, the problem with these $p_k$ and $\sqrt{p_k}$ functions only exists for an infinitesimally small initial time interval.
Whenever all inputs $U_l(t)$, $l \in [K]$ to the $u_l$ components in (\ref{limit_sampling_probability_2}) become strictly positive, then from that time onward, there is local Lipschitz continuity for the $p_k$ and $\sqrt{p_k}$ functions, which together with boundedness, ensures that there is a unique strong solution to the SDE.
In Theorem \ref{diffusion_thm3}, we use $\epsilon$-warm-start (recall Definition \ref{def:epsilon-warm-start}) to ensure local Lipschitz continuity holds.
The proof of Theorem \ref{diffusion_thm3} is a direct modification of those of Theorems \ref{diffusion_thm1} and \ref{diffusion_thm2}, and is thus omitted.

\begin{theorem} \label{diffusion_thm3}
Consider the Gaussian Thompson sampler with any fixed Gaussian prior (no $\gamma$-dependence) under $\epsilon$-warm-start (with initial sampling probabilities $q_k > 0$, $\sum_k q_k = 1$).
Then, Theorems \ref{diffusion_thm1} and \ref{diffusion_thm2} hold with the functions $p_k : [0,\infty)^K \times \mathbb{R}^K \to (0,1)$ defined by:
\begin{align}
p_k(u,s) = 
    \begin{cases}
        q_k & \sum_l u_l < \epsilon \vspace{1.5mm} \\
        \P\left( k = \argmax_{l \in [K]} \left\{ \dfrac{s_{l} \sigma_{l}}{u_{l}} + d_{l} + \dfrac{c_*}{\sqrt{u_{l}}} \mathcal{N}_l \right\} \right) & \sum_l u_l \ge \epsilon,
    \end{cases} \label{epsilon_warm_start_p_k}
\end{align}
where the $\mathcal{N}_l$ are independent standard Gaussian random variables.
\end{theorem}

\section{Further Insights from Diffusion Approximations} \label{diffusion_applications}

\subsection{Algorithmic Invariance Principle for EF Thompson Samplers} \label{diffusion_general}

So far in this paper, we have focused on the Gaussian Thompson sampler.
In Theorem \ref{diffusion_prop3} below, we show that the sampling behavior and process-level dynamics of general EF Thompson samplers can be approximated by those of the Gaussian Thompson sampler when the sub-optimality gaps are small.

In the literature, the minimax expected regret of EF Thompson samplers has been studied for some specific conjugate prior-likelihood pairs (e.g., beta prior and Bernoulli likelihood, and the Gaussian prior and Gaussian likelihood) \citep{agrawal_etal2013b, agrawal_etal2017}, with each case requiring a dedicated analysis tailored to the particular conjugate structure.
By contrast, Theorem \ref{diffusion_prop3} provides a unified characterization of the sampling behavior and process-level dynamics of EF Thompson samplers that applies broadly across general single-parameter exponential family likelihoods paired with general (bounded support) priors.
This is suggestive of how general EF Thompson samplers behave in minimax-like settings, though it does not directly lead to minimax or near-minimax expected regret guarantees for these algorithms. 
(Minimax sub-optimality gap scaling with the time horizon could in principle differ across EF Thompson samplers, and minimax expected regret bounds would require non-asymptotic analysis over a slightly expanded range of gap scalings in addition to the scaling of the small gap regime in Assumption \ref{assumption0}.)
Moreover, our characterization extends to TS-like algorithms based on sampling from Gaussian (Laplace) approximations of posterior distributions; see \cite{chapelle_etal2011} and Chapter 5 of \cite{russo_etal2019} for discussions of TS-like algorithms with such posterior approximations.

For Theorem \ref{diffusion_prop3}, the main step is to establish that in the small gap regime, the posterior distributions of EF Thompson samplers are approximately Gaussian.
To develop the Gaussian approximation, in this section, we assume that the arm reward distributions are from an exponential family $P^\mu$ parameterized by mean $\mu$, with the form:
\begin{align}
    P^\mu(dx) = \exp(\theta(\mu) \cdot x - \Lambda(\mu)) P(dx). \label{diffusion_exponential_family}
\end{align}
In (\ref{diffusion_exponential_family}), $P$ is a base distribution, $\theta(\mu) \in \mathbb{R}$ is the value of the ``tilting'' parameter resulting in a mean of $\mu$, and $\Lambda$ is the cumulant generating function.
Let $(\underline{\mu},\overline{\mu})$ denote the open interval of all possible mean values achievable by the family $P^\mu$, with finite values of the tilting parameter $\theta(\mu) \in \mathbb{R}$.
For the small gap regime, suppose Assumption \ref{assumption0} holds, with $Q^{\gamma}_k = P^{\mu^{\gamma}_k}$ for the distributions $Q^{\gamma}_k$ with means $\mu^{\gamma}_k$ from Assumption \ref{assumption0}.
For the $\sigma_k$ in (\ref{assumption0_sigma}), here we have $\sigma_k = \sigma_*$ for all $k$, where $\sigma_*^2$ is the variance of $P^{\mu_*}$, with the $\mu_*$ from (\ref{assumption0_mu}).

For simplicity, suppose we know of a (fixed/no $\gamma$-dependence) bounded, open interval $\mathcal{I}$ (with $\underline{\mu} < \inf \mathcal{I} < \sup \mathcal{I} < \overline{\mu}$) containing the reward means of all arms, i.e., $\mu_* \in \mathcal{I}$ and $\mu_k^\gamma \in \mathcal{I}$ for all $k \in [K]$ and sufficiently small $\gamma > 0$.
We consider EF Thompson samplers with posterior updating based on the likelihood of the exponential family $P^\mu$ (with mean $\mu \in \mathcal{I}$), together with any (fixed/no $\gamma$-dependence) prior distribution (for the mean $\mu$) with density that is bounded, supported on $\mathcal{I}$, and continuous and positive on a neighborhood of $\mu_*$.
For simplicity, we use the same prior for every arm, with independence of priors across arms.

The above setup leads to Theorem \ref{diffusion_prop3} below for the EF Thompson sampler in the small gap regime.
The proof of Theorem \ref{diffusion_prop3} is provided in Section \ref{sec:sections_2_4_proofs}.
It uses a custom version of the Bernstein-von Mises (BvM) theorem, i.e., a Gaussian approximation for the posterior distribution, which can be found in Proposition \ref{diffusion_prop1} in Appendix \ref{diffusion_appA1}.
Proposition \ref{diffusion_prop1} establishes the BvM theorem in a strong mode of convergence: almost surely, uniformly over the family of reward distributions.
For Theorem \ref{diffusion_prop3}, the uniformity is needed because the arm means $\mu_k^\gamma$ change with $\gamma$, and the almost sure mode ensures that the approximation holds simultaneously for all sample sizes beyond a threshold.

\begin{theorem}[Algorithmic Invariance Principle for EF Thompson Samplers] \label{diffusion_prop3}
Consider the above setup, with a $K$-armed bandit in the small gap regime of Assumption \ref{assumption0} (with iid rewards for each arm).
Suppose the arm reward distributions belong to an exponential family of the form in (\ref{diffusion_exponential_family}), and that the corresponding EF Thompson sampler uses a prior with bounded density that is continuous and positive in a neighborhood of $\mu_*$.

Then, for the EF Thompson sampler under $\epsilon$-warm-start (with initial sampling probabilities $q_k > 0$, $\sum_k q_k = 1$), under the random table model of reward feedback, we have for the processes $(U^\gamma,S^\gamma)$ from (\ref{diffusion_r_process1})-(\ref{diffusion_s_process1}):
\begin{align}
    (U^{\gamma},S^{\gamma}) \Rightarrow (U,S) \nonumber
\end{align}
as $\gamma \downarrow 0$ in $D_{2K}[0,\infty)$, where $(U,S)$ is the unique strong solution to the SDE in (\ref{diffusion_sde1})-(\ref{diffusion_sde3}) of Theorem \ref{diffusion_thm1}, with the functions $p_k : [0,\infty)^K \times \mathbb{R}^K \to (0,1)$ defined by:
\begin{align}
p_k(u,s) = 
    \begin{cases}
        q_k & \sum_l u_l < \epsilon \vspace{1.5mm} \\
        \P\left( k = \argmax_{l \in [K]} \left\{ \dfrac{s_{l} \sigma_*}{u_{l}} + d_{l} + \dfrac{\sigma_*}{\sqrt{u_{l}}} \mathcal{N}_l \right\} \right) & \sum_l u_l \ge \epsilon,
    \end{cases} \label{well-specified_sampling_probability}
\end{align}
where the $\mathcal{N}_l$ are independent standard Gaussian random variables.

Equivalently, under the reward stack model of reward feedback, we have for the processes $(U^\gamma,Z^\gamma \circ U^\gamma)$ from (\ref{diffusion_r_process1}) and (\ref{diffusion_z_process1}):
\begin{align}
    (U^\gamma,Z^\gamma \circ U^\gamma) \Rightarrow (U,B \circ U) \nonumber
\end{align}
as $\gamma \downarrow 0$ in $D_{2K}[0,\infty)$, where $U$ is the unique (in distribution) non-anticipative weak solution to the stochastic ODE in (\ref{diffusion_ode1})-(\ref{diffusion_ode2}) of Theorem \ref{diffusion_thm2}, with standard $K$-dimensional Brownian motion $B$ and the functions $p_k$ in (\ref{well-specified_sampling_probability}).

Furthermore, for regret, (\ref{regret_weak_convergence}) continues to hold.
\end{theorem}

\begin{remark}
    The conclusion of Theorem \ref{diffusion_prop3} (with the $p_k(u,s)$ in (\ref{well-specified_sampling_probability})) matches that of Theorem \ref{diffusion_thm3} (with the $p_k(u,s)$ in (\ref{epsilon_warm_start_p_k})) when (in the context of Theorem \ref{diffusion_thm3}) the limit variances $\sigma_k^2$ from (\ref{assumption0_sigma}) match the variance $c_*^2$ used in the Gaussian likelihood of the Gaussian Thompson sampler.
\end{remark}

\subsection{Algorithmic Invariance Principle for Non-parametric Bootstrap Sampler} \label{diffusion_bootstrap}

The bootstrap and related ideas such as subsampling have recently been proposed as mechanisms for exploration in bandit problems \citep{baransi_etal2014, eckles_etal2014, osband_etal2015, tang_etal2015, elmachtoub_etal2017, vaswani_etal2018, kveton_etal2019a, kveton_etal2019b, russo_etal2019, kveton_etal2020a, kveton_etal2020b, baudry_etal2020}.
In this section, we consider the non-parametric bootstrap sampler discussed in the Introduction, which is one natural implementation of bootstrapping to induce exploration in bandit problems.
For the non-parametric bootstrap sampler, in each time period, a single non-parametric bootstrapped sample mean is generated for each arm, and the arm with the largest one is played (with ties broken arbitrarily).
The bootstrap resampling is done independently across arms, conditional on the observed data, which is in accordance with the fact that each arm reward distribution is learned separately in the multi-armed bandit model.

In Theorem \ref{diffusion_prop5} below, we show that for general reward distributions, the sampling behavior and process-level dynamics of the non-parametric bootstrap sampler can be approximated by those of the Gaussian Thompson sampler when the sub-optimality gaps are small.
This is similar in spirit to Theorem \ref{diffusion_prop3}, but here the reward distributions do not need to belong to an exponential family.
Here, we allow for a general family of reward distributions $P^\mu$ that is parameterized by mean $\mu$ (assumed for simplicity) and satisfies mild regularity conditions (discussed below).
In settings with small sub-optimality gaps, given the effectiveness of the Gaussian Thompson sampler, Theorem \ref{diffusion_prop5} suggests that the non-parametric bootstrap sampler can also be an effective means of balancing exploration and exploitation.
Of course, the non-parametric bootstrap sampler has the additional benefit of not requiring distributional choices in algorithm design.

The mild regularity conditions on $P^\mu$ are that $\mu \in \mathcal{I}$, where $\mathcal{I} \subset \mathbb{R}$ is a bounded, open interval, and that $\mu \mapsto \sigma^\mu$ is continuous on $\mathcal{I}$, where $(\sigma^\mu)^2$ denotes the variance of $P^\mu$.
(Neither $P^\mu$ nor $\mathcal{I}$ is part of the algorithm design of the non-parametric bootstrap sampler.)
Also, we assume the conditions in (\ref{bootstrap_uniform_integrability}) and (\ref{variance_lower_bound_1}) hold.
For the small gap regime, suppose Assumption \ref{assumption0} holds, with $Q^{\gamma}_k = P^{\mu^{\gamma}_k}$ for the distributions $Q^{\gamma}_k$ with means $\mu^{\gamma}_k$ from Assumption \ref{assumption0}.
For the $\sigma_k$ in (\ref{assumption0_sigma}), here we have $\sigma_k = \sigma_*$ for all $k$, where $\sigma_*^2$ is the variance of $P^{\mu_*}$, with the $\mu_*$ from (\ref{assumption0_mu}).

The above setup leads to Theorem \ref{diffusion_prop5} below for the non-parametric bootstrap sampler in the small gap regime.
The proof of Theorem \ref{diffusion_prop5} is provided in Section \ref{sec:sections_2_4_proofs}.
The proof is similar to that of Theorem \ref{diffusion_prop3}, except we use a custom Gaussian approximation for the bootstrapped sample mean, which is developed in Proposition \ref{diffusion_prop2} in Appendix \ref{diffusion_appA2}.
Proposition \ref{diffusion_prop2} establishes the bootstrap Gaussian approximation in a strong mode of convergence: almost surely, uniformly over the family of reward distributions.
For Theorem \ref{diffusion_prop5} (as was the case for Theorem \ref{diffusion_prop3}), the uniformity is needed because the arm means $\mu_k^\gamma$ change with $\gamma$, and the almost sure mode ensures that the approximation holds simultaneously for all sample sizes beyond a threshold.

\begin{theorem}[Algorithmic Invariance Principle for Non-parametric Bootstrap Sampler] \label{diffusion_prop5}
Consider the above setup, with a $K$-armed bandit in the small gap regime of Assumption \ref{assumption0} (with iid rewards for each arm).
With $X^\mu \sim P^\mu$ for each $\mu \in \mathcal{I}$, suppose that $\mu \mapsto \sigma^\mu$ is continuous on $\mathcal{I}$ and the following are satisfied:
\begin{align}
    & \lim_{y \to \infty} \sup_{\mu \in \mathcal{I}} \E[(X^\mu)^2 \indic{(X^\mu)^2 > y}] = 0 \label{bootstrap_uniform_integrability} \\
    & \inf_{\mu \in \mathcal{I}} \sigma^\mu > 0. \label{variance_lower_bound_1}
\end{align}
Then, for the non-parametric bootstrap sampler under $\epsilon$-warm-start (with initial sampling probabilities $q_k > 0$, $\sum_k q_k = 1$), under the random table model of reward feedback, we have for the processes $(U^\gamma,S^\gamma)$ from (\ref{diffusion_r_process1})-(\ref{diffusion_s_process1}):
\begin{align}
    (U^{\gamma},S^{\gamma}) \Rightarrow (U,S) \nonumber
\end{align}
as $\gamma \downarrow 0$ in $D_{2K}[0,\infty)$, where $(U,S)$ is the unique strong solution to the SDE in (\ref{diffusion_sde1})-(\ref{diffusion_sde3}) of Theorem \ref{diffusion_thm1}, with the functions $p_k$ in (\ref{well-specified_sampling_probability}).

Equivalently, under the reward stack model of reward feedback, we have for the processes $(U^\gamma,Z^\gamma \circ U^\gamma)$ from (\ref{diffusion_r_process1}) and (\ref{diffusion_z_process1}):
\begin{align}
    (U^\gamma,Z^\gamma \circ U^\gamma) \Rightarrow (U,B \circ U) \nonumber
\end{align}
as $\gamma \downarrow 0$ in $D_{2K}[0,\infty)$, where $U$ is the unique (in distribution) non-anticipative weak solution to the stochastic ODE in (\ref{diffusion_ode1})-(\ref{diffusion_ode2}) of Theorem \ref{diffusion_thm2}, with standard $K$-dimensional Brownian motion $B$ and the functions $p_k$ in (\ref{well-specified_sampling_probability}).

Furthermore, for regret, (\ref{regret_weak_convergence}) continues to hold.
\end{theorem}

\subsection{Model Mis-specification} \label{diffusion_mis-specification}

In this section, we show that in the small gap regime of Assumption \ref{assumption0}, the regret of the Gaussian Thompson sampler, and that of other TS variants like EF Thompson samplers, are insensitive to mis-specification of the reward distributions.
Asymptotically, in the small gap regime, only the limit means and variances (as in (\ref{assumption0_mu})-(\ref{assumption0_sigma})) of the reward distributions influence the dynamics of the Gaussian Thompson sampler.
So, in Theorems \ref{diffusion_thm1}-\ref{diffusion_thm3}, mis-specification corresponds to mis-match between the limit variances $\sigma_k^2$ in (\ref{assumption0_sigma}) and the variance $c_*^2$ specified in the Gaussian likelihood.

In Proposition \ref{diffusion_prop4} below, 
we establish that in the small gap regime of Assumption \ref{assumption0}, the regret (as expressed in (\ref{diffusion_regret_connection}) in Remark \ref{rmk0}) of the Gaussian Thompson sampler (on the $1/\sqrt{\gamma}$ scale) is continuous with respect to the limit variances $\sigma := (\sigma_k, \, k \in [K])$.
As mentioned in the Introduction, this contrasts with the results in the instance-dependent Lai-Robbins asymptotic regime \citep{lai_etal1985}.
In that setting, as recently shown in \cite{fan_etal2021}, the slightest amount of reward distribution mis-specification (e.g., setting the variance parameter of a bandit algorithm to be just slightly less than the true variance of the rewards), can cause the regret performance to sharply deteriorate (from scaling as $\log(n)$ to polynomial in $n$ with horizon $n$).
Furthermore, previously in Section \ref{diffusion_general}, we showed that EF Thompson samplers can be approximated by the Gaussian Thompson sampler in the small gap regime.
This suggests that in the small gap regime, the insensitivity of TS to model mis-specification extends to other settings as well.

\begin{proposition} \label{diffusion_prop4}
Let $(U,S)$ denote the unique strong solution to the SDE (\ref{diffusion_sde1})-(\ref{diffusion_sde3}) in Theorem \ref{diffusion_thm1} with the $\sigma$ dependence expressed in (\ref{limit_sampling_probability}).   
Then, the distribution of $(U,S)$ is continuous with respect to $\sigma$, i.e., for any bounded continuous function $f : D_{2K}[0,\infty) \to \mathbb{R}$, the mapping $\sigma \mapsto \E^\sigma[f(U,S)]$ is continuous.
Moreover, for any fixed $t > 0$,
\begin{align}
    \lim_{\gamma \downarrow 0} \sqrt{\gamma} \, \E^\sigma[\textnormal{Reg}^{\gamma}(\lfloor t/\gamma \rfloor)] = \sum_{k \in [K]} \E^\sigma[U_k(t)] \Delta_k, \nonumber
\end{align}
where $\sigma \mapsto \E^\sigma[U_k(t)]$ is a positive, continuous mapping for each arm $k \in [K]$.

The analogous result holds for the unique (in distribution) non-anticipative weak solution to the stochastic ODE in (\ref{diffusion_ode1})-(\ref{diffusion_ode2}) of Theorem \ref{diffusion_thm2}, as well as for both SDE and stochastic ODE solutions in the setting of Theorem \ref{diffusion_thm3} with the $\sigma$-dependence expressed in (\ref{epsilon_warm_start_p_k}).
\end{proposition}

\subsection{Batched Updates} \label{diffusion_batched}

In some settings, it may be impractical to update a bandit algorithm after each time period.
Instead, updates are batched so that the algorithm commits to playing an (adaptively determined) arm for an interval of time (which could also be adaptively determined).
Then, the algorithm is updated all at once with the data collected during the interval.
For a time horizon of $O(1/\gamma)$, suppose the batch sizes are pre-determined before the start of the experiment and are $o(1/\gamma)$.
Then, in the small gap regime, we would obtain weak convergence to the same SDEs and stochastic ODEs as in the setting of ordinary non-batched TS.
Indeed, a time interval of $o(1/\gamma)$ in the discrete pre-limit system corresponds to (after multiplying by $\gamma$) an infinitesimally small time interval in the continuous limit system.
This suggests that as long as the number of batches increases to infinity (possibly at an arbitrarily slow rate) as $\gamma \downarrow 0$, and each batch is not too large ($o(1/\gamma)$ time periods), then the dynamics of TS will be approximately the same as in the non-batched setting.
To make this precise, we have the following corollary, whose straightforward proof is omitted.
\begin{corollary} \label{diffusion_prop10}
For the Gaussian Thompson sampler with batches of size $o(1/\gamma)$, Theorems \ref{diffusion_thm1}, \ref{diffusion_thm2} and \ref{diffusion_thm3} hold with the same conclusions as in the non-batched setting.
\end{corollary}
The discussion and corollary above correspond nicely to results in the literature regarding optimal batching for bandits in the minimax gap regime from the perspective of expected regret.
As shown in \cite{cesabianchi_etal2013}, \cite{perchet_etal2016} and \cite{gao_etal2019}, in the minimax regime, $O(\log\log(1/\gamma))$ number of batches is necessary and sufficient (sufficient for specially designed algorithms) to achieve the optimal order of expected regret.

We can also consider settings with batches of size $O(1/\gamma)$.
Fix a sequence increasing to infinity: $0 = a_0 < a_1 < a_2 < \dots$.
For each $\gamma > 0$, let $a_i^\gamma$ be positive integers for $i=1,2,\dots$ such that $\lim_{\gamma \downarrow 0} \gamma a_i^\gamma = a_i$ for each $i$.
For batched updating, we perform TS updates in time periods $a_i^\gamma$.
For initialization, in time periods $1,\dots,a_1^\gamma - 1$, each arm $k$ is sampled with fixed probability $q_k > 0$ ($\sum_k q_k = 1$).
Then, we have the following result, where the resulting SDE and stochastic ODE limits are Euler discretizations of the original processes with steps (in continuous time) of size $a_{i+1} - a_{i}$ for $i=1,2,\dots$.
\begin{corollary} \label{diffusion_prop11}
For the Gaussian Thompson sampler with batches of size $O(1/\gamma)$ as described above, Theorems \ref{diffusion_thm1}, \ref{diffusion_thm2} and \ref{diffusion_thm3} hold with the SDE limit:
\begin{align}
    dU_k(t) & = p_k(U(a_{i}),S(a_{i})) dt \nonumber \\
    dS_k(t) & = \sqrt{p_k(U(a_{i}),S(a_{i}))} dB_k(t), \quad t \in [a_{i},a_{i+1}), \,\, i=1,2,\dots \nonumber \\
    U_k(a_{1}) & = q_k a_1, \,\,\,\,\, S_k(a_{1}) = \sqrt{q_k} B_k(a_{1}), \quad k \in [K], \nonumber
\end{align}
and the stochastic ODE limit:
\begin{align}
    dU_k(t) & = p_k(U(a_{i}),B \circ U(a_{i})) dt, \quad t \in [a_{i},a_{i+1}), \,\, i=1,2,\dots \nonumber \\
    U_k(a_{1}) & = q_k a_1, \quad k \in [K], \nonumber
\end{align}
with the functions $p_k$ given by (\ref{limit_sampling_probability}) or (\ref{limit_sampling_probability_2}).
\end{corollary}

\section{Proofs for Main Results}

\subsection{Proofs for SDE Approximation} \label{diffusion_sde}

In this section, we prove the SDE approximation in Theorem \ref{diffusion_thm1} (from Section \ref{diffusion_mab_1}).
We first discuss a (random) step function construction (from Section 6 of \cite{kurtz_etal1991}) that can approximate functions in $D_m[0,\infty)$ uniformly with any desired accuracy.
We define this \textit{$\epsilon$-uniform step function approximation} in Definition \ref{diffusion_definition1} and define integration with step function integrands in Definition \ref{diffusion_definition2}.

Then, we introduce Lemma \ref{diffusion_lemma3}, which establishes that the $\epsilon$-uniform step function approximation is a continuous mapping that preserves adaptedness and martingale properties.
As indicated in Lemma \ref{diffusion_lemma3}, by applying the $\epsilon$-uniform step function approximation to pre-limit and limit integrands, the corresponding approximated pre-limit and limit stochastic integrals are mathematically well-defined, and we have weak convergence of the former to the latter.
Lemma \ref{diffusion_lemma3} also helps with verification that components of the limit process are Brownian motion.

Next, we provide the proof of Theorem \ref{diffusion_thm1}, which uses Lemma \ref{diffusion_lemma3} together with the CMT and standard weak convergence arguments.
We conclude the section with the proof of Lemma \ref{diffusion_lemma3}, followed by Lemmas \ref{diffusion_lemma1} and \ref{diffusion_lemma2} and their proofs, which establish tightness of stochastic processes and convergence to Brownian motion, as used in the proof of Theorem \ref{diffusion_thm1}.

\begin{definition}[$\epsilon$-uniform Step Function Approximation] \label{diffusion_definition1}
For any $\epsilon > 0$, we construct a random step function mapping $\chi^\epsilon : D_m[0,\infty) \to D_m[0,\infty)$ as follows.
We use the $\ell^1$ norm, with $\| w \| := \sum_{i=1}^m |w_i|$ for $w \in \mathbb{R}^m$.
For any $z \in D_m[0,\infty)$, inductively define the stopping times $\tau_j$ starting with $\tau_0 = 0$:
\begin{align}
\tau_{j+1} & = \inf\{t > \tau_j : \max(\|z(t) - z(\tau_j)\|, \| z(t-) - z(\tau_j) \|) \ge \epsilon \rho_j \}, \label{diffusion_step1} 
\end{align}
where $\rho_j \overset{\text{iid}}{\sim} \text{Unif}(1/2,1)$.
(Note that each $z$ has its own corresponding sequence of stopping times $\tau_j$.)
Then, define $\chi^\epsilon(z) \in D_m[0,\infty)$ by
\begin{align}
\chi^\epsilon(z)(t) = z(\tau_j), \quad t \in [\tau_j, \tau_{j+1}), \label{diffusion_step2} 
\end{align}
so that $\chi^\epsilon(z)$ is a step function, and a.s.,
\begin{align}
    \sup_{t \ge 0} \| \chi^\epsilon(z)(t) - z(t) \| \le \epsilon. \label{epsilon-uniform-approximation}
\end{align}
\end{definition}

\begin{definition}[Integration with Step Functions] \label{diffusion_definition2}
Let $f,g \in D[a,\infty)$, where $f$ is a step function with jump points $t_1 < \dots < t_j$ on the interval $[a,b]$ (set $t_0 = a$ and $t_{j+1} = b$).
We will always use the following definition of integration for step function integrand $f$ with respect to integrator $g$ on $(a,b]$:
\begin{align}
\int_a^b f(t-) dg(t) = \sum_{i=0}^j f(t_i)\left( g(t_{i+1}) - g(t_i) \right). \nonumber
\end{align}
\end{definition}

\begin{lemma}[Properties of $\epsilon$-uniform Step Function Approximation] \label{diffusion_lemma3}
For $n \ge 1$, let $L^n \in D_{l}[0,\infty)$, $H^n \in D_{m}[0,\infty)$ and $W^n \in D_{m}[0,\infty)$ be sequences of processes adapted to a corresponding sequence of filtrations $\mathcal{F}^n = (\mathcal{F}^n_t, \, t \ge 0)$.
For any $\epsilon > 0$, let $\chi^\epsilon : D_{m}[0,\infty) \to D_{m}[0,\infty)$ be constructed as in Definition \ref{diffusion_definition1}, and define filtrations $\mathcal{G}^n = (\mathcal{G}^n_t, \, t \ge 0)$ by $\mathcal{G}^n_t = \mathcal{F}^n_t \vee \widetilde{\mathcal{G}}$, where $\widetilde{\mathcal{G}} = \sigma(\rho_j, \, j \ge 0)$ (with the $\rho_j$ used to construct $\chi^\epsilon$) is independent of $\mathcal{F}^n$ for all $n$.
If $W^n$ is an $\mathcal{F}^n$-martingale and $(L^n,H^n,W^n) \Rightarrow (L,H,W)$ in $D_{l+2m}[0,\infty)$ as $n \to \infty$, then the following hold. \vspace{1mm}
\begin{enumerate}
    \item[(i)] $(L^n,H^n,W^n,\chi^\epsilon(H^n)) \Rightarrow (L,H,W,\chi^\epsilon(H))$ in $D_{l+3m}[0,\infty)$ as $n \to \infty$ \vspace{1mm}
    \item[(ii)] $(L^n,H^n,W^n,\chi^\epsilon(H^n))$ is $\mathcal{G}^n$-adapted and $W^n$ is a $\mathcal{G}^n$-martingale
    \vspace{1mm}
    \item[(iii)] If additionally, $\sup_n \E[W_k^n(t)^2] < \infty$ for each $k \in [m]$ and $t \ge 0$, and marginally (with respect to its own natural filtration) $W$ is a standard $m$-dimensional Brownian motion, then $W$ remains a standard $m$-dimensional Brownian motion with respect to the augmented natural filtration of $(L,H,W,\chi^\epsilon(H))$. 
    Moreover, with $\xi^n = (\xi_k^n, \, k \in [m])$ and $\xi = (\xi_k, \, k \in [m])$ defined via
    \begin{align}
        \xi^n_k(t) & = \int_0^t \chi_k^\epsilon(H^n)(v-) dW_k^n(v) \label{prelimit_stochastic_integral} \\
        \xi_k(t) & = \int_0^t \chi_k^\epsilon(H)(v-) dW_k(v), \label{limit_stochastic_integral}
    \end{align}
    we have
    \begin{align}
        (L^n,H^n,W^n,\xi^n) \Rightarrow (L,H,W,\xi). \label{weak_convergence_stochastic_integral}
    \end{align}
    in $D_{l+3m}[0,\infty)$ as $n \to \infty$.
\end{enumerate}
\end{lemma}

\proof{Proof of Theorem \ref{diffusion_thm1}.}
We start with the discrete approximation (\ref{diffusion_discrete_sde1})-(\ref{diffusion_discrete_sde5}) from our derivation in Section \ref{diffusion_mab_1}.
We denote the joint processes via $(U^{\gamma},S^{\gamma},B^{\gamma},M^{\gamma}) = (U_k^{\gamma},S_k^{\gamma},B_k^{\gamma},M_k^{\gamma}, \, k \in [K])$, and recall that they are processes in $D_{4K}[0,\infty)$.

Our proof strategy is as follows.
We will show that for every subsequence of $(U^{\gamma},S^{\gamma})$, there is a further subsequence which converges weakly to a limit that is a solution to the SDE.
Because the drift and dispersion functions $p_k$ and $\sqrt{p_k}$ of the SDE in (\ref{diffusion_sde1})-(\ref{diffusion_sde2}) are locally Lipschitz continuous (i.e., Lipschitz continuous on any compact subset of the domain $[0,\infty)^K \times \mathbb{R}^K$) and bounded, the SDE has a (global) unique strong solution. 
(For references regarding this result, see Remark \ref{sde_strong_solution_existence_uniqueness}.)
Thus, $(U^{\gamma},S^{\gamma})$ must converge weakly to the unique strong solution of the SDE.

By Lemma \ref{diffusion_lemma1} (stated and proved after the current proof), the joint processes $(U^{\gamma},S^{\gamma},B^{\gamma},M^{\gamma})$ are tight in $D_{4K}[0,\infty)$, and thus, Prohorov's Theorem ensures that for each subsequence, there is a further subsequence which converges weakly to some limit process $(U,S,B,M) = (U_k,S_k,B_k,M_k, \, k \in [K])$ (see Chapter 3 of \cite{ethier_etal1986}, Chapters 1 and 3 of \cite{billingsley_1999}, or Chapter 11 of \cite{whitt_2002}).
From now on, we work with this further subsequence, and for notational simplicity, we still index this further subsequence by $\gamma$.
So, we have
\begin{align}
    (U^{\gamma},S^{\gamma},B^{\gamma},M^{\gamma}) \Rightarrow (U,S,B,M). \label{diffusion_g0}
\end{align}
Because $M^{\gamma}$ is a suitably normalized sum of bounded martingale differences, we have $M_k^{\gamma}(t) \overset{\P}{\to} 0$ for each $k \in [K]$ and any $t > 0$ as $\gamma \downarrow 0$, and thus, $M$ is the $D_K[0,\infty)$ zero process.

Now define the processes $A^{\gamma} = (A_k^{\gamma}, \, k \in [K])$ and $A = (A_k, \, k \in [K])$, where
\begin{align}
A_k^{\gamma}(t) & = p_k^{\gamma}(U^{\gamma}(t),S^{\gamma}(t)) \label{diffusion_g1} \\
A_k(t) & = p_k(U(t),S(t)). \label{diffusion_g2}
\end{align}
Note that $p_k^{\gamma}(u,s) \to p_k(u,s)$ as $\gamma \downarrow 0$ uniformly for $(u,s)$ in compact subsets of $[0,\infty)^K \times \mathbb{R}^K$, and $p_k(u,s)$ is continuous at all $(u,s) \in [0,\infty)^K \times \mathbb{R}^K$.
These properties also hold for $\sqrt{p_k^{\gamma}(u,s)}$ and $\sqrt{p_k(u,s)}$.
So, by the Generalized CMT (Lemma \ref{diffusion_lemma0}) applied to the processes $A_k^{\gamma}(t)$, $A_k(t)$, $\sqrt{A_k^{\gamma}(t)}$ and $\sqrt{A_k(t)}$, we have from (\ref{diffusion_g0}),
\begin{align}
(U^{\gamma},S^{\gamma},B^{\gamma},M^{\gamma},A^{\gamma},\sqrt{A^{\gamma}}) \Rightarrow (U,S,B,M,A,\sqrt{A}), \label{diffusion_t1}
\end{align}
where we denote $\sqrt{A^{\gamma}} = (\sqrt{A_k^{\gamma}}, \, k \in [K])$ and $\sqrt{A} = (\sqrt{A_k}, \, k \in [K])$.
Additionally, define the processes $\widetilde{U}^{\gamma} = (\widetilde{U}_k^{\gamma}, \, k \in [K])$ and $\widetilde{U} = (\widetilde{U}_k, \, k \in [K])$, where
\begin{align}
\widetilde{U}_k^{\gamma}(t) & = \int_0^t A_k^{\gamma}(v) dv \label{diffusion_rntilde} \\
\widetilde{U}_k(t) & = \int_0^t A_k(v) dv. \label{diffusion_rtilde}
\end{align}
Recall that
\begin{align*}
U^{\gamma}_k(t) = \gamma \sum_{i=0}^{\lfloor t/\gamma \rfloor - 1} A_k^{\gamma}(i \gamma) + M^{\gamma}_k(t).
\end{align*}
For each $k \in [K]$, because $M^{\gamma}_k$ converges weakly to the $D[0,\infty)$ zero process and also
\begin{align*}
\sup_{t \ge 0} \abs{\gamma \sum_{i=0}^{\lfloor t/\gamma \rfloor - 1} A_k^{\gamma}(i \gamma) - \widetilde{U}_k^{\gamma}(t)} \le \gamma,
\end{align*}
we have for any $T > 0$,
\begin{align}
\sup_{0 \le t \le T} \abs{U_k^{\gamma}(t) - \widetilde{U}_k^{\gamma}(t)} \overset{\P}{\to} 0. \label{diffusion_t0}
\end{align}
Thus, by the continuity of integration with respect to the Skorohod metric (Theorem 11.5.1 of \cite{whitt_2002}) and the CMT, we have from (\ref{diffusion_t1}),
\begin{align}
(U^{\gamma},S^{\gamma},B^{\gamma},\widetilde{U}^{\gamma},\sqrt{A^{\gamma}}) \Rightarrow (U,S,B,\widetilde{U},\sqrt{A}). \label{diffusion_t2}
\end{align}

Let $\epsilon > 0$.
Let $\chi^\epsilon$ be the random step function mapping defined in (\ref{diffusion_step1})-(\ref{diffusion_step2}).
Using (\ref{diffusion_t2}) and Lemma \ref{diffusion_lemma3}(i), applying $\chi^\epsilon$ to $\sqrt{A^{\gamma}}$ and $\sqrt{A}$, we have
\begin{align}
(U^{\gamma},S^{\gamma},B^{\gamma},\widetilde{U}^{\gamma},\chi^\epsilon(\sqrt{A^{\gamma}})) \Rightarrow (U,S,B,\widetilde{U},\chi^\epsilon(\sqrt{A})). \label{diffusion_t3}
\end{align}
Let $\widetilde{\mathcal{H}}^{\gamma} = (\widetilde{\mathcal{H}}^{\gamma}_t, \, t \ge 0)$ denote the continuous, piecewise constant (and right-continuous) interpolation of the discrete-time filtration $\mathcal{H}^{\gamma} = (\mathcal{H}^{\gamma}_j, \, j \ge 0)$ defined in (\ref{h_j_gamma}).
For fixed $\gamma > 0$, the processes on the left side of (\ref{diffusion_t2}) are adapted to $\widetilde{\mathcal{H}}^{\gamma}$, and $B^\gamma$ is a square-integrable $\widetilde{\mathcal{H}}^{\gamma}$-martingale.
By Lemma \ref{diffusion_lemma3}(ii), the processes on the left side of (\ref{diffusion_t3}) are adapted to the expanded version of the filtration $\widetilde{\mathcal{H}}^{\gamma}$ that includes the exogenous randomization used to construct $\chi^\epsilon$, and $B^\gamma$ continues to be a square-integrable martingale with respect to the expanded filtration.
We can then define stochastic integration for the pre-limit processes $\chi^\epsilon(\sqrt{A^{\gamma}})$ and $B^{\gamma}$ in (\ref{diffusion_t3}) via
\begin{align}
\widehat{S}^{\gamma}_k(t) & = \int_0^t \chi_k^\epsilon(\sqrt{\displaystyle A^{\gamma}})(v-) \,\, dB_k^{\gamma}(v), \label{diffusion_ytilden}
\end{align}
with $\widehat{S}^{\gamma} = (\widehat{S}^{\gamma}_k, \, k \in [K])$.
By Lemma \ref{diffusion_lemma2} (stated and proved after the current proof), $B$ is marginally a standard $K$-dimensional Brownian motion. 
Then, by Lemma \ref{diffusion_lemma3}(iii), $B$ remains a standard $K$-dimensional Brownian motion with respect to the augmented natural filtration of the limit processes $(U,S,B,\widetilde{U},\chi^\epsilon(\sqrt{A}))$ in (\ref{diffusion_t3}).
So, we can also define stochastic integration for the limit processes $\chi^\epsilon(\sqrt{A})$ and $B$ in (\ref{diffusion_t3}) via
\begin{align}
\widehat{S}_k(t) & = \int_0^t \chi_k^\epsilon(\sqrt{A})(v-) \,\, dB_k(v), \label{diffusion_yhat}
\end{align}
with $\widehat{S} = (\widehat{S}_k, \, k \in [K])$.
Moreover, using Lemma \ref{diffusion_lemma3}(iii) and (\ref{diffusion_t3}), we have weak convergence of the stochastic integrals in (\ref{diffusion_ytilden}) to those in (\ref{diffusion_yhat}), jointly with the weak convergence of other components in (\ref{diffusion_t3}):
\begin{align}
(U^{\gamma},S^{\gamma},B^{\gamma},\widetilde{U}^{\gamma},\widehat{S}^{\gamma}) \Rightarrow (U,S,B,\widetilde{U},\widehat{S}). \label{diffusion_t4}
\end{align}
Recall from (\ref{diffusion_discrete_sde2}), (\ref{diffusion_discrete_sde5}) and (\ref{diffusion_g1}), that for each $k \in [K]$, 
\begin{align}
S_k^{\gamma}(t) = \int_0^t \sqrt{A_k^{\gamma}(v-)} \, dB_k^{\gamma}(v). \label{diffusion_yn}
\end{align}
Also, define the process $\widetilde{S} = (\widetilde{S}_k, \, k \in [K])$, where
\begin{align}
\widetilde{S}_k(t) & = \int_0^t \sqrt{A_k(v-)} \, dB_k(v). \label{diffusion_ytilde}
\end{align}

For each $k \in [K]$, we now show that $\widehat{S}^{\gamma}_k$ in (\ref{diffusion_ytilden}) approximates $S_k^{\gamma}$ in (\ref{diffusion_yn}), and $\widehat{S}_k$ in (\ref{diffusion_yhat}) approximates $\widetilde{S}_k$ in (\ref{diffusion_ytilde}).
For a square integrable martingale process $\xi$, we use $t \mapsto \langle \xi, \xi \rangle_t$ to denote its (predictable) quadratic variation process.
For each $k \in [K]$ and any $T > 0$, we have
\begin{align}
\E\left[\sup_{0 \le t \le T} \abs{S_k^{\gamma}(t) - \widehat{S}_k^{\gamma}(t)} \right] & \le 2 \E\left[ \left( \int_0^T \left( \chi_k^\epsilon(\sqrt{\displaystyle A^{\gamma}})(t-) - \sqrt{A_k^{\gamma}(t-)} \right) dB_k^{\gamma}(t) \right)^2 \right]^{1/2} \label{doob_maximal_inequality} \\
& = 2 \E\left[ \int_0^T \left( \chi_k^\epsilon(\sqrt{\displaystyle A^{\gamma}})(t-) - \sqrt{A_k^{\gamma}(t-)} \right)^2 d\langle B_k^{\gamma}, B_k^{\gamma} \rangle_t \right]^{1/2} \label{ito_isometry} \\
& \le 2 \E\left[ \int_0^T \epsilon^2 d\langle B_k^{\gamma}, B_k^{\gamma} \rangle_t \right]^{1/2} \label{epsilon_uniform_bound} \\
& = 2 \epsilon \sqrt{\gamma} \frac{1}{\sigma_k} \E\left[ \sum_{i=0}^{\lfloor T/\gamma \rfloor - 1} \E\left[ \frac{I_k^{\gamma}(i+1) (X_k^{\gamma}(i+1) - \mu_k^{\gamma})^2}{\displaystyle p_k^{\gamma}(U^{\gamma}(i \gamma),S^{\gamma}(i \gamma))} \; \biggl| \; \mathcal{H}_{i}^{\gamma} \right] \right]^{1/2} \nonumber \\
& \le 2 \epsilon \frac{\sigma_k^\gamma}{\sigma_k} \sqrt{T}, \label{diffusion_t5}
\end{align}
where (\ref{doob_maximal_inequality}) follows from Doob's Maximal Inequality (Theorem 3.8(iv) of Chapter 1 of \cite{karatzas_etal1998}), (\ref{ito_isometry}) follows from the It\^{o} Isometry for square-integrable martingale integrators, and (\ref{epsilon_uniform_bound}) follows from the fact that $\chi^\epsilon$ is an $\epsilon$-uniform approximation (as in (\ref{epsilon-uniform-approximation})).
Similarly, for each $k \in [K]$ and any $T > 0$,
\begin{align}
\E\left[\sup_{0 \le t \le T} \abs{\widehat{S}_k(t) - \widetilde{S}_k(t)} \right] \le 2 \epsilon \sqrt{T}. \label{diffusion_t6}
\end{align}
Putting together (\ref{diffusion_t0}), (\ref{diffusion_t4}), and (\ref{diffusion_t5})-(\ref{diffusion_t6}), and sending $\epsilon \downarrow 0$, we obtain:
\begin{align}
U & = \widetilde{U} \label{diffusion_t10} \\
S & = \widetilde{S}. \label{diffusion_t7}
\end{align}
Recalling the definition of $A_k$ in (\ref{diffusion_g2}), $\widetilde{U}$ in (\ref{diffusion_rtilde}), and $\widetilde{S}$ in (\ref{diffusion_ytilde}), we see from (\ref{diffusion_t10})-(\ref{diffusion_t7}) that the limit processes $(U,S,B)$ satisfy the SDE:
\begin{align}
U_k(t) & = \int_0^t p_k(U(v),S(v)) dv \nonumber \\
S_k(t) & = \int_0^t \sqrt{p_k(U(v),S(v))} dB_k(v), \quad k \in [K]. \nonumber
\end{align}
\halmos
\endproof

\proof{Proof of Lemma \ref{diffusion_lemma3}.}
Parts (i) and (ii) follow from Lemma 6.1 of \cite{kurtz_etal1991}.
(See also the proof of Theorem 2.2 of \cite{kurtz_etal1991}.)
We now establish part (iii).

The part (iii) assumption that $\sup_n \E[W_k^n(t)^2] < \infty$ implies that $W_k^n(t)$ is a uniformly integrable sequence for each $k \in [m]$ and $t > 0$.
Using parts (i) and (ii), together with this uniform integrability, Theorem 5.3 of \cite{whitt_2007} ensures that $W$ remains a martingale with respect to the augmented natural filtration of $(L,H,W,\chi^\epsilon(H))$. 
(See also Problem 7 from Chapter 7 of \cite{ethier_etal1986}.)
Since $W$ is marginally a standard $m$-dimensional Brownian motion, it has quadratic variation $\langle W_k, W_k \rangle_t = t$ for all $k$ and cross-variation $\langle W_k, W_l \rangle_t = 0$ for all $k \ne l$ a.s.
Then, using the L\'{e}vy characterization of multi-dimensional Brownian motion (Theorem 3.16 from Chapter 3 of \cite{karatzas_etal1998}), $W$ remains a standard $m$-dimensional Brownian motion with respect to the augmented natural filtration of $(L,H,W,\chi^\epsilon(H))$.

Lastly, we establish (\ref{weak_convergence_stochastic_integral}).
The stochastic integrals in (\ref{prelimit_stochastic_integral}) and (\ref{limit_stochastic_integral}) are well-defined with predictable step function integrands and square-integrable martingale integrators.
To obtain weak convergence of the stochastic integrals in (\ref{prelimit_stochastic_integral}) to those in (\ref{limit_stochastic_integral}), and moreover the joint weak convergence in (\ref{weak_convergence_stochastic_integral}), consider the following observations.
Let $x^n \in D_m[0,\infty)$ and $y^n \in D_m[0,\infty)$ be sequences such that $(x^n,y^n) \to (x,y)$ in $D_{2m}[0,\infty)$ as $n \to \infty$.
Suppose also that for each component $k \in [m]$, $x_k^n$ is a step function and the number of discontinuities of $x_k^n$ in any bounded time interval is uniformly bounded in $n$.
Then, defining $z^n \in D_m[0,\infty)$ and $z \in D_m[0,\infty)$ with components 
\begin{align}
    z_k^n(t) & = \int_0^t x_k^n(v-) dy_k^n(v) \nonumber \\
    z_k(t) & = \int_0^t x_k(v-) dy_k(v), \nonumber
\end{align}
we have 
\begin{align}
    (x^n,y^n,z^n) \to (x,y,z) \nonumber
\end{align} 
in $D_{3m}[0,\infty)$ as $n \to \infty$.
As noted in \cite{kurtz_etal1991}, by the continuity of stochastic integration with step function integrands as considered above, we obtain the weak convergence of the stochastic integrals in (\ref{prelimit_stochastic_integral}) to those in (\ref{limit_stochastic_integral}), and moreover the desired joint weak convergence in (\ref{weak_convergence_stochastic_integral}).
\halmos
\endproof

\begin{lemma} \label{diffusion_lemma1}
The processes $(U^{\gamma},S^{\gamma},B^{\gamma},M^{\gamma})$ defined in (\ref{diffusion_discrete_sde1})-(\ref{diffusion_discrete_sde5}) are tight in $D_{4K}[0,\infty)$.
\end{lemma}

\proof{Proof of Lemma \ref{diffusion_lemma1}.}
We recall that the processes have the following expressions for $k \in [K]$.
\begin{align}
U^{\gamma}_k(t) & = \gamma \sum_{i=1}^{\lfloor t/\gamma \rfloor} I_k^{\gamma}(i) \label{diffusion_l1} \\
S^{\gamma}_k(t) & = \sqrt{\gamma} \frac{1}{\sigma_k} \sum_{i=1}^{\lfloor t/\gamma \rfloor} I_k^{\gamma}(i) (X_k^{\gamma}(i) - \mu_k^{\gamma}) \label{diffusion_l2} \\
M^{\gamma}_k(t) & = \gamma \sum_{i=0}^{\lfloor t/\gamma \rfloor - 1} \left( I_k^{\gamma}(i+1) - p_k^{\gamma}(U^{\gamma}(i \gamma),S^{\gamma}(i \gamma)) \right) \label{diffusion_l3} \\
B^{\gamma}_k(t) & = \sqrt{\gamma} \frac{1}{\sigma_k} \sum_{i=0}^{\lfloor t/\gamma \rfloor - 1} \frac{I_k^{\gamma}(i+1) (X_k^{\gamma}(i+1) - \mu_k^{\gamma})}{\sqrt{\displaystyle p_k^{\gamma}(U^{\gamma}(i \gamma),S^{\gamma}(i \gamma))}} \label{diffusion_l4}
\end{align}
Note that (\ref{diffusion_l1})-(\ref{diffusion_l2}) are just different expressions of the same quantities in (\ref{diffusion_discrete_sde1})-(\ref{diffusion_discrete_sde2}).
Also, the process in (\ref{diffusion_l1}) is uniformly bounded and increasing, and those in (\ref{diffusion_l2})-(\ref{diffusion_l4}) are square-integrable martingales.

By Lemma \ref{diffusion_lemma5}, to show tightness of the joint processes $(U^{\gamma},S^{\gamma},B^{\gamma},M^{\gamma})$, we just need to show tightness of each component sequence of processes and each pairwise sum of component sequences of processes.
We use Lemma \ref{diffusion_lemma6} to verify tightness in each case.
Condition (\ref{diffusion_tightness1}) can be directly verified using a sub-martingale maximal inequality (Theorem 3.8(i) of Chapter 1 of \cite{karatzas_etal1998}), along with a union bound when dealing with pairwise sums of component processes.
Conditions (\ref{diffusion_tightness2})-(\ref{diffusion_tightness3}) can also be directly verified.
(Here, the corresponding sequences are $\xi^\gamma$ and $A_\delta^{\gamma}(T)$, which are indexed by $\gamma > 0$ with $\gamma \downarrow 0$.)
Since $\lim_{\gamma \downarrow 0} \sigma_k^\gamma = \sigma_k$ for each $k \in [K]$, there exists $\gamma_0 > 0$ sufficiently small such that $\max_{k \in [K]} \sup_{\gamma \in (0,\gamma_0]} (\sigma_k^\gamma / \sigma_k)^2 \le 2$.
Since we are only interested in limit behavior as $\gamma \downarrow 0$, it suffices to restrict attention to $\gamma \in (0,\gamma_0]$.
For any $T > 0$, $\delta > 0$ and $\gamma \in (0,\gamma_0]$, (\ref{diffusion_tightness2}) holds for $\xi^\gamma$ equal to each individual component process of $(U^{\gamma},S^{\gamma},B^{\gamma},M^{\gamma})$ by setting $A_\delta^{\gamma}(T) = 2 \max(\delta + \gamma, (\delta + \gamma)^2)$, and (\ref{diffusion_tightness2}) holds for $\xi^\gamma$ equal to each pairwise sum of component processes of $(U^{\gamma},S^{\gamma},B^{\gamma},M^{\gamma})$ by setting $A_\delta^{\gamma}(T) = 8 \max(\delta + \gamma, (\delta + \gamma)^2)$ (using the bound: $(x+y)^2 \le 2x^2 + 2y^2$).
Then, for (\ref{diffusion_tightness3}), we have $\lim_{\delta \downarrow 0} \limsup_{\gamma \downarrow 0} A_\delta^{\gamma}(T) = 0$ in all cases.
\halmos
\endproof

\begin{lemma} \label{diffusion_lemma2}
Marginally, $B^{\gamma} \Rightarrow B$ as $\gamma \downarrow 0$, where $B$ is standard $K$-dimensional Brownian motion.
\end{lemma}

\proof{Proof of Lemma \ref{diffusion_lemma2}.}
We apply the martingale functional central limit theorem (CLT) stated in Lemma \ref{diffusion_lemma7}.
Below, we verify (\ref{diffusion_martingale1}) and (\ref{diffusion_martingale2}) to ensure Lemma \ref{diffusion_lemma7} holds.

\noindent \underline{Verification of (\ref{diffusion_martingale1})} \\
Because $I_j^{\gamma}(i) I_k^{\gamma}(i) = 0$ for $j \ne k$ and all $i \ge 1$ (only one arm is played in each time period $i$), we have $\Sigma_{jk} = 0$ for $j \ne k$.
For the diagonal elements, we have $\Sigma_{kk} = 1$ for each $k \in [K]$, as the following argument shows.
As shorthand, denote $p_{k}^{\gamma}(i) := p_k^{\gamma}(U^{\gamma}(i \gamma),S^{\gamma}(i \gamma))$.
Then,
\begin{align}
    & \gamma \left(\frac{\sigma_k^{\gamma}}{\sigma_k}\right)^2 \sum_{i=0}^{\lfloor t/\gamma \rfloor - 1} \E\left[ \frac{I_k^{\gamma}(i+1)}{p_{k}^{\gamma}(i)} \left(\frac{X_k^{\gamma}(i+1) - \mu_k^{\gamma}}{\sigma_k^{\gamma}}\right)^2 \; \bigg| \; \mathcal{H}_{i}^{\gamma} \right] \nonumber \\
    & = \gamma \left(\frac{\sigma_k^{\gamma}}{\sigma_k}\right)^2 \sum_{i=0}^{\lfloor t/\gamma \rfloor - 1} \frac{\E\left[ I_k^{\gamma}(i+1) \; \bigg| \; \mathcal{H}_{i}^{\gamma} \right]}{p_{k}^{\gamma}(i)} \E\left[ \left(\frac{X_k^{\gamma}(i+1) - \mu_k^{\gamma}}{\sigma_k^{\gamma}} \right)^2 \; \bigg| \; \mathcal{H}_{i}^{\gamma} \right] \label{measurable_independent} \\
    & = \gamma \left(\frac{\sigma_k^{\gamma}}{\sigma_k}\right)^2 \lfloor t/\gamma \rfloor \to t \label{quadratic_variation_convergence}
\end{align}
as $\gamma \downarrow 0$.
Here, (\ref{measurable_independent}) follows from $p_{k}^{\gamma}(i) = p_k^{\gamma}(U^{\gamma}(i \gamma),S^{\gamma}(i \gamma))$ being $\mathcal{H}_{i}^{\gamma}$-measurable, and $I_k^{\gamma}(i+1)$ and $(X_k^{\gamma}(i+1) - \mu_k^{\gamma})^2/(\sigma_k^{\gamma})^2$ being independent conditional on $\mathcal{H}_{i}^{\gamma}$.
The convergence in (\ref{quadratic_variation_convergence}) follows from (\ref{assumption0_sigma}) in Assumption \ref{assumption0}.

\noindent \underline{Verification of (\ref{diffusion_martingale2})} \\
For each $k \in [K]$, denote 
\begin{align}
    \xi_k^{\gamma}(i+1) = \frac{I_k^{\gamma}(i+1) (X_k^{\gamma}(i+1) - \mu_k^{\gamma})}{\sqrt{p_{k}^{\gamma}(i)} \cdot \sigma_k}. \nonumber
\end{align}
By Markov's inequality, it suffices to show that
\begin{align}
    \sup_{0 \le i \le \lfloor t/\gamma \rfloor - 1} \E\left[ \xi_k^{\gamma}(i+1)^2 \indic{\abs{\xi_k^{\gamma}(i+1)} > \epsilon/\sqrt{\gamma}} \right] \to 0 \label{diffusion_M2_sufficient}
\end{align}
as $\gamma \downarrow 0$.

We have the following three observations.
1) $(U^{\gamma},S^{\gamma})$ is a tight sequence, as established in Lemma \ref{diffusion_lemma1}, which implies stochastic boundedness of each component with respect to the supremum norm. 
2) $p_k^{\gamma}(u,s) \to p_k(u,s)$ as $\gamma \downarrow 0$ uniformly for $(u,s)$ in compact subsets of $[0,\infty)^K \times \mathbb{R}^K$.
3) $p_k(u,s)$ is continuous and strictly positive for all $(u,s) \in [0,\infty)^K \times \mathbb{R}^K$.
Given these three observations, for any $\eta > 0$, there exists $\delta \in (0,1)$ such that for $\gamma$ sufficiently close to zero,
\begin{align}
    \P\left( \inf_{v \in [0,t]} p_k^{\gamma}(U^{\gamma}(v),S^{\gamma}(v)) < \delta \right) \le \eta. \label{tightness_sampling_probability_bound}
\end{align}
We then have
\begin{align}
    & \E\left[ \xi_k^{\gamma}(i+1)^2 \indic{\abs{\xi_k^{\gamma}(i+1)} > \epsilon/\sqrt{\gamma}} \indic{p_{k}^{\gamma}(i) < \delta} \right] \nonumber \\ 
    & \le \E\left[ \xi_k^{\gamma}(i+1)^2 \indic{p_{k}^{\gamma}(i) < \delta} \right] \nonumber \\
    & = \E\left[ \indic{p_{k}^{\gamma}(i) < \delta} \frac{\E[ I_k^{\gamma}(i+1) \mid \mathcal{H}_{i}^{\gamma} ]}{p_{k}^{\gamma}(i)} \E\left[ \left( \frac{X_k^{\gamma}(i+1) - \mu_k^{\gamma}}{\sigma_k^{\gamma}} \right)^2 \, \bigg| \, \mathcal{H}_{i}^{\gamma} \right] \right] \left(\frac{\sigma_k^{\gamma}}{\sigma_k}\right)^2 \label{diffusion_measurability} \\
    & = \P\left( p_{k}^{\gamma}(i) < \delta \right) \left(\frac{\sigma_k^{\gamma}}{\sigma_k}\right)^2 \label{diffusion_conditional_independence} \\
    & \le \P\left( \inf_{v \in [0,t]} p_k^{\gamma}(U^{\gamma}(v),S^{\gamma}(v)) < \delta \right) \left(\frac{\sigma_k^{\gamma}}{\sigma_k}\right)^2 \nonumber \\
    & \le 2\eta, \label{diffusion_tightness_bound}
\end{align}
where (\ref{diffusion_measurability}) follows from $U^{\gamma}(i \gamma)$ and $S^{\gamma}(i \gamma)$ being $\mathcal{H}_{i}^{\gamma}$-measurable, (\ref{diffusion_conditional_independence}) follows from conditional independence of $I_k^{\gamma}(i+1)$ and $(X_k^{\gamma}(i+1) - \mu_k^{\gamma})^2/(\sigma_k^{\gamma})^2$, and (\ref{diffusion_tightness_bound}) holds for $\gamma$ sufficiently close to zero, which ensures (\ref{tightness_sampling_probability_bound}) and $( \sigma_k^{\gamma}/\sigma_k )^2 \le 2$.

Additionally, we have
\begin{align}
    & \E\left[ \xi_k^{\gamma}(i+1)^2 \indic{\abs{\xi_k^{\gamma}(i+1)} > \epsilon/\sqrt{\gamma}} \indic{p_{k}^{\gamma}(i) \ge \delta} \right] \nonumber \\ 
    & = \E\left[ \frac{\indic{p_{k}^{\gamma}(i) \ge \delta}}{p_{k}^{\gamma}(i)} \E\left[ I_k^{\gamma}(i+1) \left( \frac{X_k^{\gamma}(i+1) - \mu_k^{\gamma}}{\sigma_k} \right)^2 \indic{\abs{\xi_k^{\gamma}(i+1)} > \epsilon/\sqrt{\gamma}} \, \bigg| \, \mathcal{H}_{i}^{\gamma} \right] \right] \nonumber \\
    & \le \frac{1}{\delta} \E\left[ \E\left[ \left( \frac{X_k^{\gamma}(i+1) - \mu_k^{\gamma}}{\sigma_k} \right)^2 \indic{\abs{\xi_k^{\gamma}(i+1)} > \epsilon/\sqrt{\gamma}} \, \bigg| \, \mathcal{H}_{i}^{\gamma} \right] \right] \nonumber \\
    & \le \frac{1}{\delta} \E\left[ \E\left[ \abs{ \frac{X_k^{\gamma}(i+1) - \mu_k^{\gamma}}{\sigma_k} }^{2+\alpha} \right]^{2/(2+\alpha)} \P\left(\abs{\xi_k^{\gamma}(i+1)} > \epsilon/\sqrt{\gamma} \, \Big| \, \mathcal{H}_{i}^{\gamma} \right)^{\alpha/(2+\alpha)} \right] \label{diffusion_holder} \\
    & \le \frac{C}{\delta} \E\left[ \P\left(\abs{\xi_k^{\gamma}(i+1)} > \epsilon/\sqrt{\gamma} \, \Big| \, \mathcal{H}_{i}^{\gamma} \right)^{\alpha/(2+\alpha)} \right], \label{diffusion_assumption0_lyapunov}
\end{align}
where (\ref{diffusion_holder}) follows from H\"older's inequality, and (\ref{diffusion_assumption0_lyapunov}) follows from (\ref{assumption0_lyapunov}) in Assumption \ref{assumption0}, with constant $C > 0$.
Furthermore, a.s.,
\begin{align}
    \P\left(\abs{\xi_k^{\gamma}(i+1)} > \epsilon/\sqrt{\gamma} \, \Big| \, \mathcal{H}_{i}^{\gamma} \right) & \le \frac{\gamma}{\epsilon^2} \frac{1}{p_{k}^{\gamma}(i)} \E\left[ I_k^{\gamma}(i+1) \left( \frac{X_k^{\gamma}(i+1) - \mu_k^{\gamma}}{\sigma_k^{\gamma}} \right)^2 \, \bigg| \, \mathcal{H}_{i}^{\gamma} \right] \left(\frac{\sigma_k^{\gamma}}{\sigma_k}\right)^2 \nonumber \\
    & = \frac{\gamma}{\epsilon^2} \left(\frac{\sigma_k^{\gamma}}{\sigma_k}\right)^2. \nonumber
\end{align}
So, by the bounded convergence theorem, the right side of (\ref{diffusion_assumption0_lyapunov}) converges to zero as $\gamma \downarrow 0$.

Therefore, from (\ref{diffusion_tightness_bound}) and (\ref{diffusion_assumption0_lyapunov}), we have
\begin{align}
    \limsup_{\gamma \downarrow 0} \sup_{0 \le i \le \lfloor t/\gamma \rfloor - 1} \E\left[ \xi_k^{\gamma}(i+1)^2 \indic{\abs{\xi_k^{\gamma}(i+1)} > \epsilon/\sqrt{\gamma}} \right] \le 2\eta, \nonumber
\end{align}
and sending $\eta \downarrow 0$ yields (\ref{diffusion_M2_sufficient}).
\halmos
\endproof

\subsection{Proofs for Stochastic ODE Approximation} \label{diffusion_rode}

In this section, we prove Theorem \ref{diffusion_thm2}, Theorem \ref{diffusion_thm_sde_sode_equivalence} and Proposition \ref{time_change_representation_conversion} from Section \ref{diffusion_mab_2}. 
Theorem \ref{diffusion_thm2} shows that in the small gap regime of Assumption \ref{assumption1} (with stationary rewards for each arm) and the reward stack model of reward feedback, the pre-limit processes converge weakly to the unique (in distribution) non-anticipative weak solution to the stochastic ODE, with the uniqueness ensured by Theorem \ref{diffusion_thm_sde_sode_equivalence}.
In Theorem \ref{diffusion_thm_sde_sode_equivalence}, we show that a non-anticipative solution to the stochastic ODE is also a solution to the corresponding SDE.
So, if the SDE has a unique strong solution, then the stochastic ODE must have a unique (in distribution) non-anticipative weak solution.
In Proposition \ref{time_change_representation_conversion}, we show the converse result that a solution to the SDE is also a non-anticipative solution to the corresponding stochastic ODE.

\vspace{2mm}
\indent \textit{Proof of Theorem \ref{diffusion_thm2}.} 
\vspace{0mm} \\
We start with the joint processes $(U^{\gamma},Z^{\gamma},M^{\gamma})$ with components defined in (\ref{stationary_partial_sums}), (\ref{diffusion_r_process1}) and (\ref{diffusion_discrete_ode3}).
From Assumption \ref{assumption1} and Remark \ref{rmk4}, the individual components of $Z^{\gamma}$ are tight in $D[0,\infty)$, and their weak limit points have continuous sample paths.
Also, the individual components of $U^{\gamma}$ and $M^{\gamma}$ are tight in $D[0,\infty)$ (which can be argued as in Lemma \ref{diffusion_lemma1}), and their weak limit points have continuous sample paths.
The marginal tightness results together with the sample path continuity of all weak limit points ensures that the joint processes $(U^{\gamma},Z^{\gamma},M^{\gamma})$ are tight in $D_{3K}[0,\infty)$.
(There is no need here to use Lemma \ref{diffusion_lemma5} to establish joint tightness.)
By tightness and Prohorov's Theorem, any subsequence of $(U^{\gamma},Z^{\gamma},M^{\gamma})$ has a further subsequence that converges weakly.
Consider any weakly convergent subsequence of $(U^{\gamma},Z^{\gamma},M^{\gamma})$, which for notational simplicity we still index using $\gamma$, and let $(U,Z,M)$ denote its weak limit.
From Assumption \ref{assumption1} and Remark \ref{rmk4}, we know that $Z$ is a standard $K$-dimensional Brownian motion (with respect to its own augmented natural filtration).
Moreover, as in the proof of Theorem \ref{diffusion_thm1}, $M$ is the $D_K[0,\infty)$ zero process.

By the continuity of function composition (Theorem 13.2.2 of \cite{whitt_2002}), since the individual components of $Z$ have continuous sample paths and those of $U$ have non-decreasing sample paths, we have by the CMT,
\begin{align}
(U^{\gamma},Z^{\gamma},M^{\gamma},Z^{\gamma} \circ U^{\gamma}) \Rightarrow (U,Z,M,Z \circ U), \label{diffusion_t0_rode}
\end{align}
in $D_{4K}[0,\infty)$, with $Z^\gamma \circ U^\gamma$ as defined in (\ref{diffusion_z_process1}) and $Z \circ U = (Z_k(U_k), \, k \in [K]) \in D_K[0,\infty)$.
Define the processes $A^{\gamma} = (A_k^{\gamma}, \, k \in [K])$ and $A = (A_k, \, k \in [K])$, where
\begin{align}
A_k^{\gamma}(t) & = p_k^{\gamma}(U^{\gamma}(t),Z^{\gamma} \circ U^{\gamma}(t)) \label{diffusion_g11} \\
A_k(t) & = p_k(U(t),Z \circ U(t)). \label{diffusion_g22}
\end{align}
Since $p_k^{\gamma}(u,s) \to p_k(u,s)$ as $\gamma \downarrow 0$ uniformly for $(u,s)$ in compact subsets of $[0,\infty)^K \times \mathbb{R}^K$, and $p_k(u,s)$ is continuous at all $(u,s) \in [0,\infty)^K \times \mathbb{R}^K$, by the Generalized CMT (Lemma \ref{diffusion_lemma0}) applied to the processes in (\ref{diffusion_g11})-(\ref{diffusion_g22}), we have from (\ref{diffusion_t0_rode}),
\begin{align}
(U^{\gamma},Z^{\gamma},M^{\gamma},A^{\gamma}) \Rightarrow (U,Z,M,A). \label{diffusion_t1_rode}
\end{align}
Additionally, define the processes $\widetilde{U}^{\gamma} = (\widetilde{U}_k^{\gamma}, \, k \in [K])$ and $\widetilde{U} = (\widetilde{U}_k, \, k \in [K])$, where
\begin{align}
\widetilde{U}_k^{\gamma}(t) & = \int_0^t A_k^{\gamma}(v) dv \nonumber \\
\widetilde{U}_k(t) & = \int_0^t A_k(v) dv. \label{diffusion_rtilde_rode}
\end{align}
Recall that
\begin{align*}
U^{\gamma}_k(t) & = \gamma \sum_{i=0}^{\lfloor t/\gamma \rfloor - 1} A_k^{\gamma}(i \gamma) + M^{\gamma}_k(t).
\end{align*}
For each $k \in [K]$, because $M^{\gamma}_k$ converges weakly to the $D[0,\infty)$ zero process and also
\begin{align*}
\sup_{t \ge 0} \abs{\gamma \sum_{i=0}^{\lfloor t/\gamma \rfloor - 1} A_k^{\gamma}(i \gamma) - \widetilde{U}_k^{\gamma}(t)} \le \gamma,
\end{align*}
we have for any $T > 0$,
\begin{align}
\sup_{0 \le t \le T} \abs{U_k^{\gamma}(t) - \widetilde{U}_k^{\gamma}(t)} \overset{\P}{\to} 0. \label{diffusion_t2_rode}
\end{align}
By the continuity of integration with respect to the Skorohod metric (Theorem 11.5.1 of \cite{whitt_2002}) and the CMT, we have from (\ref{diffusion_t1_rode}),
\begin{align}
(U^{\gamma},Z^{\gamma},\widetilde{U}^{\gamma}) \Rightarrow (U,Z,\widetilde{U}). \label{diffusion_t3_rode}
\end{align}
Together, (\ref{diffusion_t2_rode})-(\ref{diffusion_t3_rode}) yield
\begin{align*}
(U^{\gamma},Z^{\gamma},U^{\gamma}) \Rightarrow (U,Z,\widetilde{U}),
\end{align*}
and recalling the definition of $\widetilde{U}$ in (\ref{diffusion_rtilde_rode}) and (\ref{diffusion_g22}), we obtain:
\begin{align}
(U^{\gamma},Z^{\gamma}) \Rightarrow (U,Z), \nonumber
\end{align}
where $(U,Z)$ satisfies
\begin{align}
U_k(t) & = \int_0^t p_k(U(v),Z \circ U(v)) dv, \quad k \in [K]. \label{convergence2_for_Theorem3.4}
\end{align}
Moreover, by the continuity of function composition, we have
\begin{align}
(U^{\gamma}, Z^{\gamma} \circ U^{\gamma}, Z^{\gamma}) \Rightarrow (U, Z \circ U, Z). \label{convergence1_for_Theorem3.4}
\end{align}

Next, we show that for $Z$ and the functions $p_k$, $U$ is a non-anticipative solution (as in Definition \ref{sode_non-anticipative}) to the stochastic ODE in (\ref{convergence2_for_Theorem3.4}).
For this, we use Lemma \ref{ethier-kurtz-theorem}, which is stated after the current proof.
Lemma \ref{ethier-kurtz-theorem} is adapted from Theorem 3.4(a) from Chapter 6 of \cite{ethier_etal1986}.
Define the filtration $\mathcal{G}^\gamma = (\mathcal{G}^\gamma_u, \, u \in [0,\infty)^K)$ via
\begin{align}
    \mathcal{G}^\gamma_u = \mathcal{F}_u^\gamma \vee \sigma(\varphi_j, \, j=1,\dots,\lfloor \Sigma_k u_k / \gamma \rfloor), \label{f_gamma_filtration_with_randomization}
\end{align}
where the filtration $\mathcal{F}^\gamma = (\mathcal{F}_u^\gamma, \, u \in [0,\infty)^K)$ is as defined in (\ref{h_u_gamma}), and $\varphi_j$ is the exogenous iid random variable used at time $j$ by TS to randomly draw an arm (in the original discrete-time system).
Then, we can see that the $U^\gamma(t)$, $t \ge 0$ are $\mathcal{G}^\gamma_u$-stopping times, i.e., $\bigcap_{k \in [K]} \{ U_k^\gamma(t) \le u_k \} \in \mathcal{G}_u^\gamma$ for any $u \in [0,\infty)^K$.
Moreover, because the $\varphi_j$ TS randomization variables are independent of $\mathcal{F}^\gamma$, we have that (\ref{nonanticipative_fdd}) (from Assumption \ref{assumption1}) holds with $\mathcal{G}^\gamma_u$ in the place of $\mathcal{F}^\gamma_u$, i.e., for any continuous functions $f_k : \mathbb{R} \to \mathbb{R}$, $k \in [K]$ vanishing at infinity, and any $u \in [0,\infty)^K$, $v > 0$,
\begin{align}
    \lim_{\gamma \downarrow 0} \E\left[ \, \abs{ \, \E\left[ f_k(Z_k^\gamma(u_k + v)) \mid \mathcal{G}_u^\gamma \right] - \E\left[ f_k(Z_k^\gamma(u_k) + \sqrt{v} \mathcal{N}_k) \mid Z_k^\gamma(u_k) \right] \, } \, \right] = 0, \label{semigroup_convergence_1}
\end{align}
where for each $k \in [K]$, $\mathcal{N}_k$ is a standard Gaussian random variable that is independent of $Z_k^\gamma$.
From (\ref{semigroup_convergence_1}), using the independence between different components $Z_k^\gamma$, $k \in [K]$ and the boundedness of the functions $f_k$, $k \in [K]$, we have
\begin{align}
    \lim_{\gamma \downarrow 0} \E\Biggl[ \, \Biggl| \, \E\Biggl[ \prod_{k \in [K]} f_k(Z_k^\gamma(u_k + v)) \, \biggl| \, \mathcal{G}^\gamma_u \Biggr] - \prod_{k \in [K]} \E\Bigl[ f_k(Z_k^\gamma(u_k) + \sqrt{v} \mathcal{N}_k) \, \Bigl| \, Z_k^\gamma(u_k) \Bigr] \, \Biggl| \, \Biggr] = 0. \label{semigroup_convergence_2}
\end{align}
Moreover, using (\ref{diffusion_t2_rode}) and the convergence $p_k^{\gamma}(u,s) \to p_k(u,s)$ as $\gamma \downarrow 0$ uniformly for $(u,s)$ in compact subsets of $[0,\infty)^K \times \mathbb{R}^K$, we have
\begin{align}
    \abs{U_k^\gamma(t) - \int_0^t p_k(U^\gamma(v), Z^\gamma \circ U^\gamma(v)) dv} \overset{\P}{\to} 0, \label{u_convergence}
\end{align}
for each $t \ge 0$ as $\gamma \downarrow 0$.
With (\ref{semigroup_convergence_2}), (\ref{u_convergence}) and (\ref{convergence1_for_Theorem3.4}), the corresponding conditions (\ref{Brownian_semigroup_convergence}), (\ref{stopping_time_convergence}) and (\ref{joint_3m_weak_convergence}) from Lemma \ref{ethier-kurtz-theorem} are verified.
Also, the functions $p_k : [0,\infty)^K \times \mathbb{R}^K \to (0,1)$ are continuous.
Thus, Lemma \ref{ethier-kurtz-theorem} establishes for $Z$ and the functions $p_k$, the desired non-anticipative property of $U$.

Finally, we can apply Theorem \ref{diffusion_thm_sde_sode_equivalence} to conclude that all such weakly converging subsequences $(U^\gamma, Z^\gamma \circ U^\gamma)$ have the same weak limit.
Specifically, by Theorem \ref{diffusion_thm_sde_sode_equivalence}, with $(U,Z \circ U)$ denoting the weak limit (as in (\ref{convergence1_for_Theorem3.4}), which satisfies the stochastic ODE in (\ref{convergence2_for_Theorem3.4})), we have that $(U,S) := (U,Z \circ U)$ is also a solution to the corresponding SDE, which has a unique strong solution.
This establishes the desired weak convergence in Theorem \ref{diffusion_thm2} to the unique (in distribution) non-anticipative weak solution to the stochastic ODE.
\halmos

\begin{lemma}[Adapted from Theorem 3.4(a), Chapter 6 of \cite{ethier_etal1986}] \label{ethier-kurtz-theorem}
For $n \ge 1$, let $W^n = (W_k^n, \, k \in [m]) \in D_m[0,\infty)$ be a sequence of processes.
For each $n$, let $\eta^n(t) = (\eta_k^n(t), \, k \in [m])$, $t \ge 0$ be a component-wise non-decreasing family of $\mathcal{G}_u^n$-stopping times that is right-continuous in $t$, where $\mathcal{G}^n = (\mathcal{G}_u^n, \, u \in [0,\infty)^m)$ is a filtration that satisfies $\mathcal{G}_u^n \supset \sigma(W_k^n(t_k), \, t_k \le u_k, \, k \in [m])$.
Suppose that for any continuous functions $f_k : \mathbb{R} \to \mathbb{R}$, $k \in [m]$ vanishing at infinity, and any $u \in [0,\infty)^m$, $v > 0$,
\begin{align}
    \lim_{n \to \infty} \E\Biggl[ \, \Biggl| \, \E\Biggl[ \prod_{k \in [m]} f_k(W_k^n(u_k + v)) \, \Biggl| \, \mathcal{G}^n_u \Biggr] - \prod_{k \in [m]} \E\Bigl[ f_k(W_k^n(u_k) + \sqrt{v} \mathcal{N}_k) \, \Bigl| \, W_k^n(u_k) \Bigr] \, \Biggl| \, \Biggr] = 0, \label{Brownian_semigroup_convergence}
\end{align}
where for each $k \in [m]$, $\mathcal{N}_k$ is a standard Gaussian random variable that is independent of $W_k^n$.
Suppose also that for each $k \in [m]$, there exists a continuous function $\beta_k : [0,\infty)^m \times \mathbb{R}^m \to (0,1)$ such that
\begin{align}
    \abs{\eta_k^n(t) - \int_0^t \beta_k(\eta^n(v), W^n \circ \eta^n(v)) dv} \overset{\P}{\to} 0, \label{stopping_time_convergence}
\end{align}
for each $t \ge 0$ as $n \to \infty$.
Additionally, suppose that
\begin{align}
    (\eta^n, W^n \circ \eta^n, W^n) \Rightarrow (\eta,W \circ \eta, W) \label{joint_3m_weak_convergence}
\end{align}
in $D_{3m}[0,\infty)$ as $n \to \infty$.
Then, $W = (W_k, \, k \in [m])$ is a standard $m$-dimensional Brownian motion, and for $W$ and the functions $\beta_k$, $\eta = (\eta_k, \, k \in [m])$ is a non-anticipative solution (as in Definition \ref{sode_non-anticipative}) to the stochastic ODE:
\begin{align}
    \eta_k(t) = \int_0^t \beta_k(\eta(v), W \circ \eta(v)) dv, \quad k \in [m]. \nonumber
\end{align}
\end{lemma}

\indent \textit{Proof of Theorem \ref{diffusion_thm_sde_sode_equivalence}.}
\vspace{0mm} \\
Let $B$ be a standard $K$-dimensional Brownian motion on a probability space $(\Omega,\mathbb{F},\P)$, and let $\mathcal{F} = (\mathcal{F}_u, \, u \in [0,\infty)^K)$ be the augmented filtration defined in (\ref{explicit_augmented_Brownian_filtration}).
Let $U$ be a non-anticipative solution, as in Definition \ref{sode_non-anticipative}, to the stochastic ODE:
\begin{align}
    U_k(t) = \int_0^t p_k(U(v), B \circ U(v)) dv, \quad k \in [K], \label{ode_start}
\end{align}
with the standard $K$-dimensional Brownian motion $B$.
So, there exists a filtration $\mathcal{G} = (\mathcal{G}_u, \, u \in [0,\infty)^K)$ for which conditions (i)-(iii) of Definition \ref{sode_non-anticipative} hold.
For any $\theta \in \mathbb{R}^K$, define for $u \in [0,\infty)^K$:
\begin{align}
    \psi_\theta(u) = \prod_{k \in [K]} \exp\left( i \theta_k B_k(u_k) + \frac{1}{2} \theta_k^2 u_k \right). \nonumber
\end{align}
By condition (ii) of Definition \ref{sode_non-anticipative}, for any $\theta \in \mathbb{R}^K$, $\psi_\theta(u)$ is a $\mathcal{G}_u$-martingale.
(See Definition \ref{directed_set_martingale}, given after the current proof, for martingales indexed by directed sets like $[0,\infty)^K$.)
Moreover, by condition (iii) of Definition \ref{sode_non-anticipative}, for each $t \ge 0$, $U(t)$ is a $\mathcal{G}_u$-stopping time.
Thus, the conditions of Theorem 6.3(a) of \cite{kurtz_etal1980b} are satisfied.
Define $S = (S_k, \, k \in [K])$ by $S_k(t) := B_k(U_k(t))$.
Using Theorem 6.3(a), for any $k,l \in [K]$ with $k \ne l$, both $S_k(t)$ and $S_k(t) S_l(t)$ are continuous local martingales with respect to the filtration $\widetilde{\mathcal{G}} = (\widetilde{\mathcal{G}}_t, \, t \ge 0)$ defined by $\widetilde{\mathcal{G}}_t := \mathcal{G}_{U(t)}$.
Using the fact that $U_k(t) \le t$ a.s., together with Doob's Maximal Inequality (Theorem 3.8(iv) from Chapter 1 of \cite{karatzas_etal1998}), we have for each $k \in [K]$,
\begin{align}
    \E\left[\sup_{v \in [0,t]} S_k^2(v)\right] \le \E\left[\left(\sup_{v \in [0,t]} \abs{B_k(v)} \right)^2\right] \le 4 \E[B_k^2(t)] = 4t. \label{square_integrability}
\end{align}
From (\ref{square_integrability}), we can conclude that the $S_k(t)$ are continuous square-integrable martingales and the $S_k(t) S_l(t)$ for $k \ne l$ are continuous martingales with respect to the filtration $\widetilde{\mathcal{G}}_t$.
Since the $S_k(t) S_l(t)$ for $k \ne l$ are continuous martingales, by Theorem 5.13 from Chapter 1 of \cite{karatzas_etal1998}, their quadratic co-variations $\langle S_k, S_l \rangle_t$ must be zero.
Also, by Lemma \ref{diffusion_lemma4}, developed after the current proof,
\begin{align}
    \langle S_k, S_k \rangle_t = U_k(t). \label{time_changed_quadratic_variation}
\end{align} 
So, in summary, for $k,l \in [K]$,
\begin{align}
    \langle S_k, S_l \rangle_t & = U_k(t) \indic{k = l}. \label{s_quadratic_covariation}
\end{align}
Moreover, note that 
\begin{align}
    d \langle S_k, S_k \rangle_t = d U_k(t) = p_k(U(t),S(t)) dt. \label{integrator_quadratic_variation}
\end{align}

Now, define $\widetilde{B} = (\widetilde{B}_k, \, k \in [K])$ by
\begin{align}
    \widetilde{B}_k(t) := \int_0^t \frac{1}{\sqrt{p_k(U(v),S(v))}} dS_k(v), \quad k \in [K]. \label{constructed_brownian_motion_1}
\end{align}
The stochastic integral in (\ref{constructed_brownian_motion_1}) is well defined for each $k \in [K]$ because the integrand $1/\sqrt{p_k(U(t),S(t))}$ is progressively measurable with respect to $\widetilde{\mathcal{G}}_t$ (since $U(t)$, $S(t)$ are continuous and adapted to $\widetilde{\mathcal{G}}_t$ and thus progressively measurable, and the functions $p_k$ are positive and Borel measurable), the integrator $S_k(t)$ is a continuous square-integrable martingale with respect to $\widetilde{\mathcal{G}}_t$, and using (\ref{integrator_quadratic_variation}), we have
\begin{align}
    \E\left[ \int_0^t \frac{1}{p_k(U(v),S(v))} d \langle S_k, S_k \rangle_v \right] = t < \infty. \nonumber
\end{align}
For each $k \in [K]$, $\widetilde{B}_k(t)$ is a continuous square-integrable martingale with respect to $\widetilde{\mathcal{G}}_t$ (Definition 2.9 from Chapter 3 of \cite{karatzas_etal1998}).
Moreover, for any $k,l \in [K]$, by Proposition 2.17 from Chapter 3 of \cite{karatzas_etal1998},
\begin{align}
    \langle \widetilde{B}_k, \widetilde{B}_l \rangle_t = \int_0^t \frac{1}{\sqrt{p_k(U(v),S(v))}} \frac{1}{\sqrt{p_l(U(v),S(v))}} d \langle S_k, S_l \rangle_v. \nonumber
\end{align}
Using (\ref{s_quadratic_covariation}) and (\ref{integrator_quadratic_variation}), we have $\langle \widetilde{B}_k, \widetilde{B}_l \rangle_t = t \indic{k = l}$. 
So, by the L\'{e}vy characterization of multi-dimensional Brownian motion (Theorem 3.16 from Chapter 3 of \cite{karatzas_etal1998}), $\widetilde{B}$ is a standard $K$-dimensional Brownian motion with respect to the filtration $\widetilde{\mathcal{G}}_t$. 
Moreover, from (\ref{constructed_brownian_motion_1}) and Corollary 2.20 from Chapter 3 of \cite{karatzas_etal1998}, we have for each $k \in [K]$,
\begin{align}
    \int_0^t \sqrt{p_k(U(v),S(v))} d\widetilde{B}_k(v) = \int_0^t \sqrt{p_k(U(v),S(v))} \frac{1}{\sqrt{p_k(U(v),S(v))}} dS_k(v) = S_k(t). \nonumber
\end{align}
Thus, we have the representation:
\begin{align}
    U_k(t) & = \int_0^t p_k(U(v), S(v)) dv \label{sde_recovery_1} \\
    S_k(t) & = \int_0^t \sqrt{p_k(U(v),S(v))} d\widetilde{B}_k(v), \quad k \in [K]. \label{sde_recovery_2}
\end{align}
So, the non-anticipative solution $U$ to the stochastic ODE in (\ref{ode_start}) with the Brownian motion $B$, together with $S = B \circ U$, solves the SDE in (\ref{sde_recovery_1})-(\ref{sde_recovery_2}) with the Brownian motion $\widetilde{B}$.
\halmos

\vspace{2mm}

\begin{definition} \label{directed_set_martingale}
    On a probability space $(\Omega, \mathbb{F}, \mathbb{P})$, let $\mathcal{F} = (\mathcal{F}_u, \, u \in [0,\infty)^m)$ be a filtration indexed by (directed set) $[0,\infty)^m$.
    A real-valued process $\xi$ indexed by $[0,\infty)^m$ is an \textit{$\mathcal{F}_u$-martingale} if $\xi(u)$ is $\mathcal{F}_u$-measurable for each $u \in [0,\infty)^m$, $\E[\abs{\xi(u)}] < \infty$ for each $u \in [0,\infty)^m$, and $\E[\xi(v) \mid \mathcal{F}_u] = \xi(u)$ for all $u, v \in [0,\infty)^m$ such that $u_k \le v_k$ for all $k \in [m]$.
\end{definition}

\vspace{2mm}

\begin{lemma} \label{diffusion_lemma4}
    In the proof of Theorem \ref{diffusion_thm_sde_sode_equivalence}, (\ref{time_changed_quadratic_variation}) holds, i.e., $\langle S_k, S_k \rangle_t = U_k(t)$ for each $k \in [K]$.
\end{lemma}

\indent \textit{Proof of Lemma \ref{diffusion_lemma4}.}
\vspace{0mm} \\ 
    Recall from the proof of Theorem \ref{diffusion_thm_sde_sode_equivalence} that for each $k \in [K]$, $S_k(t) = B_k(U_k(t))$ (by definition) and $B_k(U_k(t))$ is a continuous square-integrable martingale with respect to the filtration $\widetilde{\mathcal{G}}_t := \mathcal{G}_{U(t)}$.
    For each $k$, $B_k(u_k)^2 - u_k$, with index taken to be $u \in [0,\infty)^K$ (not just $u_k \in [0,\infty)$), is a continuous $\mathcal{G}_u$-martingale (recall Definition \ref{directed_set_martingale} above).
    Also, for $0 \le s \le t$, we have that $U(s)$ and $U(t)$ are bounded $\mathcal{G}_u$-stopping times with $0 \le U_k(s) \le U_k(t) \le t$ for all $k$.
    Then, for each $k$, we have
    \begin{align}
        \E[ B_k(U_k(t))^2 - U_k(t) \mid \mathcal{G}_{U(s)} ] = B_k(U_k(s))^2 - U_k(s), \nonumber
    \end{align}
    using Theorem 8.7 from Chapter 2 of \cite{ethier_etal1986}, which is an optional stopping theorem for filtrations and martingales indexed by directed sets like $[0,\infty)^K$ (recall Remark \ref{kurtz_results} and Definition \ref{directed_set_filtration_stopping_time}). 
    So, for each $k$, $B_k(U_k(t))^2 - U_k(t)$ is a $\widetilde{\mathcal{G}}_t$ martingale.
    The desired result then follows from the Doob-Meyer decomposition (Definition 5.3 from Chapter 1 of \cite{karatzas_etal1998}).
\halmos

\vspace{2mm}

\indent \textit{Proof of Proposition \ref{time_change_representation_conversion}.}
\vspace{0mm} \\ 
Let $\widetilde{B}$ be a standard $K$-dimensional Brownian motion on a probability space $(\Omega,\mathbb{F},\P)$, and let $\widetilde{\mathcal{F}} = (\widetilde{\mathcal{F}}_t, \, t \ge 0)$ be the augmented natural filtration corresponding to $\widetilde{B}$.
Let $(U,S)$ be a strong solution to the SDE (\ref{sde_sode_equivalence_1})-(\ref{sde_sode_equivalence_3}) on this probability space with respect to the standard Brownian motion $\widetilde{B}$.
Writing (\ref{sde_sode_equivalence_2}) in integral form, because the $p_k$ functions are bounded, we have that
\begin{align*}
S_k(t) = \int_0^t \sqrt{p_k(U(v),S(v))} d\widetilde{B}_k(v), \quad k \in [K]
\end{align*}
are continuous $\widetilde{\mathcal{F}}_t$-martingales with quadratic variation processes
\begin{align*}
\langle S_k, S_k \rangle_t & = \int_0^t p_k(U(v),S(v)) dv, \quad k \in [K].
\end{align*}
For $k \ne l$, the quadratic co-variation processes $\langle S_k, S_l \rangle_t = 0$ since $\widetilde{B}_k$ and $\widetilde{B}_l$ are independent.
From (\ref{sde_sode_equivalence_1}), we see that $\langle S_k, S_k \rangle_t = U_k(t)$, $k \in [K]$, which are continuous and strictly increasing processes since the $p_k$ functions are bounded and strictly positive.
Define for each $k \in [K]$:
\begin{align*}
U_k^{-1}(t) = \inf\{v \ge 0 : U_k(v) > t\}.
\end{align*}
Also, we define $U_k(\infty) = \lim_{t \to \infty} U_k(t)$, and if $U_k(\infty) < \infty$, then we define $S_k(\infty) := \lim_{t \uparrow U_k(\infty)} S_k(U_k^{-1}(t))$.

By a possible extension of the original probability space $(\Omega, \mathbb{F}, \mathbb{P})$, we have independent standard Brownian motions $W_k$, $k \in [K]$ that are also independent of $\widetilde{\mathcal{F}}$.
For $k \in [K]$, define $B = (B_k, \, k \in [K])$ via:
\begin{align}
    B_k(t) = \begin{cases}
        S_k(U_k^{-1}(t)), & t < U_k(\infty) \\
        S_k(\infty) + W_k(t - U_k(\infty)), & t \ge U_k(\infty).
    \end{cases} \nonumber
\end{align} 
(This appends on an additional independent standard Brownian motion if $U_k(\infty) < \infty$.)
By construction, we have $B_k(U_k(t)) = S_k(t)$, and substituting this representation into the SDE in (\ref{sde_sode_equivalence_1}), we obtain the stochastic ODE representation:
\begin{align}
U_k(t) = \int_0^t p_k(U(v),B \circ U(v)) dv, \quad k \in [K]. \label{diffusion_equivalent_ode}
\end{align}

We now establish that $B$ is a standard $K$-dimensional Brownian motion, and that $U$ is a non-anticipative (as in Definition \ref{sode_non-anticipative}) solution to the stochastic ODE in (\ref{diffusion_equivalent_ode}).
Define the filtration $\mathcal{G} = (\mathcal{G}_u, \, u \in [0,\infty)^K)$ via
\begin{align}
    \mathcal{G}_u = \mathcal{F}_u \vee \bigcap_{k \in [K]} \widetilde{\mathcal{F}}_{U^{-1}_k(u_k)}, \nonumber
\end{align}
where the filtration $\mathcal{F} = (\mathcal{F}_u, \, u \in [0,\infty)^K)$ is defined via
\begin{align}
    \mathcal{F}_u = \sigma\bigl( B_k(t_k), \, t_k \le u_k, \, k \in [K]\bigr) \vee \sigma(\mathcal{L}), \nonumber
\end{align}
with $\mathcal{L} \subset \mathbb{F}$ denoting the collection of all $\mathbb{P}$-probability zero sets.
Since $\mathcal{F}_u \subset \mathcal{G}_u$ for all $u \in [0,\infty)^K$, condition (i) of Definition \ref{sode_non-anticipative} is satisfied.
Moreover, for any $\theta \in \mathbb{R}^K$, define for $u \in [0,\infty)^K$:
\begin{align}
    \psi_\theta(u) = \prod_{k \in [K]} \exp\left( i \theta_k B_k(u_k) + \frac{1}{2} \theta_k^2 u_k \right). \nonumber
\end{align}
By Theorem 6.3(b) of \cite{kurtz_etal1980b}, for any $\theta \in \mathbb{R}^K$, $\psi_\theta(u)$ is a $\mathcal{G}_u$-martingale.
(Recall Definition \ref{directed_set_martingale} for martingales indexed by directed sets like $[0,\infty)^K$.)
Also, for each $k \in [K]$, $B_k(t)$ has continuous sample paths.
Thus, $B$ is a standard $K$-dimensional Brownian motion.
Moreover, condition (ii) of Definition \ref{sode_non-anticipative} is satisfied, i.e., with $\xi^u(\cdot) = (B_k(u_k + \cdot), \, k \in [K])$ for $u \in [0,\infty)^K$, we have for all $u \in [0,\infty)^K$,
\begin{align}
    \P(\xi^u \in \cdot \mid \mathcal{G}_u) = \P(\xi^u \in \cdot \mid \mathcal{F}_u). \nonumber
\end{align}
Lastly, for any $t \ge 0$, condition (iii) of Definition \ref{sode_non-anticipative} follows from:
\begin{align}
    \bigcap_{k \in [K]} \{ U_k(t) \le u_k \} = \Bigl\{ \min_{k \in [K]} U_k^{-1}(u_k) \ge t \Bigr\} \in \bigcap_{k \in [K]} \widetilde{\mathcal{F}}_{U^{-1}_k(u_k)} \subset \mathcal{G}_u. \nonumber
\end{align}
\halmos

\subsection{Proofs for Algorithmic Invariance Principle} \label{sec:sections_2_4_proofs}

\proof{Proof of Theorem \ref{diffusion_prop3}.}
Here, we establish Theorem \ref{diffusion_prop3} for the stochastic ODE limit using the reward stack model of reward feedback.
The SDE limit using the random table model of reward feedback follows from the stochastic ODE limit and Theorem \ref{diffusion_thm_sde_sode_equivalence}, together with the distributional equivalence in the pre-limit between the random table and reward stack models under Assumption \ref{assumption0} with iid rewards.

Under $\epsilon$-warm-start (from Definition \ref{def:epsilon-warm-start}), with fixed, positive sampling probabilities $q_k$, $k \in [K]$, the weak convergence of $(U^\gamma, Z^\gamma \circ U^\gamma)$ over the time range $[0,\epsilon]$ is straightforward to establish using the arguments from the proof of Theorem \ref{diffusion_thm2}.
So, we focus on establishing the weak convergence $(U^\gamma, Z^\gamma \circ U^\gamma)$ over the time range $[\epsilon,\infty)$.
We verify that the sampling probabilities for EF Thompson samplers have the desired form with $p_k(u,s)$ as in (\ref{well-specified_sampling_probability}).

At time $j+1$ with $j \ge \lfloor \epsilon/\gamma \rfloor$, conditional on $\mathcal{H}^{\gamma}_j$ (as defined in (\ref{h_j_gamma})), for each arm $k \in [K]$, we sample once from the posterior distribution of $\mu_k^{\gamma}$ and denote the sample by $\widetilde{\mu}_k^{\gamma}(j+1)$.
For each arm $k \in [K]$, let $\widehat{\mu}_k^{\gamma}(j+1)$ denote the sample mean estimate at time $j+1$.
(For the exponential family model, the sample mean is the maximum likelihood estimator (MLE) for the mean, and is used as the centering quantity for the Gaussian posterior approximation in Proposition \ref{diffusion_prop1}.)
Recall that the $U_k^{\gamma}$ and $Z_k^{\gamma}(U_k^{\gamma})$ processes have the expressions from (\ref{diffusion_r_process1}) and (\ref{diffusion_z_process1}).
Note that here, $\sigma_k = \sigma_*$ for all $k \in [K]$.
Then, the probability of playing arm $k \in [K]$ is given by:
\begin{align}
& \P\bigl( k = \argmax_{l \in [K]} \; \widetilde{\mu}_l^{\gamma}(j+1) \; \big| \; \mathcal{H}^{\gamma}_j \bigr) \nonumber \\
& = \P\left( k = \argmax_{l \in [K]} \left\{ \frac{Z_l^{\gamma}(U_l^{\gamma}(j \gamma)) \sigma_*}{U_l^{\gamma}(j \gamma)} + d_l^{\gamma} + \frac{1}{\sqrt{\gamma}} \left(\widetilde{\mu}_l^{\gamma}(j+1) - \widehat{\mu}_l^{\gamma}(j+1)\right) \right\} \; \bigg| \; U^{\gamma}(j \gamma), Z^{\gamma} \circ U^{\gamma}(j \gamma) \right) \nonumber \\
& = \P\left( k = \argmax_{l \in [K]} \left\{ \frac{Z_l^{\gamma}(U_l^{\gamma}(j \gamma)) \sigma_*}{U_l^{\gamma}(j \gamma)} + d_l^{\gamma} + \frac{\sigma_l^{\gamma}}{\sqrt{U_l^{\gamma}(j \gamma)}} \mathcal{N}_l \right\} \; \bigg| \; U^{\gamma}(j \gamma), Z^{\gamma} \circ U^{\gamma}(j \gamma) \right) + \widetilde{r}_k^\gamma(j) \label{diffusion_BvM_approximation1} \\
& = p_k^{\gamma}(U^{\gamma}(j \gamma), Z^{\gamma} \circ U^{\gamma}(j \gamma)) + \widetilde{r}_k^\gamma(j), \label{diffusion_BvM_approximation2}
\end{align}
where the probability in (\ref{diffusion_BvM_approximation1}) is taken with respect to independent standard Gaussian random variables $\mathcal{N}_l$, and the $p_k^{\gamma}(u,s)$ in (\ref{diffusion_BvM_approximation2}) have the expression:
\begin{align}
    p_k^{\gamma}(u,s) = \P\left( k = \argmax_{l \in [K]} \left\{ \frac{s_{l} \sigma_*}{u_{l}} + d_l^{\gamma} + \frac{\sigma_l^{\gamma}}{\sqrt{u_{l}}} \mathcal{N}_l \right\} \right). \label{general_pre-limit_sampling_probability_epsilon-warm-start}
\end{align}

We show that in (\ref{diffusion_BvM_approximation1}), the $\widetilde{r}_k^\gamma(j)$ terms satisfy
\begin{align}
    \max_{k \in [K]} \sup_{j \ge \lfloor \epsilon/\gamma \rfloor} \abs{\widetilde{r}_k^\gamma(j)} \overset{\P}{\to} 0 \label{remainder_term_uniform_control_bvm}
\end{align}
as $\gamma \downarrow 0$.
To see this, first note that for each arm $k \in [K]$, the $\epsilon$-warm-start sample size $\sim \lfloor \epsilon / \gamma \rfloor q_k$ a.s. as $\gamma \downarrow 0$, with $q_k > 0$ fixed (and $\sum_k q_k = 1$).
In particular, a.s. for each $k \in [K]$,
\begin{align}
    \lim_{\gamma \downarrow 0} U_k^\gamma(\epsilon) = \epsilon q_k = U_k(\epsilon). \nonumber
\end{align}
(We can couple the random variables used for randomization during the $\epsilon$-warm-start across all $\gamma > 0$ to obtain the a.s. convergence result as $\gamma \downarrow 0$.)
Moreover, the joint posterior distribution for all $K$ arms is the product of the $K$ marginal posterior distributions (in accordance with the fact that each arm reward distribution is learned separately in the multi-armed bandit model), and the posterior approximation in Proposition \ref{diffusion_prop1} applies jointly for all $K$ arms in this way.
Then, (\ref{remainder_term_uniform_control_bvm}) follows from Proposition \ref{diffusion_prop1}, which develops convergence using total variation distance and ensures that the Gaussian approximations are valid for all measurable events.

Additionally, with $p_k(u,s)$ as in (\ref{well-specified_sampling_probability}), we have that $p_k^{\gamma}(u,s) \to p_k(u,s)$ uniformly for $(u,s)$ in compact subsets of $[\epsilon q_1,\infty) \times \dots \times [\epsilon q_K,\infty) \times \mathbb{R}^K$.
The restriction $u_k \ge \epsilon q_{k} > 0$ for each $k \in [K]$ is due to the initial sampling with constant, positive probabilities $(q_k, \, k \in [K])$ in the $\epsilon$-warm-start procedure.

This sequence of derivations parallels that of (\ref{diffusion_prob11})-(\ref{diffusion_prob13}) in Section \ref{diffusion_mab_2}. 
Continuing from (\ref{diffusion_BvM_approximation2}), the arguments from the proof of Theorem \ref{diffusion_thm2} can be applied to yield the desired stochastic ODE approximation over the time range $[\epsilon,\infty)$.
In particular, (\ref{remainder_term_uniform_control_bvm}) ensures
\begin{align}
    \max_{k \in [K]} \gamma \sum_{j = \lfloor \epsilon/\gamma \rfloor}^{\lfloor t/\gamma \rfloor} \abs{\widetilde{r}_k^{\gamma}(j)} \overset{\P}{\to} 0 \nonumber
\end{align}
for any $t > \epsilon > 0$ as $\gamma \downarrow 0$.
\halmos
\endproof

\vspace{2mm}

\proof{Proof of Theorem \ref{diffusion_prop5}.}
The proof is similar to the proof of Theorem \ref{diffusion_prop3}.
The difference is that we are now using the non-parametric bootstrap sampler instead of the EF Thompson sampler.

Here, we establish Theorem \ref{diffusion_prop5} for the stochastic ODE limit using the reward stack model of reward feedback.
The SDE limit using the random table model of reward feedback follows from the stochastic ODE limit and Theorem \ref{diffusion_thm_sde_sode_equivalence}, together with the distributional equivalence in the pre-limit between the random table and reward stack models under Assumption \ref{assumption0} with iid rewards.

Under $\epsilon$-warm-start (from Definition \ref{def:epsilon-warm-start}), with fixed, positive sampling probabilities $q_k$, $k \in [K]$, the weak convergence of $(U^\gamma, Z^\gamma \circ U^\gamma)$ over the time range $[0,\epsilon]$ is straightforward to establish using the arguments from the proof of Theorem \ref{diffusion_thm2}.
So, we focus on establishing the weak convergence of $(U^\gamma, Z^\gamma \circ U^\gamma)$ over the time range $[\epsilon,\infty)$.
We verify that the sampling probabilities for the non-parametric bootstrap sampler have the desired form with $p_k(u,s)$ as in (\ref{well-specified_sampling_probability}).

At time $j+1$ with $j \ge \lfloor \epsilon/\gamma \rfloor$, conditional on $\mathcal{H}^{\gamma}_j$ (as defined in (\ref{h_j_gamma})), for each arm $k \in [K]$, we compute one bootstrap sample mean $\mu_k^{*\gamma}(j+1)$.
For each arm $k \in [K]$, let $\widehat{\mu}_k^{\gamma}(j+1)$ denote the sample mean estimate at time $j+1$.
Recall that the $U_k^{\gamma}$ and $Z_k^{\gamma}(U_k^{\gamma})$ processes have the expressions from (\ref{diffusion_r_process1}) and (\ref{diffusion_z_process1}).
Note that here, $\sigma_k = \sigma_*$ for all $k \in [K]$.
Then, as in the proof of Theorem \ref{diffusion_prop3}, the probability of playing arm $k \in [K]$ is given by:
\begin{align}
& \P\bigl( k = \argmax_{l \in [K]} \; \mu_l^{*\gamma}(j+1) \; \big| \; \mathcal{H}^{\gamma}_j \bigr) \nonumber \\
& = \P\left( k = \argmax_{l \in [K]} \left\{ \frac{Z_l^{\gamma}(U_l^{\gamma}(j \gamma)) \sigma_*}{U_l^{\gamma}(j \gamma)} + d_l^{\gamma} + \frac{1}{\sqrt{\gamma}} \left(\mu_l^{*\gamma}(j+1) - \widehat{\mu}_l^{\gamma}(j+1)\right) \right\} \; \bigg| \; \mathcal{H}^{\gamma}_j \right) \nonumber \\
& = \P\left( k = \argmax_{l \in [K]} \left\{ \frac{Z_l^{\gamma}(U_l^{\gamma}(j \gamma)) \sigma_*}{U_l^{\gamma}(j \gamma)} + d_l^{\gamma} + \frac{\sigma_l^{\gamma}}{\sqrt{U_l^{\gamma}(j \gamma)}} \mathcal{N}_l \right\} \; \bigg| \; U^{\gamma}(j \gamma), Z^{\gamma} \circ U^{\gamma}(j \gamma) \right) + r_k^{*\gamma}(j) \label{diffusion_BS_approximation1} \\
& = p_k^{\gamma}(U^{\gamma}(j \gamma), Z^{\gamma} \circ U^{\gamma}(j \gamma)) + r_k^{*\gamma}(j), \label{diffusion_BS_approximation2}
\end{align}
where the probability in (\ref{diffusion_BS_approximation1}) is taken with respect to independent standard Gaussian random variables $\mathcal{N}_l$, and the $p_k^{\gamma}(u,s)$ in (\ref{diffusion_BS_approximation2}) have the expression in (\ref{general_pre-limit_sampling_probability_epsilon-warm-start}).

We show that in (\ref{diffusion_BS_approximation1}), the $r_k^{*\gamma}(j)$ terms satisfy
\begin{align}
    \max_{k \in [K]} \sup_{j \ge \lfloor \epsilon/\gamma \rfloor} \abs{r_k^{*\gamma}(j)} \overset{\P}{\to} 0 \label{remainder_term_uniform_control_bootstrap}
\end{align}
as $\gamma \downarrow 0$.
To see this, first note that for each arm $k \in [K]$, the $\epsilon$-warm-start sample size $\sim \lfloor \epsilon / \gamma \rfloor q_k$ a.s. as $\gamma \downarrow 0$, with $q_k > 0$ fixed (and $\sum_k q_k = 1$).
In particular, a.s. for each $k \in [K]$,
\begin{align}
    \lim_{\gamma \downarrow 0} U_k^\gamma(\epsilon) = \epsilon q_k = U_k(\epsilon). \nonumber
\end{align}
(We can couple the random variables used for randomization during the $\epsilon$-warm-start across all $\gamma > 0$ to obtain the a.s. convergence result as $\gamma \downarrow 0$.)
Moreover, the joint bootstrap distribution for all $K$ arms is the product of the $K$ marginal bootstrap distributions (in accordance with the fact that each arm reward distribution is learned separately in the multi-armed bandit model), and the bootstrap approximation in Proposition \ref{diffusion_prop2} applies jointly for all $K$ arms in this way.
As shorthand, denote for $k \in [K]$, $j \ge \lfloor \epsilon/\gamma \rfloor$, and $x \in \mathbb{R}$:
\begin{align}
    \xi_k^{\gamma,j} & = \frac{Z_k^{\gamma}(U_k^{\gamma}(j \gamma)) \sigma_*}{U_k^{\gamma}(j \gamma)} + d_k^{\gamma} \nonumber \\
    F_k^{\gamma,j}(x) & = \P\left( \xi_k^{\gamma,j} + \frac{1}{\sqrt{\gamma}} \left(\mu_k^{*\gamma}(j+1) - \widehat{\mu}_k^{\gamma}(j+1)\right) \le x \; \bigg| \; \mathcal{H}^{\gamma}_j \right) \nonumber \\
    G_k^{\gamma,j}(x) & = \Phi\left( (x - \xi_k^{\gamma,j}) \frac{\sqrt{U_k^{\gamma}(j \gamma)}}{\sigma_k^\gamma} \right) \nonumber \\
    \delta_k^\gamma(j) & = \sup_{x \in \mathbb{R}} \abs{ F_k^{\gamma,j}(x) - G_k^{\gamma,j}(x) }, \nonumber
\end{align}
where $\Phi(\cdot)$ is the standard Gaussian CDF.
Then, by Proposition \ref{diffusion_prop2},
\begin{align}
    \max_{k \in [K]} \sup_{j \ge \lfloor \epsilon/\gamma \rfloor} \delta_k^{\gamma}(j) \overset{\P}{\to} 0 \label{weak_convergence_bootstrap}
\end{align}
as $\gamma \downarrow 0$.
From (\ref{diffusion_BS_approximation1}), we have for each $k \in [K]$,
\begin{align}
    r_k^{*\gamma}(j) \le \int_{\mathbb{R}} \left( \prod_{l \ne k} F_l^{\gamma,j}(x) \right) dF_k^{\gamma,j}(x) - \int_{\mathbb{R}} \left( \prod_{l \ne k} G_l^{\gamma,j}(x) \right) dG_k^{\gamma,j}(x), \label{bootstrap_upper_bound} 
\end{align}
and
\begin{align}
    r_k^{*\gamma}(j) \ge \int_{\mathbb{R}} \left( \prod_{l \ne k} F_l^{\gamma,j}(x-) \right) dF_k^{\gamma,j}(x) - \int_{\mathbb{R}} \left( \prod_{l \ne k} G_l^{\gamma,j}(x) \right) dG_k^{\gamma,j}(x). \label{bootstrap_lower_bound} 
\end{align}
The distinction between (\ref{bootstrap_upper_bound}) and (\ref{bootstrap_lower_bound}) is that (\ref{bootstrap_upper_bound}) involves the terms $F_l^{\gamma,j}(x)$, whereas (\ref{bootstrap_lower_bound}) involves the terms (left limits) $F_l^{\gamma,j}(x-)$.
No matter how ties are broken for the non-parametric bootstrap sampler, (\ref{bootstrap_upper_bound}) and (\ref{bootstrap_lower_bound}) provide upper and lower bounds for $r_k^{*\gamma}(j)$. 

The right side of (\ref{bootstrap_upper_bound}) is equal to:
\begin{align}
    \int_{\mathbb{R}} \left( \prod_{l \ne k} F_l^{\gamma,j}(x) - \prod_{l \ne k} G_l^{\gamma,j}(x) \right) dF_k^{\gamma,j}(x) + \int_{\mathbb{R}} \left( \prod_{l \ne k} G_l^{\gamma,j}(x) \right) d(F_k^{\gamma,j}(x) - G_k^{\gamma,j}(x)). \label{integral_difference_1}
\end{align}
The magnitude of the first integral in (\ref{integral_difference_1}) can be upper bounded by $\sum_{l \ne k} \delta_l^{\gamma}(j)$.
For the second integral in (\ref{integral_difference_1}), via integration by parts,
\begin{align}
    & \int_{\mathbb{R}} \left( \prod_{l \ne k} G_l^{\gamma,j}(x) \right) d(F_k^{\gamma,j}(x) - G_k^{\gamma,j}(x)) \nonumber \\
    & = \left( \prod_{l \ne k} G_l^{\gamma,j}(x) \right) ( F_k^{\gamma,j}(x) - G_k^{\gamma,j}(x) ) \Bigl|_{-\infty}^\infty - \int_{\mathbb{R}} ( F_k^{\gamma,j}(x) - G_k^{\gamma,j}(x) ) d\prod_{l \ne k} G_l^{\gamma,j}(x). \label{integral_difference_2}
\end{align}
The magnitude of the right side of (\ref{integral_difference_2}) can be upper bounded by $2 \delta_k^{\gamma}(j)$.
So, the magnitude of the right side of (\ref{bootstrap_upper_bound}) can be upper bounded by $2 K \max_{k \in [K]} \delta_k^{\gamma}(j)$.
Similarly, the magnitude of the right side of (\ref{bootstrap_lower_bound}) can also be upper bounded by $2 K \max_{k \in [K]} \delta_k^{\gamma}(j)$.
Thus,
\begin{align}
    \max_{k \in [K]} \sup_{j \ge \lfloor \epsilon/\gamma \rfloor} \abs{r_k^{*\gamma}(j)} \le 2 K \max_{k \in [K]} \sup_{j \ge \lfloor \epsilon/\gamma \rfloor} \delta_k^{\gamma}(j), \nonumber
\end{align}
and using (\ref{weak_convergence_bootstrap}), we have established (\ref{remainder_term_uniform_control_bootstrap}). 

Additionally, with $p_k(u,s)$ as in (\ref{well-specified_sampling_probability}), we have that $p_k^{\gamma}(u,s) \to p_k(u,s)$ uniformly for $(u,s)$ in compact subsets of $[\epsilon q_1,\infty) \times \dots \times [\epsilon q_K,\infty) \times \mathbb{R}^K$.
The restriction $u_k \ge \epsilon q_{k} > 0$ for each $k \in [K]$ is due to the initial sampling with constant, positive probabilities $(q_k, \, k \in [K])$ in the $\epsilon$-warm-start procedure.

This sequence of derivations parallels that of (\ref{diffusion_prob11})-(\ref{diffusion_prob13}) in Section \ref{diffusion_mab_2}. 
Continuing from (\ref{diffusion_BS_approximation2}), the arguments from the proof of Theorem \ref{diffusion_thm2} can be applied to yield the desired stochastic ODE approximation over the time range $[\epsilon,\infty)$.
In particular, (\ref{remainder_term_uniform_control_bootstrap}) ensures
\begin{align}
    \max_{k \in [K]} \gamma \sum_{j = \lfloor \epsilon/\gamma \rfloor}^{\lfloor t/\gamma \rfloor} \abs{r_k^{*\gamma}(j)} \overset{\P}{\to} 0 \nonumber
\end{align}
for any $t > \epsilon > 0$ as $\gamma \downarrow 0$.
\halmos
\endproof

%\THEEndNotes
%\begingroup \parindent 0pt \parskip 0.0ex \def\enotesize{\normalsize} \theendnotes \endgroup

% Appendix here
% Options are (1) APPENDIX (with or without general title) or
%             (2) APPENDICES (if it has more than one unrelated sections)
% Outcomment the appropriate case if necessary
%
% \begin{APPENDIX}{<Title of the Appendix>}
% \end{APPENDIX}
%
%   or
%
% \begin{APPENDICES}
% \section{<Title of Section A>}
% \section{<Title of Section B>}
% etc
% \end{APPENDICES}

\begin{APPENDICES}

\section{Gaussian Approximations for Posterior Distributions} \label{diffusion_appA1}

In this appendix, we consider the same setup as in Section \ref{diffusion_general}. 
Recall that the arm reward distributions are from an exponential family $P^\mu$ parameterized by mean $\mu$ (as expressed in (\ref{diffusion_exponential_family})), with means $\mu$ known to belong to a bounded, open interval $\mathcal{I}$.
Our goal here is to develop Proposition \ref{diffusion_prop1} below, which is a custom version of the BvM theorem.
This version establishes weak convergence of the rescaled posterior distribution to a Gaussian distribution, a.s. as the sample size $n \to \infty$, and uniformly over the possible data-generating distributions $P^\mu$, $\mu \in \mathcal{I}$.
The reason we develop the result uniformly over the possible data-generating distributions is the following.
For a time horizon of $O(1/\gamma)$, the small gap regime of Assumption \ref{assumption0} involves mean parameters $\mu_k^\gamma$ in a $\sqrt{\gamma}$-neighborhood of some $\mu_*$.
As $\gamma \downarrow 0$ (and the time horizon goes to infinity), the mean parameters $\mu_k^\gamma$ change.
So, for a given large time horizon, the Gaussian approximation to the posterior in Proposition \ref{diffusion_prop1} should be valid simultaneously for all distributions $P^\mu$ corresponding to a range of mean parameters $\mu$; a fixed (not depending on $\gamma$) neighborhood of $\mu_*$ suffices.
Below, we first discuss the ``uniform almost sure'' mode of convergence, and then move on to the development of Proposition \ref{diffusion_prop1}.

\vspace{2mm}
\noindent \textbf{``Uniform Almost Sure'' Convergence}

To make sense of the ``uniform almost sure'' mode of convergence, we first recall an equivalent characterization of almost sure convergence in Remark \ref{rmk2} below, followed by a precise definition of the mode of convergence in Definition \ref{def1} below.
For any particular distribution $Q$, we use $\E_Q[\cdot]$ and $\P_Q(\cdot)$ to denote expectation and probability taken with respect to $Q$.

\begin{remark} \label{rmk2}
    For a sequence of random variables $Y_1,Y_2,\dots$,
    \begin{align}
        Y_n \overset{\textnormal{a.s.}}{\to} 0 \nonumber
    \end{align}
    as $n \to \infty$, if and only if for any $\epsilon > 0$,
    \begin{align}
        \lim_{n \to \infty} \P\left( \sup_{j \ge n} \abs{Y_j} > \epsilon \right) = 0. \nonumber
    \end{align}
\end{remark}

\begin{definition} \label{def1}
    Let $\mathcal{Q}$ be a collection of probability distributions and $Z_i$ be random variables defined on the probability spaces $(\Omega,\mathbb{F},Q)_{Q \in \mathcal{Q}}$.
    We say that the sequence $Z_i$ converges to zero, a.s. uniformly in $Q \in \mathcal{Q}$, if for any $\epsilon > 0$,
    \begin{align}
        \lim_{n \to \infty} \sup_{Q \in \mathcal{Q}} \P_Q\left( \sup_{j \ge n} \abs{Z_j} > \epsilon \right) = 0. \nonumber
    \end{align}
\end{definition}

Next, we state Lemma \ref{chung_uniform_slln}, which is used in the proof of Proposition \ref{diffusion_prop1}.
This result, originally due to \cite{chung_1951}, is a strong law of large numbers (SLLN) that holds uniformly over a collection of underlying probability distributions.

\begin{lemma} \label{chung_uniform_slln}
    Let $\mathcal{Q}$ be a collection of probability distributions, and for each $Q \in \mathcal{Q}$, let $Y, Y_i \overset{\textnormal{iid}}{\sim} Q$.
    Suppose the $\mathcal{Q}$-uniform integrability condition,
    \begin{align}
        \lim_{z \to \infty} \sup_{Q \in \mathcal{Q}} \E_Q\left[ \abs{Y - \E_Q[Y]} \indic{\abs{Y - \E_Q[Y]} > z} \right] = 0, \nonumber
    \end{align}
    is satisfied.
    Then, for every $\epsilon > 0$,
    \begin{align}
        \lim_{m \to \infty} \sup_{Q \in \mathcal{Q}} \P_Q\left( \sup_{n \ge m} \abs{\frac{1}{n} \sum_{i=1}^n Y_i - \E_Q[Y]} > \epsilon \right) = 0. \nonumber
    \end{align}
\end{lemma}

\vspace{2mm}
\noindent \textbf{Development of Proposition \ref{diffusion_prop1}}

Before presenting Lemma \ref{parametric_glivenko_cantelli} and then continuing on to Proposition \ref{diffusion_prop1}, which is the main result of this appendix, we first formalize the (modest) technical conditions, \hyperref[diffusion_C1]{C1} and \hyperref[diffusion_C2]{C2} below, that are used to develop these results.
It can be easily verified that the exponential family setup detailed in (\ref{diffusion_exponential_family}) from Section \ref{diffusion_general} satisfies \hyperref[diffusion_C1]{C1} and \hyperref[diffusion_C2]{C2}.
Notation-wise, corresponding to the exponential family $P^\mu$, the log-likelihood function is denoted by $l(\mu,x)$, and derivatives of $l(\mu,x)$ with respect to $\mu$ are denoted by $l'(\mu,x)$, $l''(\mu,x)$, etc.
Recall from Section \ref{diffusion_general} that $\mathcal{I}$ is a bounded, open interval containing $\mu_*$, satisfying $\underline{\mu} < \inf \mathcal{I} < \sup \mathcal{I} < \overline{\mu}$.
(For the exponential family $P^\mu$, recall that $(\underline{\mu},\overline{\mu})$ denotes the open interval of all possible mean values achievable with finite values of the tilting parameter.)
For each $\mu \in \mathcal{I}$, let $X^\mu, X_i^\mu \overset{\text{iid}}{\sim} P^\mu$.

\begin{enumerate}
\item[(C1)] \label{diffusion_C1} For each $\delta > 0$, there is an $\epsilon(\delta) > 0$ such that for all $\mu \in \mathcal{I}$,
\begin{align}
\sup_{z : \abs{\mu - z} \ge \delta} \E[l(z,X^\mu)] \le \E[l(\mu,X^\mu)] - \epsilon(\delta). \label{diffusion_C1.1}
\end{align}
\item[(C2)] \label{diffusion_C2} 
There exist functions $\eta$ and $\kappa$ such that for all $x$ in the support of the base distribution $P$ (in (\ref{diffusion_exponential_family})),
\begin{align}
& \eta(x) \ge \sup_{\mu \in \mathcal{I}} \abs{l'(\mu,x)} \label{diffusion_C2.1} \\
& \kappa(x) \ge \sup_{\mu \in \mathcal{I}} \abs{l'''(\mu,x)}. \label{diffusion_C2.2}
\end{align}
\hspace{4.5mm} Moreover, for the cases: $f(x) = \abs{x}$, $f(x) = \eta(x) + \abs{l(\mu_0,x)}$ for some fixed $\mu_0 \in \mathcal{I}$, and $f(x) = \kappa(x)$,
\begin{align}
\lim_{y \to \infty} \sup_{\mu \in \mathcal{I}} \E\left[ f(X^\mu) \indic{f(X^\mu) > y} \right] = 0. \label{diffusion_C2.3}
\end{align}
\end{enumerate}

Applying Theorems 2.7.11 and 2.8.1 of \cite{vandervaart_etal1996} (and using the mean value theorem), we have the following result. 
\begin{lemma} \label{parametric_glivenko_cantelli}
    Suppose (\ref{diffusion_C2.3}) in \hyperref[diffusion_C2]{C2} holds for the case $f(x) = \eta(x) + \abs{l(\mu_0,x)}$ with some fixed $\mu_0 \in \mathcal{I}$ and $\eta(x)$ as defined in (\ref{diffusion_C2.1}).
    Then, $\{l(\mu,\cdot), \mu \in \mathcal{I}\}$ is a Glivenko-Cantelli class of functions uniformly in $P^\mu$, $\mu \in \mathcal{I}$, i.e., for any $\epsilon > 0$,
    \begin{align}
        \lim_{m \to \infty} \sup_{\mu \in \mathcal{I}} \P\left( \sup_{n \ge m} \sup_{z \in \mathcal{I}} \abs{ \frac{1}{n} \sum_{i = 1}^n l(z,X_i^\mu) - \E[l(z,X^\mu)] } > \epsilon \right) = 0. \nonumber
    \end{align}
\end{lemma}

We now state and prove Proposition \ref{diffusion_prop1}. 
The proof is adapted from the proof sketch for Theorem 4.2 in \cite{ghosh_etal2006}.
As before, for each $\mu \in \mathcal{I}$, let $X^\mu, X_i^\mu \overset{\text{iid}}{\sim} P^\mu$, with mean $\mu$ and corresponding variance $(\sigma^\mu)^2$.
(Below, we will write all relevant quantities with superscript $\mu$ to keep track of the distribution $P^\mu$ that we work with.)
The sample mean of $X_1^\mu,\dots,X_n^\mu$ is denoted by $\widehat{m}_n^\mu$.
Given $n$ such samples, we use $\widetilde{m}_n^\mu$ to denote a sample from the posterior distribution of the mean $\mu$.

\begin{proposition} \label{diffusion_prop1}
For the exponential family $P^\mu$ in (\ref{diffusion_exponential_family}) from Section \ref{diffusion_general}, suppose the conditions \hyperref[diffusion_C1]{C1} and \hyperref[diffusion_C2]{C2} hold for a bounded, open interval $\mathcal{I} \subset \mathbb{R}$.
Let $\mathcal{J}$ be a compact sub-interval of $\mathcal{I}$.
Let $\nu_0$ be a bounded prior density that has support contained in $\mathcal{I}$, is continuous on an open interval containing $\mathcal{J}$, and is strictly positive on $\mathcal{J}$.
Then, conditional on the data $X_i^\mu \overset{\textnormal{iid}}{\sim} P^\mu$, the centered and scaled posterior density $y \mapsto \nu_n(y \mid X_1^\mu,\dots,X_n^\mu)$ for $\sqrt{n}(\widetilde{m}_n^\mu - \widehat{m}_n^\mu)$ satisfies:
\begin{align}
\lim_{n \to \infty} \int_{\mathbb{R}} \abs{\nu_n(y \mid X_1^\mu,\dots,X_n^\mu) - \frac{1}{\sqrt{2\pi}\sigma^\mu} \exp\left(-\frac{1}{2 (\sigma^\mu)^2}y^2 \right)} dy = 0, \label{diffusion_p1e0}
\end{align}
a.s. uniformly in the underlying distribution $P^\mu$ for $\mu \in \mathcal{J}$, in the sense of Definition \ref{def1}.
\end{proposition}

\proof{Proof of Proposition \ref{diffusion_prop1}.}
The posterior density can be expressed as
\begin{align}
\nu_n(y \mid X_1^\mu,\dots,X_n^\mu) = (C_n^\mu)^{-1} \nu_0( \widehat{m}_n^\mu + y/\sqrt{n} ) \exp\left(L_{n}^\mu(\widehat{m}_n^\mu + y/\sqrt{n}) - L_{n}^\mu(\widehat{m}_n^\mu)\right), \label{diffusion_p1e1}
\end{align}
with normalization factor 
\begin{align}
    C_n^\mu = \int_{\mathbb{R}} \nu_0( \widehat{m}_n^\mu + y/\sqrt{n} ) \exp\left(L_{n}^\mu(\widehat{m}_n^\mu + y/\sqrt{n}) - L_{n}^\mu(\widehat{m}_n^\mu)\right) dy, \nonumber
\end{align}
and
\begin{align}
L_{n}^\mu(z) = \sum_{i=1}^n l(z,X_i^\mu). \nonumber
\end{align}
Consider the following difference between unnormalized densities.
\begin{align}
D_n^\mu(y) = \nu_0( \widehat{m}_n^\mu + y/\sqrt{n} ) \exp\left(L_{n}^\mu(\widehat{m}_n^\mu + y/\sqrt{n}) - L_{n}^\mu(\widehat{m}_n^\mu)\right) - \nu_0(\mu) \exp\left(-\frac{1}{2 (\sigma^\mu)^2}y^2 \right) \label{diffusion_p1e2}
\end{align}
To establish (\ref{diffusion_p1e0}), it suffices to show that
\begin{align}
\lim_{n \to \infty} \int_{\mathbb{R}} \abs{D_n^\mu(y)} dy = 0, \label{diffusion_p1e3}
\end{align}
a.s. uniformly in $\mu \in \mathcal{J}$ (i.e., a.s. uniformly in the underlying distribution $P^\mu$ for $\mu \in \mathcal{J}$, in the sense of Definition \ref{def1}).
Indeed, if (\ref{diffusion_p1e3}) holds, we must also have
\begin{align}
\lim_{n \to \infty} C_n^\mu = \nu_0(\mu) \sqrt{2\pi} \sigma^\mu, \label{normalization_factor_convergence}
\end{align}
a.s. uniformly in $\mu \in \mathcal{J}$.
Then, we would have
\begin{align}
& \int_{\mathbb{R}} \abs{\nu_n(y \mid X_1^\mu,\dots,X_n^\mu) - \frac{1}{\sqrt{2\pi} \sigma^\mu} \exp\left(-\frac{1}{2 (\sigma^\mu)^2}y^2 \right)} dy \nonumber \\
& \le (C_n^\mu)^{-1} \int_{\mathbb{R}} \abs{D_n^\mu(y)} dy + \abs{(C_n^\mu)^{-1} \nu_0(\mu) - \frac{1}{\sqrt{2\pi} \sigma^\mu}} \int_{\mathbb{R}} \exp\left(-\frac{1}{2 (\sigma^\mu)^2} y^2 \right) dy. \label{convergence_conclusion}
\end{align}
Applying (\ref{diffusion_p1e3}) and (\ref{normalization_factor_convergence}) to (\ref{convergence_conclusion}) would then lead to the desired conclusion in (\ref{diffusion_p1e0}). 
So, it suffices to show that (\ref{diffusion_p1e3}) holds.
To do this, we split the integral over $\mathbb{R}$ into two pieces on $A_n = \{ y : \abs{y} > \beta \sqrt{n} \}$ and $A_n^c = \{ y : \abs{y} \le \beta \sqrt{n} \}$, with $\beta > 0$ to be specified later in the proof.

In the first case on $A_n$, we have
\begin{align}
\int_{A_n} \abs{D_n^\mu(y)} dy & \le \int_{A_n} \nu_0(\widehat{m}_n^\mu + y/\sqrt{n}) \exp\left(L_{n}^\mu(\widehat{m}_n^\mu + y/\sqrt{n}) - L_{n}^\mu(\widehat{m}_n^\mu)\right) dy \nonumber \\
& \quad + \int_{A_n} \nu_0(\mu) \exp\left(-\frac{1}{2 (\sigma^\mu)^2} y^2 \right) dy. \label{diffusion_p1e4}
\end{align}
By the boundedness of $\nu_0(\mu)$ and $(\sigma^\mu)^2$ for $\mu \in \mathcal{J}$, the second integral on the right side of (\ref{diffusion_p1e4}) goes to zero as $n \to \infty$, uniformly in $\mu \in \mathcal{J}$.
For the first integral on the right side of (\ref{diffusion_p1e4}), using (\ref{diffusion_C2.3}) in condition \hyperref[diffusion_C2]{C2} for the case $f(x) = \abs{x}$ together with Lemma \ref{chung_uniform_slln}, it follows that
\begin{align}
    \lim_{n \to \infty} \widehat{m}_n^\mu = \mu, \label{diffusion_slln_uniform}
\end{align}
a.s. uniformly in $\mu \in \mathcal{J}$.
This, together with condition \hyperref[diffusion_C1]{C1} and Lemma \ref{parametric_glivenko_cantelli}, implies that there exists $\epsilon(\beta) > 0$ such that
\begin{align}
\sup_{y \in A_n \, : \, (\widehat{m}_n^\mu + y/\sqrt{n}) \in \mathcal{I}} \, \frac{1}{n} \left(L_{n}^\mu(\widehat{m}_n^\mu + y/\sqrt{n}) - L_{n}^\mu(\widehat{m}_n^\mu)\right) \le -\epsilon(\beta), \label{A_n_bound}
\end{align}
for sufficiently large $n$, a.s. uniformly in $\mu \in \mathcal{J}$.
(For the first integral on the right side of (\ref{diffusion_p1e4}), we only need to consider $y$ for which $(\widehat{m}_n^\mu + y/\sqrt{n}) \in \mathcal{I}$ because the prior density $\nu_0$ has support contained in $\mathcal{I}$.)
So, using (\ref{diffusion_p1e4}), (\ref{A_n_bound}) and the boundedness of the density $\nu_0$, we have
\begin{align}
    \lim_{n \to \infty} \int_{A_n} \abs{D_n^\mu(y)} dy = 0, \label{final_uniform_almost_sure_1}
\end{align}
a.s. uniformly in $\mu \in \mathcal{J}$. 
(Note that this result does not depend on how the constant $\beta > 0$ will be chosen below.)

For the second case on $A_n^c$, we analyze $\int_{A_n^c} \abs{D_n^\mu(y)} dy$.
We expand $L_{n}^\mu$ in a Taylor series about the MLE $\widehat{m}_n^\mu$, noting that by the definition of the MLE, $(L_{n}^\mu)'(\widehat{m}_n^\mu) = 0$.
We have
\begin{align}
L_{n}^\mu(\widehat{m}_n^\mu + y/\sqrt{n}) - L_{n}^\mu(\widehat{m}_n^\mu) & = \frac{1}{2} \frac{1}{n} (L_{n}^\mu)''(\widehat{m}_n^\mu) y^2 + r_n^\mu(y) \nonumber \\
& = \frac{1}{2} \left( \theta''(\widehat{m}_n^\mu) \widehat{m}_n^\mu - \Lambda''(\widehat{m}_n^\mu) \right) y^2 + r_n^\mu(y), \label{diffusion_p1e5}
\end{align}
using the fact that $l''(z,x) = \theta''(z) \cdot x - \Lambda''(z)$ (recall the definitions of $\theta(z)$ and $\Lambda(z)$ in (\ref{diffusion_exponential_family})), with 
\begin{align*}
r_n^\mu(y) = \frac{1}{6} \left( \frac{y}{\sqrt{n}} \right)^3 (L_{n}^\mu)'''(m_{n,y}^\mu),
\end{align*}
where $m_{n,y}^\mu$ is a point in between $\widehat{m}_n^\mu$ and $\widehat{m}_n^\mu + y/\sqrt{n}$.
Here, we require that $\beta > 0$ satisfy:
\begin{align}
    [\min \mathcal{J} - 2 \beta, \max \mathcal{J} + 2 \beta] \subset \mathcal{I}, \label{choose_beta_relation_1}
\end{align}
so that $m_{n,y}^\mu \in \mathcal{I}$ for $y \in A_n^c$, for $n$ sufficiently large, a.s. uniformly in $\mu \in \mathcal{J}$.
Then, using condition \hyperref[diffusion_C2]{C2} with $f(x) = \kappa(x)$ and Lemma \ref{chung_uniform_slln}, there exists $\delta > 0$ such that for sufficiently large $n$, a.s. uniformly in $\mu \in \mathcal{J}$,
\begin{align}
\abs{r_n^\mu(y)} \le \frac{1}{6} \frac{\abs{y}^3}{\sqrt{n}} \frac{1}{n} \sum_{i=1}^n \kappa(X_i^\mu) \le \frac{1}{6} \frac{\abs{y}^3}{\sqrt{n}} \left( \E\left[\kappa(X^\mu)\right] + \delta \right). \label{diffusion_p1e6}
\end{align}
For $y \in A_n^c$, (\ref{diffusion_p1e6}) can be re-expressed as
\begin{align}
\abs{r_n^\mu(y)} \le \frac{1}{6} \beta y^2 \left( \E\left[\kappa(X^\mu)\right] + \delta \right). \label{diffusion_p1e7}
\end{align}
For the first term on the right side of (\ref{diffusion_p1e5}), we have
\begin{align}
    \lim_{n \to \infty} \theta''(\widehat{m}_n^\mu) \widehat{m}_n^\mu - \Lambda''(\widehat{m}_n^\mu) = \theta''(\mu) \mu - \Lambda''(\mu) = - \frac{1}{(\sigma^\mu)^2} \label{diffusion_observed_fisher_info}
\end{align}
a.s. uniformly in $\mu \in \mathcal{J}$.
The first equality in (\ref{diffusion_observed_fisher_info}) follows from the uniform continuity of $\theta''$ and $\Lambda''$ on $\mathcal{I}$, together with the convergence result in (\ref{diffusion_slln_uniform}).
The second equality in (\ref{diffusion_observed_fisher_info}) is from a standard identity relating Fisher information and the variance $(\sigma^\mu)^2$ of the $X_i^\mu$. 
Defining 
\begin{align}
    c_0 = \inf_{\mu \in \mathcal{J}} \frac{1}{(\sigma^\mu)^2}, \label{c_0_definition}
\end{align}
and recognizing that $c_0 > 0$, we can choose $\beta > 0$ to satisfy both (\ref{choose_beta_relation_1}) and also:
\begin{align}
    -\frac{1}{3} c_0 + \frac{1}{6} \beta \left( \, \sup_{\mu \in \mathcal{J}} \E\left[\kappa(X^\mu)\right] + \delta \right) \le -\frac{1}{4} c_0. \label{choose_beta_relation_2}
\end{align}
Then, using (\ref{diffusion_p1e7}) and (\ref{diffusion_observed_fisher_info}), we have from (\ref{diffusion_p1e5}) that
\begin{align}
\sup_{y \in A_n^c} \frac{\exp\left(L_{n}^\mu(\widehat{m}_n^\mu + y/\sqrt{n}) - L_{n}^\mu(\widehat{m}_n^\mu)\right)}{\exp( - c_0 y^2 / 4)} \le 1 \label{diffusion_p1e8}
\end{align}
for sufficiently large $n$, a.s. uniformly in $\mu \in \mathcal{J}$.
(Note that using $-(1/3)c_0$ instead of $-(1/2)c_0$ on the left side of (\ref{choose_beta_relation_2}) offers some ``slack'' to ensure compatibility with the convergence in (\ref{diffusion_observed_fisher_info}).)
Thus, with $D_n^\mu(y)$ as defined in (\ref{diffusion_p1e2}), we have using (\ref{diffusion_p1e5}), (\ref{diffusion_observed_fisher_info}), (\ref{c_0_definition}) and (\ref{diffusion_p1e8}),
\begin{align}
    \abs{D_n^\mu(y)} \indic{y \in A_n^c} \le \left( \sup_{z \in \mathcal{I}} \nu_0(z) \right) \Bigl( \exp( - c_0 y^2 / 4) + \exp( - c_0 y^2 / 2) \Bigr), \label{domination_independent_of_mu}
\end{align}
for sufficiently large $n$, a.s. uniformly in $\mu \in \mathcal{J}$.
Let $g(y)$ denote the function on the right side of (\ref{domination_independent_of_mu}).
Let $\epsilon > 0$.
Then, there exists $L(\epsilon) > 0$ such that
\begin{align}
    \int_{\abs{y} > L(\epsilon)} g(y) dy \le \epsilon/2. \label{integral_upper_bound_half_epsilon}
\end{align}
Moreover, using the continuity of $\nu_0$ on an open interval containing $\mathcal{J}$, (\ref{diffusion_slln_uniform}), (\ref{diffusion_p1e5}), (\ref{diffusion_p1e6}) and (\ref{diffusion_observed_fisher_info}), we have
\begin{align}
    \lim_{n \to \infty} \sup_{\abs{z} \le L(\epsilon)} \abs{D_n^\mu(z)} = 0, \label{uniform_uniform_almost_sure}
\end{align}
a.s. uniformly in $\mu \in \mathcal{J}$.
From (\ref{domination_independent_of_mu}) and (\ref{integral_upper_bound_half_epsilon}), we have
\begin{align}
    \int_{y \in A_n^c} \abs{D_n^\mu(y)} dy & \le 2L(\epsilon) \sup_{\abs{z} \le L(\epsilon)} \abs{D_n^\mu(z)} + \int_{\abs{y} > L(\epsilon)} g(y) dy \nonumber \\
    & \le 2L(\epsilon) \sup_{\abs{z} \le L(\epsilon)} \abs{D_n^\mu(z)} + \epsilon/2. \label{decomposition_uniform_almost_sure}
\end{align}
Then, using (\ref{uniform_uniform_almost_sure}) and (\ref{decomposition_uniform_almost_sure}), we have
\begin{align}
    \limsup_{m \to \infty} \sup_{\mu \in \mathcal{J}} \P_{P^\mu}\left( \sup_{n \ge m} \int_{y \in A_n^c} \abs{D_n^\mu(y)} dy > \epsilon  \right) & \le \limsup_{m \to \infty} \sup_{\mu \in \mathcal{J}} \P_{P^\mu}\left( \sup_{n \ge m} 2L(\epsilon) \sup_{\abs{z} \le L(\epsilon)} \abs{D_n^\mu(z)} > \epsilon/2  \right) \nonumber \\
    & = 0, \nonumber
\end{align}
thus establishing that
\begin{align}
    \lim_{n \to \infty} \int_{y \in A_n^c} \abs{D_n^\mu(y)} dy = 0, \label{final_uniform_almost_sure_2}
\end{align}
a.s. uniformly in $\mu \in \mathcal{J}$.
Together, (\ref{final_uniform_almost_sure_1}) and (\ref{final_uniform_almost_sure_2}) yield (\ref{diffusion_p1e3}).
\halmos
\endproof

\section{Gaussian Approximations for the Bootstrap} \label{diffusion_appA2}

In Proposition \ref{diffusion_prop2} below, we develop a Gaussian approximation for bootstrapping the sample mean.
Recall that here, we allow for an arbitrary family (not necessarily from an exponential family) of reward distributions $P^\mu$ parameterized by mean $\mu \in \mathcal{I}$, with corresponding variances $(\sigma^\mu)^2$, where $\mathcal{I} \subset \mathbb{R}$ is a bounded, open interval.
The only requirement on the $P^\mu$ is that the condition in (\ref{diffusion_second_moment_uniform_integrable}) is satisfied.
As before, for each $\mu \in \mathcal{I}$, let $X^\mu, X_i^\mu \overset{\text{iid}}{\sim} P^\mu$.
We use $\widehat{m}_n^\mu$ and $(\widehat{\sigma}_n^\mu)^2$ to denote the sample mean and variance computed using $n$ samples $X_1^\mu,\dots,X_n^\mu$.
Also, we use $\widehat{m}_n^{*\mu}$ to denote a bootstrap of the sample mean $\widehat{m}_n^\mu$ computed using $n$ re-samples with replacement.
Proposition \ref{diffusion_prop2} holds a.s. and uniformly over data-generating distributions $P^\mu$, $\mu \in \mathcal{I}$. 
Recall that a precise description of this mode of convergence was given in Remark \ref{rmk2} and Definition \ref{def1} in Appendix \ref{diffusion_appA1}.

\begin{proposition} \label{diffusion_prop2}
Suppose that
\begin{align}
    \lim_{y \to \infty} \sup_{\mu \in \mathcal{I}} \E\left[(X^\mu)^2 \indic{(X^\mu)^2 > y}\right] = 0. \label{diffusion_second_moment_uniform_integrable}
\end{align}
and
\begin{align}
    \inf_{\mu \in \mathcal{I}} \sigma^\mu > 0. \label{uniform_variance_lower_bound_2}
\end{align}
Then,
\begin{align}
\lim_{n \to \infty} \sup_{x \in \mathbb{R}} \abs{\P\left(\sqrt{n}\left( \widehat{m}_n^{*\mu} - \widehat{m}_n^\mu \right) \le x \mid X_1^\mu,\dots,X_n^\mu \right) - \Phi\left(\frac{x}{\sigma^\mu}\right) } = 0, \label{diffusion_bootstrap_clt}
\end{align}
a.s. uniformly in $\mu \in \mathcal{I}$.
\end{proposition}

\proof{Proof of Proposition \ref{diffusion_prop2}.}
Using (\ref{diffusion_second_moment_uniform_integrable}) and Lemma \ref{chung_uniform_slln}, we have 
\begin{align}
    & \lim_{n \to \infty} \widehat{m}_n^\mu = \mu \label{diffusion_slln_uniform_2} \\
    & \lim_{n \to \infty} \widehat{\sigma}_n^\mu = \sigma^\mu, \label{diffusion_slln_uniform_3}
\end{align}
a.s. uniformly in $\mu \in \mathcal{I}$.
The rest of the proof adapts the approach for establishing bootstrap consistency described in Example 3.1 from \cite{shao_etal1995}.
By the Berry-Esseen Theorem (Appendix A.9 of \cite{shao_etal1995}), there exists some universal constant $c > 0$ such that for all $n$, a.s. for all $\mu \in \mathcal{I}$,
\begin{align}
    \sup_{x \in \mathbb{R}} \abs{\P\left(\sqrt{n}\left( \widehat{m}_n^{*\mu} - \widehat{m}_n^\mu \right) \le x \mid X_1^\mu,\dots,X_n^\mu \right) - \Phi\left( \frac{x}{\widehat{\sigma}_n^\mu} \right) } \le c (\widehat{\sigma}_n^\mu)^{-3} n^{-3/2} \sum_{i=1}^n \abs{ X_i^\mu - \widehat{m}_n^\mu }^3. \label{berry_esseen_uniform_bound}
\end{align}
For the right side of (\ref{berry_esseen_uniform_bound}), we have
\begin{align}
    \abs{X_i^\mu - \widehat{m}_n^\mu}^3 \le 4 \left( \abs{X_i^\mu - \mu}^3 + \abs{\widehat{m}_n^\mu - \mu}^3 \right), \nonumber
\end{align}
and moreover,
\begin{align}
    & \lim_{n \to \infty} \frac{1}{n^{3/2}} \sum_{i=1}^n \abs{X_i^\mu - \mu}^3 = 0 \label{M-Z-SLLN} \\
    & \lim_{n \to \infty} \frac{1}{n^{3/2}} \sum_{i=1}^n \abs{\widehat{m}_n^\mu - \mu}^3 = \lim_{n \to \infty} \frac{1}{\sqrt{n}} \abs{\widehat{m}_n^\mu - \mu}^3 = 0, \label{other_term_M-Z-SLLN}
\end{align}
a.s. uniformly in $\mu \in \mathcal{I}$, where (\ref{M-Z-SLLN}) follows from (\ref{diffusion_second_moment_uniform_integrable}) and a uniform version of the Marcinkiewicz-Zygmund SLLN (Theorem 1(ii) of \cite{waudby-smith_etal2024} with $q = 2/3$), and (\ref{other_term_M-Z-SLLN}) follows from (\ref{diffusion_slln_uniform_2}).
Then, using (\ref{uniform_variance_lower_bound_2}), (\ref{diffusion_slln_uniform_3}), (\ref{M-Z-SLLN}) and (\ref{other_term_M-Z-SLLN}), the right side of (\ref{berry_esseen_uniform_bound}) converges to zero as $n \to \infty$, a.s. uniformly in $\mu \in \mathcal{I}$.
Using (\ref{uniform_variance_lower_bound_2}), (\ref{diffusion_slln_uniform_3}), and the Lipschitz continuity properties of $\Phi(\cdot)$, we have
\begin{align}
    \lim_{n \to \infty} \sup_{x \in \mathbb{R}} \abs{\Phi\left( \frac{x}{\widehat{\sigma}_n^\mu} \right) - \Phi\left( \frac{x}{\sigma^\mu} \right)} = 0, \label{application_Polya_theorem}
\end{align}
a.s. uniformly in $\mu \in \mathcal{I}$.
Using the convergence results for (\ref{berry_esseen_uniform_bound}) and (\ref{application_Polya_theorem}), the desired conclusion in (\ref{diffusion_bootstrap_clt}) is established.
\halmos
\endproof

\section{Weak Convergence Technical Lemmas} \label{diffusion_appC}

\begin{lemma}[Generalized Continuous Mapping Theorem] \label{diffusion_lemma0}
Let $f$ and $f^n$, $n \ge 1$, be measurable functions that map from the metric space $(\mathcal{S}_1,r_1)$ to the separable metric space $(\mathcal{S}_2,r_2)$.
Let $E$ be the set of $x \in \mathcal{S}_1$ such that $f^n(x^n) \to f(x)$ fails for some sequence $x^n$, $n \ge 1$, with $x^n \to x$ in $\mathcal{S}_1$.
If $\xi^n \Rightarrow \xi$ in $(\mathcal{S}_1,r_1)$ and $P(\xi \in E) = 0$, then $f^n(\xi^n) \Rightarrow f(\xi)$ in $(\mathcal{S}_2,r_2)$.
(See Theorem 3.4.4 of \cite{whitt_2002}.)
\end{lemma}

\begin{lemma}[Tightness of Multi-dimensional Processes]  \label{diffusion_lemma5}
A sequence of process $\xi^n = (\xi_1^n,\dots,\xi_d^n)$ is tight in $D_d[0,\infty)$ if each $\xi_j^n$ and each $\xi_j^n + \xi_k^n$ are tight in $D[0,\infty)$, for all $1 \le j, k \le d$. 
(See Problem 22 of Chapter 3 of \cite{ethier_etal1986}.)
\end{lemma}

\begin{lemma}[Simple Sufficient Conditions for Tightness] \label{diffusion_lemma6}
A sequence of processes $\xi^n$ in $D[0,\infty)$ adapted to filtration $(\mathcal{F}^n_t, \, t \ge 0)$ is tight if, for each $T > 0$,
\begin{align}
\lim_{a \to \infty} \sup_n \P\left(\sup_{0 \le t \le T} \abs{\xi^n(t)} > a\right) = 0, \label{diffusion_tightness1} \tag{T1}
\end{align}
and there exists a collection of non-negative random variables $\{ A^n_\delta(T), \, n \ge 1, \delta > 0 \}$ such that
\begin{align}
\E\left[ \left(\xi^n(t+v) - \xi^n(t)\right)^2 \mid \mathcal{F}^n_t \right] \le \E\left[ A^n_\delta(T) \mid \mathcal{F}^n_t \right] \label{diffusion_tightness2} \tag{T2}
\end{align}
a.s. for $0 \le t \le T$ and $0 \le v \le \delta$, and
\begin{align}
\lim_{\delta \downarrow 0} \limsup_{n \to \infty} \E\left[ A^n_\delta(T) \right] = 0. \label{diffusion_tightness3} \tag{T3}
\end{align}
(See Lemma 3.11 from \cite{whitt_2007}, which is adapted from \cite{ethier_etal1986}.)
\end{lemma}

\begin{lemma}[Martingale Functional Central Limit Theorem] \label{diffusion_lemma7}
Let $Y^n(i) \in \mathbb{R}^m$, $n \ge 1$ be a martingale difference sequence adapted to the filtration $\mathcal{F}^n_i$ for $i = 1,2,\dots$.
Suppose for any $t > 0$, the following conditions (\ref{diffusion_martingale1}) and (\ref{diffusion_martingale2}) hold as $n \to \infty$. \\
There exists a symmetric positive-definite matrix $\Sigma$ such that
\begin{align}
    \frac{1}{n} \sum_{i=1}^{\lfloor nt \rfloor} \E\left[Y^n(i) Y^n(i)^{\top} \mid \mathcal{F}^n_{i-1}\right] \overset{\P}{\to} t \Sigma. \label{diffusion_martingale1} \tag{M1}
\end{align}
For any $\epsilon > 0$ and each component $k = 1,\dots,m$,
\begin{align}
    \frac{1}{n} \sum_{i=1}^{\lfloor nt \rfloor} \E\left[ Y^n_k(i)^2 \indic{\abs{Y^n_k(i)} > \epsilon \sqrt{n}} \mid \mathcal{F}^n_{i-1} \right] \overset{\P}{\to} 0. \label{diffusion_martingale2} \tag{M2}
\end{align}
Then,
\begin{align}
    \frac{1}{\sqrt{n}} \sum_{i=1}^{\lfloor n \cdot \rfloor} Y^n(i) \Rightarrow W(\cdot) \nonumber
\end{align}
in $D_m[0,\infty)$, where $W$ is $m$-dimensional Brownian motion with covariance matrix $\Sigma$.
\end{lemma}

\end{APPENDICES}

% Acknowledgments here
%\ACKNOWLEDGMENT{We would like to express our sincere gratitude to [acknowledge individuals, organizations, or institutions] for their invaluable contributions to this research. We are also grateful to [mention any additional acknowledgements, such as technical assistance, data providers, or colleagues] for their support and assistance throughout the course of this work.}

% References here (outcomment the appropriate case)

% CASE 1: BiBTeX used to constantly update the references
%   (while the paper is being written).
%\bibliographystyle{informs2014} % outcomment this and next line in Case 1
%\bibliography{<your bib file(s)>} % if more than one, comma separated

%\bibliographystyle{informs2014} % outcomment this and next line in Case 1
%\bibliography{sample} % if more than one, comma separated

% CASE 2: BiBTeX used to generate mypaper.bbl (to be further fine tuned)
%\input{mypaper.bbl} % outcomment this line in Case 2

%If you don't use BiBTex, you can manually itemize references as shown below.

\bibliographystyle{informs2014} % or try abbrvnat or unsrtnat
\bibliography{references} % refers to references.bib

%%%%%%%%%%%%%%%%%
\end{document}